\documentclass[11pt]{article}

\usepackage{zingpaper}
\usepackage{colortbl}
\usepackage{placeins}
\usepackage{flafter}
\ifdefined\IfSocketPlugExistsT
  \IfSocketPlugExistsT{build/column/outputbox}{floats-space-footnotes}{%
    \AssignSocketPlug{build/column/outputbox}{floats-space-footnotes}}
\fi
\papertitle{\texorpdfstring{\fontsize{20}{24}\selectfont}{}Zing-1.0: A Playable World System\texorpdfstring{\\}{ }Built on Controllable Rendering}
\paperauthors{Zing Team}    
\paperaffiliations{SeedLeap.ai}
\papernote{Technical Report}
\paperrightlogo{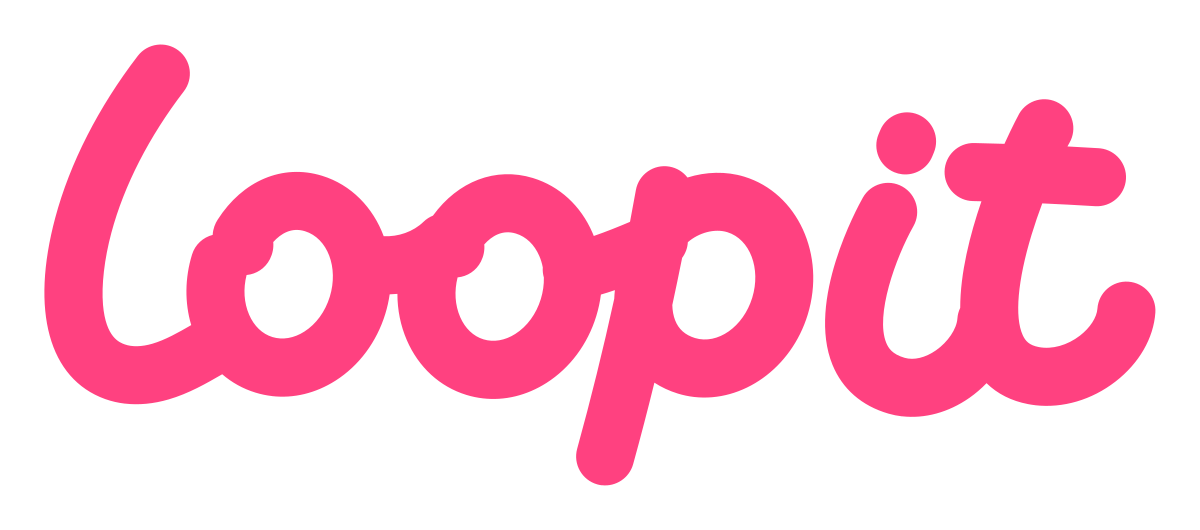}
\paperrighticon{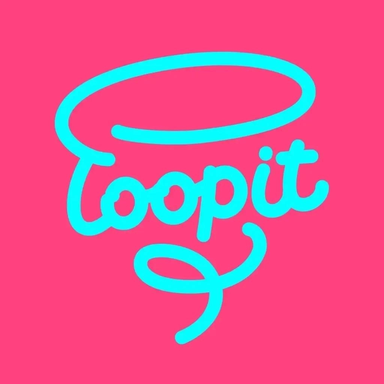}
\paperabstract{%
  We present \textbf{Zing-1.0}, a \textbf{Playable World System} built on \textbf{Controllable Rendering}. An executable engine resolves player actions under the world's rules and produces proxy video from the updated state. A streaming video model synthesizes the player's view from this proxy, and the runtime delivers it to inform subsequent actions. We make four contributions:
  \mbox{\textbf{(1) Principle and system.}} We systematically formulate Controllable Rendering and define the interfaces among world construction, world execution, observation synthesis, and the interaction runtime. To our knowledge, this is the first implementation-independent specification of a Playable World System.
  \mbox{\textbf{(2) AI-native engine.}} WorldKit provides structured authoring tools and reusable character, camera, and control components for coding agents to construct executable proxy worlds. It supports extensible interactions and precise geometric motion control while separating executable behavior from final visual appearance, allowing agents to focus on world-specific rules and interactions rather than low-level control implementation and photorealistic asset production.
  \mbox{\textbf{(3) Streaming renderer.}} We adapt MiniMax-H3 for block-causal autoregressive generation and incremental decoding. Proxy video constrains structure and motion, while text and separate scene and subject references specify appearance. We train the generator on data spanning diverse tasks and distill it to generate each block in a few denoising steps. The renderer leads the compared streaming editors on OpenVE (3.90) and LongV2VBench (3.44). On StreamEval, Zing-1.0 scores 57.86 on instruction editing and 54.06 on reference editing, exceeding the published baselines on both tracks.
  \mbox{\textbf{(4) Interaction infrastructure.}} We build an Interaction Runtime that routes player actions to WorldKit, coordinates proxy capture and streaming rendering, and returns generated observations to the player while supporting in-play prompt and reference updates. Cross-block joint execution and sparse attention with compressed K/V reduce inference overhead. With the DiT on a single H200 and encoding and decoding deployed as separate services on one H100 each, the serving system achieves a steady-state throughput of 23.82 FPS at $832\times480$.%
}

\paperlinks{%
  \paperlinkicon{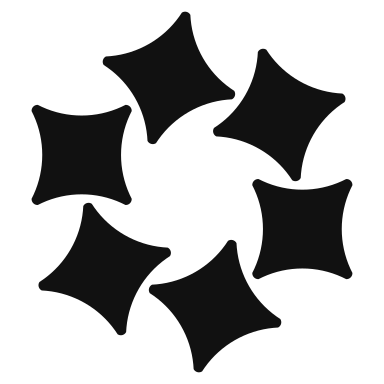}\textbf{Project:} \url{https://zing.loopit.me/}%
}

\begin{document}

\makezingtitle

\suppressfloats[t]
\begin{figure}[t]
  \centering
  \input{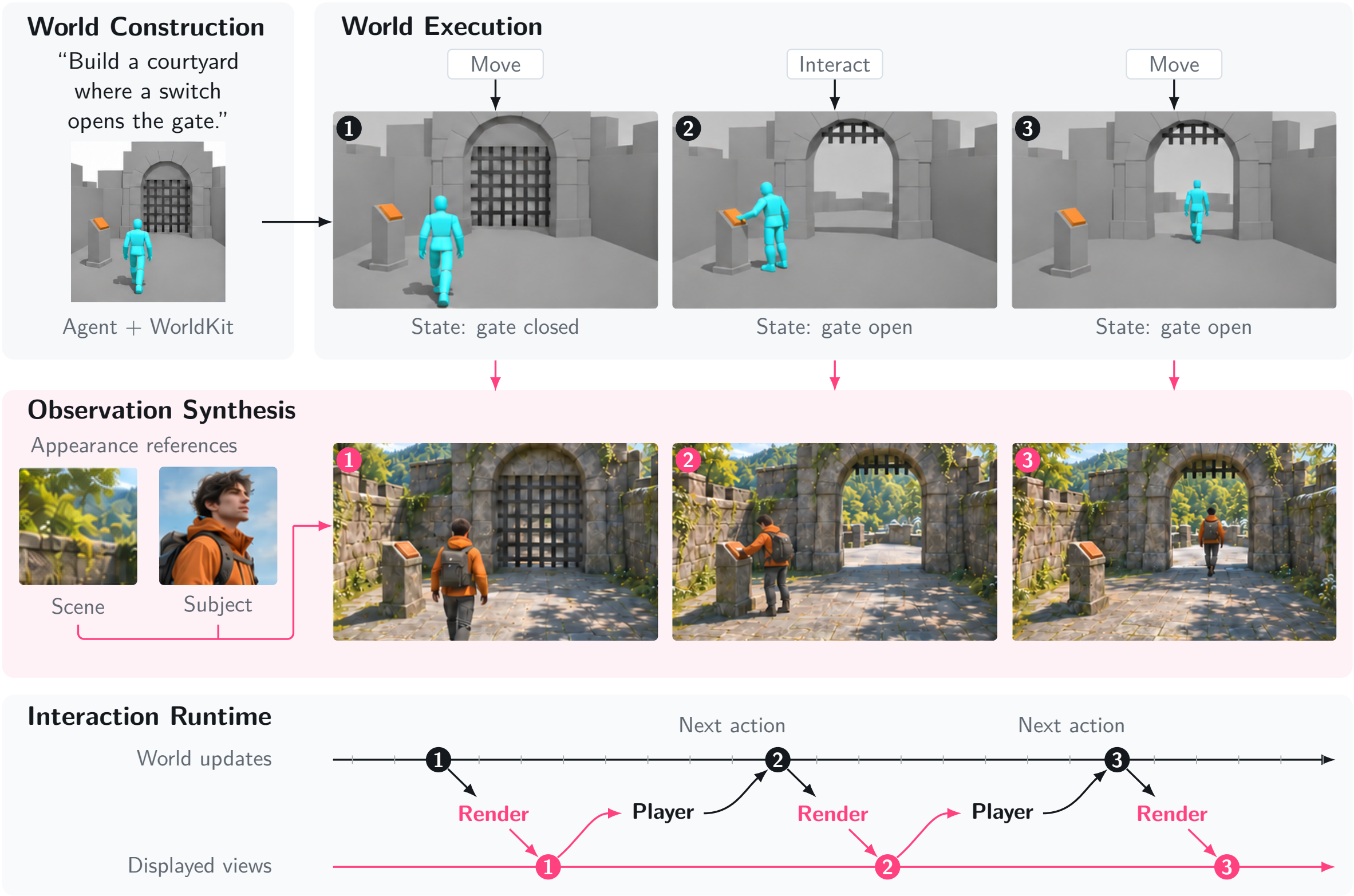}
  \caption{Conceptual overview of Zing-1.0 as a Playable World System. World construction produces an executable WorldKit world. Player actions update its state, and proxy frames guide observation synthesis under separate scene and subject references. The interaction runtime links displayed views to their source updates and returns feedback to the player.}
  \label{fig:teaser}
\end{figure}

\section{Introduction}
\label{sec:introduction}

Interactive video world models~\citep{parkerholder2025genie3,hadsell2026genie4,alibaba2026happyoyster} support closed-loop interaction through a generated video stream. The model predicts subsequent frames from the visual history and current user input, and the resulting observations inform the user's next decision. Within this loop, users can issue gameplay commands through a keyboard or mouse~\citep{valevski2025gamengen,decart2024oasis,li2025gamecraft}, direct the viewpoint through camera trajectories~\citep{he2025cameractrlii,nam2026worldcam,robbyant2026lingbotworld} or discrete camera motions~\citep{huang2026disco}, or guide the evolution of the scene through online prompt updates~\citep{longlive}. A single predictor can also combine inputs, pairing discrete keyboard actions with continuous mouse inputs~\citep{he2026matrixgame2,wang2026matrixgame3} or camera poses~\citep{sun2026worldplay}. Zing-0.5~\citep{chen2026zing05} extends joint control beyond navigation by combining keyboard commands for moving and turning the controlled subject with text prompts that can alter the environment or introduce events during generation.

In these models, the generated video must show how the environment responds to the player's action. A key distinction is that a player command specifies an action to attempt, not its outcome. The world state describes the environment at a given time, including the entities present and their current properties and relationships. Given a command and the current state, the environment's rules and constraints determine whether the action succeeds and what state transition follows~\citep{kaelbling1998pomdp,worldlabs2026taxonomy}. In Zing-0.5~\citep{chen2026zing05} and similar interactive video world models, the command directly conditions the prediction of the next observation without a separate execution step. Their video generator implicitly models the state transition while synthesizing its visible consequences. An observation can therefore be visually plausible even when the transition it implies violates the environment's constraints. Current video world models do not yet reliably sustain both rule-consistent state evolution and visually coherent generation over extended interactions~\citep{ding2026playworld,ying2026wbench}. This motivates resolving actions independently of the synthesis of their visible consequences.

\enlargethispage{-2\baselineskip}
A recent line of systems does this by moving state evolution out of the video model: an engine or an executable program resolves each command and maintains the world state, and a video model renders the result from state-derived inputs~\citep{chen2026codeworldmodel,zhan2026magpie,huang2026programmableworldmodel,chi2026agentgarten,zhang2026matrixgame4,wu2026boxworld}. We formulate the principle behind this move as \textbf{Controllable Rendering}: an explicit, executable control process decides what happens in the world, and a generative video model renders what the player sees. A system follows the principle when (C1) every change of world state results from executing a command under the world's rules and is never inferred from generated pixels, and (C2) the video model receives the result of that execution rather than the command itself (\secref{sec:cr-spec}). Controllable Rendering is therefore not an additional control signal for a video model. It reassigns responsibility between two technologies with complementary limits: a conventional engine executes rules exactly but depends on authored assets for its final imagery~\citep{epicunrealgameplay}, whereas a video model generates open-ended appearance but cannot guarantee that rules hold. Existing systems describe this division through their own components (\secref{sec:related-work}). What a complete system must do has not been specified independently of an implementation, and the specification of appearance and the association between displayed frames and state updates are often left implicit.

We therefore specify a \mbox{\textbf{Playable World System}}: the complete loop that Controllable Rendering requires, stated as four responsibilities and the interfaces between them rather than as the modules of one implementation (\secref{sec:cr-spec}). \mbox{\textbf{World Construction}} supplies an executable world and, separately, its appearance conditions. \mbox{\textbf{World Execution}} resolves commands under the world's rules and advances the state. \mbox{\textbf{Observation Synthesis}} generates the player's view from conditions derived from that state together with the appearance conditions. \mbox{\textbf{Interaction Runtime}} delivers each observation in time to inform the next action and keeps it associated with the state update it depicts. Construction and execution form the control side, synthesis forms the rendering side, and the runtime closes the loop between them. Generated observations influence the world only through the player's subsequent actions.

\textbf{Zing-1.0} implements this specification with an AI-native game engine on the control side and a streaming video model on the rendering side. Coding agents use WorldKit to author, playtest, and package executable proxy worlds, and WorldKit executes player commands in the delivered world. As the state advances, WorldKit renders it as a \emph{proxy video} from the player's viewpoint. Its simplified RGB frames convey scene layout, camera motion, and entity motion without specifying final appearance. The renderer conditions on this video rather than raw player inputs. We adapt MiniMax-H3~\citep{minimax2026h3} for block-causal rendering while retaining its multi-reference conditioning, so that separate images can specify scene and subject appearance. Text indicates the intended correspondence between each reference and its scene or subject in the proxy world. The runtime streams the generated observations back to the player and associates them with the state updates they depict.

This division gives one world capabilities from both sides (\figref{fig:teaser}). \textbf{From control}, WorldKit supports \emph{extensible interactions}: mechanics such as vehicle driving, object pickup, and switch interactions are defined in the engine, and their visible effects reach the renderer through the same proxy interface. It also provides \emph{geometric motion control} by applying precise displacements and turn angles before rendering, rather than leaving their magnitude to the video model. \textbf{From rendering}, Zing-1.0 enables \emph{rapid appearance iteration} by re-rendering the same executable proxy with new scene or subject references, without requiring changes to its assets, and \emph{in-play prompt-driven events} that alter what is rendered during an active session. As in Zing-0.5~\citep{chen2026zing05}, event prompts affect subsequent frames; in Zing-1.0, they do so without directly updating WorldKit state. \tabref{tab:playable-world-comparison} compares these capabilities with representative interactive video world models~\citep{parkerholder2025genie3,hadsell2026genie4,alibaba2026happyoyster}, generative 3D world models~\citep{worldlabs2025marble,worldlabs2026atlas}, and conventional AAA engines~\citep{epicunrealgameplay}.

\begingroup
\setlength{\intextsep}{6pt}
\begin{table}[H]
  \centering
  \caption{Qualitative comparison of Controllable Rendering, as realized in Zing-1.0, with other approaches to playable worlds.}
  \label{tab:playable-world-comparison}
  \small
  \setlength{\tabcolsep}{4pt}
  \renewcommand{\arraystretch}{1.75}
  \renewcommand{\tabularxcolumn}[1]{m{#1}}
  \setlength{\aboverulesep}{0pt}
  \setlength{\belowrulesep}{0pt}
  \begin{tabularx}{\linewidth}{>{\columncolor{zingSoft}\color{zingPink}\raggedright\arraybackslash\itshape}m{0.265\linewidth}>{\columncolor{zingPinkWash}\centering\arraybackslash\bfseries}X*{3}{>{\centering\arraybackslash}X}}
    \arrayrulecolor{black}\toprule[0.9pt]
    \cellcolor{zingRule!50}{\color{black}\upshape\bfseries Capability} & \cellcolor{zingPink!16}Zing-1.0\par {\scriptsize\mdseries\mbox{Controllable Rendering}} & \textbf{Interactive video world model} & \textbf{Generative 3D world model} & \textbf{Conventional AAA engine} \\
    \arrayrulecolor{zingRule}\midrule[0.5pt]
    Rapid appearance iteration & $\pmb{\checkmark}$ & $\checkmark$ & $\checkmark$ & $\times$ \\
    Extensible interactions & High & Low & Low & High \\
    Geometric motion control & High & Medium & High & High \\
    In-play prompt-driven events & High & High & Low & Low \\
    \arrayrulecolor{black}\bottomrule[0.6pt]
  \end{tabularx}
\end{table}
\endgroup

\Needspace{11\baselineskip}
We make four contributions: a principle with its system specification, and three components that realize it in Zing-1.0.
\begin{itemize}
  \item \textbf{Principle and system.} We formulate Controllable Rendering as a named principle with an explicit criterion and specify the Playable World System that it requires. The specification connects \emph{World Construction}, \emph{World Execution}, \emph{Observation Synthesis}, and the \emph{Interaction Runtime} around one executable world, and Zing-1.0 implements all four. To our knowledge, this is the first specification of the complete loop at the level of responsibilities and interfaces, independent of any one implementation.
  \item \textbf{Engine.} Control requires worlds that agents can author and that execute reproducibly. WorldKit provides structured interfaces and prebuilt character, camera, and control assets for language-model-assisted \emph{World Construction}. During \emph{World Execution}, it executes the resulting worlds and outputs proxy video, leaving final appearance to the renderer.
  \item \textbf{Renderer.} Rendering requires following the proxy's structure while taking appearance from references and text. For \emph{Observation Synthesis}, we adapt the visual VAE and diffusion transformer of MiniMax-H3 into a block-causal renderer that combines proxy video with separate scene and subject references, using aligned proxy--target pairs, general video-editing data, and staged causal training and distillation.
  \item \textbf{Infrastructure.} The loop requires observations that arrive in time. Within the \emph{Interaction Runtime}, we optimize the DiT inference path through joint execution, frame-aligned sparse attention, and compressed KV state, and coordinate proxy capture, generation, decoding, and playback to align delivered observations with the corresponding world updates.
\end{itemize}

\section{Related Work}
\label{sec:related-work}

\paragraph{Interactive video world models.}
Action-conditioned video generation has progressed from reproducing the visual dynamics of particular games to supporting navigation in novel generated scenes~\citep{valevski2025gamengen,yu2025gamefactory,he2026matrixgame2}. Camera-conditioned models specify viewpoint motion geometrically through camera trajectories~\citep{he2025cameractrlii,nam2026worldcam,robbyant2026lingbotworld}. Beyond navigation, online text instructions allow users to request changes in character behavior or scene events during a rollout~\citep{longlive,tang2025gamecraft2}. Several models support both navigation and text-directed interventions in the same rollout~\citep{gao2026lingbotinfinity,chen2026hynomas}. Zing-0.5~\citep{chen2026zing05} uses separate conditioning streams for keyboard actions and text instructions so that either input can change as generation proceeds.
Over longer interactions, spatial memory helps preserve consistency across viewpoints~\citep{xiao2025worldmem,nam2026worldcam,sun2026worldplay}. To improve consistency with game mechanics, StatePlay~\citep{lin2026stateplay} jointly predicts game state and video. GameDirector~\citep{lin2026gamedirector} separates rule enforcement from rendering, but updates game state using generated observations and guides the video model through prompts. Controllable Rendering instead determines action outcomes by executing commands under the environment's rules, without inferring state from generated frames. In Zing-1.0, WorldKit executes each command before rendering and the renderer never receives the raw command, so displacement, collisions, and scores are determined by engine state rather than estimated by the video model.

\paragraph{Agent-based world construction.}
Under Controllable Rendering, the control side must exist before play, and agent-based construction is what makes it cheap to produce. Existing work ranges from producing editable 3D scenes to delivering testable games. WorldClaw~\citep{guo2026worldclaw} plans terrain and object placement from text and refines the resulting scene with editable assets. AutoUE~\citep{yin2026autoue} and Code2Games~\citep{wu2026code2games} address the additional task of coupling scenes to executable interactions in Unreal Engine: AutoUE generates gameplay code and runs automated playtests, while Code2Games uses compilation and playtesting feedback to revise generated worlds. In the browser-game setting, Game-Creation-Agents~\citep{tencent2026gca} verifies implemented mechanics through actual interaction with the generated game. For an engine-driven system, the relevant output of construction is not only a scene but an executable environment whose rules determine how its state changes. It need not include final appearance. In these systems, final appearance still depends on conventional engine assets and rendering, while high-quality character animation, movement, and camera behavior require suitable assets and carefully integrated runtime components.
Zing-1.0 asks the agent to build only a proxy world, supplies character, camera, and control as prebuilt assets, and leaves final appearance to the renderer.

\paragraph{Controllable Rendering systems.}
The systems closest to ours follow Controllable Rendering under other names, including code world models, programmable world models, and code worlds~\citep{chen2026codeworldmodel,huang2026programmableworldmodel,fang2026coder,chi2026agentgarten}. Each describes the division through its own components: a game engine and a render server~\citep{zhan2026magpie}, simulators and a shared neural renderer~\citep{chi2026agentgarten}, or an engine, a coding agent, and a video model~\citep{zhang2026matrixgame4}. They differ in which responsibilities they specify and in the interface between control and rendering.
Executable worlds constrain the layout and motion of generated observations through state-derived inputs, from proxy videos and rendered buffers~\citep{chen2026codeworldmodel,huang2026programmableworldmodel,lin2026alayaflash,liu2026anygamerenderer} to object layouts with explicit state attributes~\citep{wu2026boxworld}. Several interactive systems pair these structural controls with a first-frame image to specify appearance~\citep{edenlive2026,zhan2026magpie}, while Wizard~\citep{wu2026wizard} also uses skeletal proxies to bind separate character references to their subjects.
Streaming renderers use sliding-window bidirectional attention~\citep{lu2026streamrenderh3} or block-causal generation~\citep{chi2026agentgarten,zhang2026matrixgame4,fang2026solarrenderer}. To reduce streaming latency, Zing-1.0 uses short autoregressive blocks to limit proxy buffering and caches past keys and values to avoid recomputation. Its renderer conditions on separate scene and subject reference images.

\paragraph{Differences from prior systems.}
Zing-1.0 differs from these systems in three respects. First, appearance is specified per entity and persists throughout play. Magpie, Code World Model, and Matrix-Game 4.0 anchor appearance to a first frame~\citep{zhan2026magpie,chen2026codeworldmodel,zhang2026matrixgame4}, and AgentGarten uses a single appearance reference~\citep{chi2026agentgarten}. Zing-1.0 assigns separate reference images to the scene and to each subject, so the same executable world can be re-rendered by replacing its references.
Second, the renderer is also a general streaming video editor. To our knowledge, prior systems in this line do not report general video-editing benchmarks, whereas we evaluate the Zing-1.0 renderer on OpenVE~\citep{he2025openve}, LongV2VBench~\citep{xiao2026joyai}, and StreamEval~\citep{streameval2026}.
Third, the world that is validated is the world that is executed. A coding agent compiles, previews, and playtests a world before sealing it, the same engine runs the sealed package offline and online, and each proxy frame carries the simulation step that produced it.

\section{Playable World System}
\label{sec:system-design}

\subsection{Principle and System Architecture}
\label{sec:cr-spec}

Controllable Rendering separates world-state evolution from observation synthesis. An explicit, executable control process maintains the world state and updates it by executing player commands under the world's rules, whereas a generative video model synthesizes observations from the resulting state. Two conditions define this separation.
\begin{itemize}
  \item \textbf{C1 (state from execution).} Every change of world state results from executing a command under the world's rules.
  \item \textbf{C2 (rendering from state).} The video model is conditioned on the result of execution, not on the command itself.
\end{itemize}
Under C1, rendering errors may affect the player's decisions but do not directly alter world state, a property also implemented in earlier engine-based systems~\citep{zhan2026magpie,chi2026agentgarten}. Under C2, the outcome and magnitude of an action are determined by execution before rendering, while text and appearance references condition visual synthesis without directly updating world state.

A \emph{Playable World System} implements this separation through four responsibilities with explicit interfaces. \tabref{tab:pws-spec} specifies their required behavior and realization in Zing-1.0. Between execution and synthesis, a state-to-observation interface derives rendering conditions from updated world state, conveying scene structure, entity states, and interaction outcomes relevant to the player's view~\citep{vast2026projecteden,chi2026agentgarten}. Although appearance may vary, generated observations must preserve the visible consequences of execution that are relevant to interaction. These responsibilities define functional boundaries without prescribing a particular state representation or requiring separate implementations.

\begin{table}[H]
  \centering
  \caption{\textbf{Playable World System specification.} The scope, required behavior, and Zing-1.0 implementation of each responsibility.}
  \label{tab:pws-spec}
  \small
  \setlength{\tabcolsep}{5pt}
  \setlength{\extrarowheight}{2pt}
  \renewcommand{\arraystretch}{1.22}
  \renewcommand{\tabularxcolumn}[1]{m{#1}}
  \setlength{\aboverulesep}{0pt}
  \setlength{\belowrulesep}{0pt}
  \newcommand{\pwsscope}[1]{\newline{\footnotesize\upshape\color{zingMuted}#1}}
  \begin{tabularx}{\linewidth}{>{\columncolor{zingSoft}\raggedright\arraybackslash\itshape\leavevmode\color{zingPink}}m{0.205\linewidth}>{\raggedright\arraybackslash}X>{\columncolor{zingPinkWash}\raggedright\arraybackslash}m{0.335\linewidth}}
    \arrayrulecolor{black}\toprule[0.9pt]
    \cellcolor{zingRule!50}{\color{black}\upshape\bfseries Responsibility} & \textbf{Required behavior} & \cellcolor{zingPink!16}\textbf{In Zing-1.0} \\
    \arrayrulecolor{zingRule}\midrule[0.5pt]
    World Construction\pwsscope{Control} & Deliver an executable world with rules and initial state while specifying appearance separately. & Coding agents author, test, and package proxy worlds with WorldKit Creator. \\
    \midrule[0.35pt]
    World Execution\pwsscope{Control} & Update world state by executing player commands under the world's rules (C1). & WorldKit updates world state and produces RGB proxy frames. \\
    \midrule[0.35pt]
    Observation Synthesis\pwsscope{Rendering} & Generate the player's view from state-derived conditions and separate appearance inputs (C2). & A block-causal renderer uses proxy video, text, and appearance references. \\
    \midrule[0.35pt]
    Interaction Runtime\pwsscope{Across Control \& Rendering} & Forward player actions to world execution and return timely observations associated with their source state updates. & The runtime routes inputs, tracks proxy frames, and pipelines rendering and delivery back to the player. \\
    \arrayrulecolor{black}\bottomrule[0.6pt]
  \end{tabularx}
\end{table}

\begin{figure}[!t]
  \centering
  \begingroup
  \input{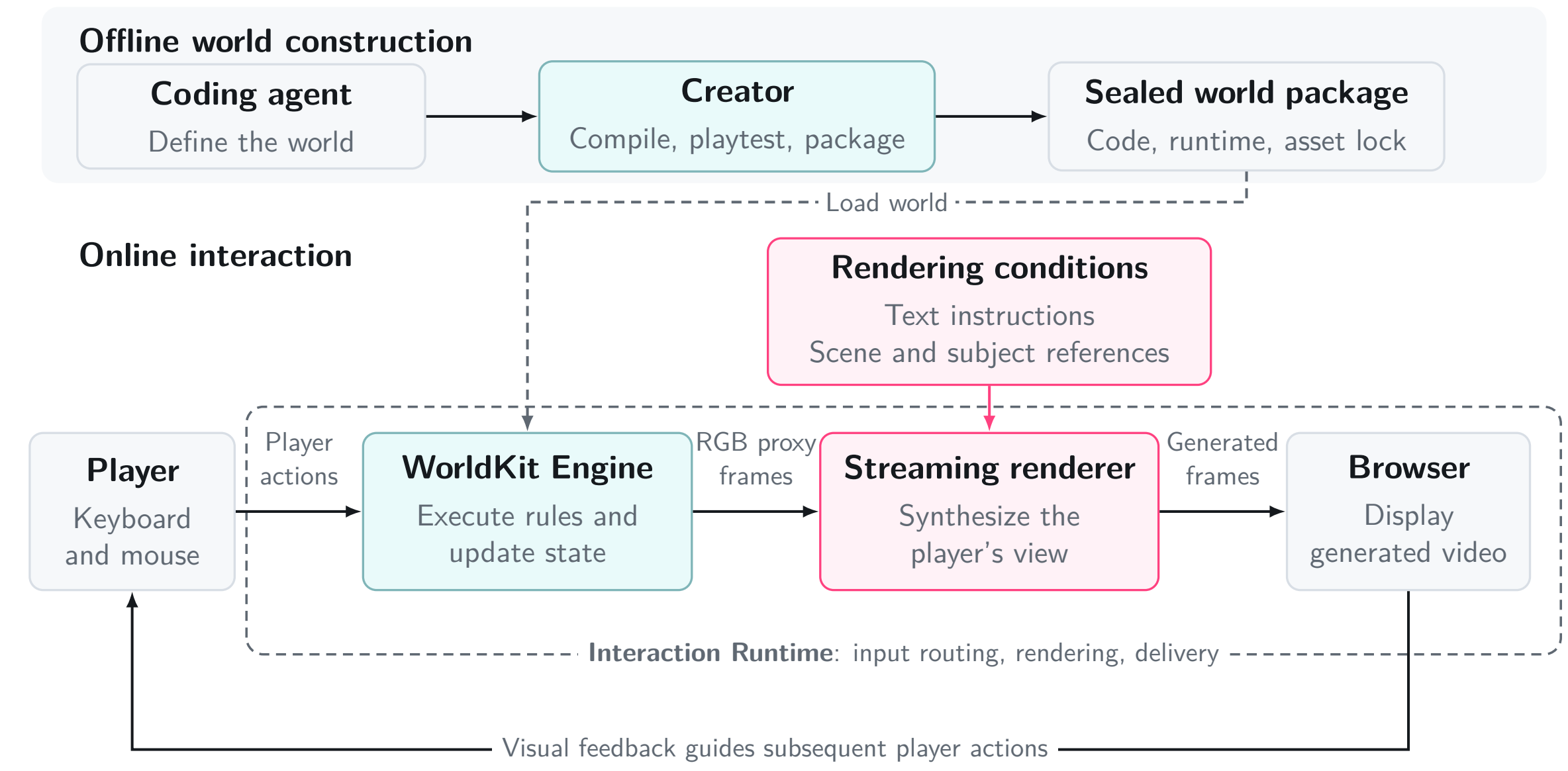}
  \resizebox{0.98\linewidth}{!}{%
  \begin{tikzpicture}[font=\sffamily\fontsize{9}{11}\selectfont]
    \path[use as bounding box] (-.45,-.6) rectangle (15.95,7.55);
    \fill[zingSoft,rounded corners=5pt] (-.2,5.65) rectangle (15.6,7.5);
    \node[rtitle] at (.05,7.15) {Offline world construction};
    \node[rstate,text width=3.3cm,minimum height=1.05cm] (agent) at (2,6.35) {{\bfseries\fontsize{10}{12}\selectfont Coding agent}\\[3pt]\textcolor{zingMuted}{Define the world}};
    \node[rsource,text width=3.8cm,minimum height=1.05cm] (creator) at (7.1,6.35) {{\bfseries\fontsize{10}{12}\selectfont Creator}\\[3pt]\textcolor{zingMuted}{Compile, playtest, package}};
    \node[rstate,text width=3.8cm,minimum height=1.05cm] (world) at (12.45,6.35) {{\bfseries\fontsize{10}{12}\selectfont Sealed world package}\\[3pt]\textcolor{zingMuted}{Code, runtime, asset lock}};
    \draw[rarrow] (agent.east)--(creator.west);
    \draw[rarrow] (creator.east)--(world.west);
    \node[rtitle] at (.05,4.9) {Online interaction};
    \draw[draw=zingMuted,line width=.7pt,densely dashed,rounded corners=5pt] (1.95,.7) rectangle (15.75,3.3);
    \node[rpink,text width=4.3cm,minimum height=1.45cm] (cond) at (9.75,4.3) {{\bfseries\fontsize{10}{12}\selectfont Rendering conditions}\\[3pt]\textcolor{zingMuted}{Text instructions\\Scene and subject references}};
    \node[rstate,text width=1.8cm,minimum height=1.65cm] (player) at (.75,2.2) {{\bfseries\fontsize{10}{12}\selectfont Player}\\[3pt]\textcolor{zingMuted}{Keyboard and mouse}};
    \node[rsource,text width=3.1cm,minimum height=1.65cm] (engine) at (4.9,2.2) {{\bfseries\fontsize{10}{12}\selectfont WorldKit Engine}\\[3pt]\textcolor{zingMuted}{Execute rules and update state}};
    \node[rpink,text width=3.2cm,minimum height=1.65cm] (renderer) at (9.75,2.2) {{\bfseries\fontsize{10}{12}\selectfont Streaming renderer}\\[3pt]\textcolor{zingMuted}{Synthesize the player's view}};
    \node[rstate,text width=2.2cm,minimum height=1.65cm] (browser) at (14.15,2.2) {{\bfseries\fontsize{10}{12}\selectfont Browser}\\[3pt]\textcolor{zingMuted}{Display generated video}};
    \draw[rguide] (world.south)--(12.45,5.45)--node[midway,rlabel,fill=white,inner sep=2pt] {Load world} (4.9,5.45)--(engine.north);
    \draw[rarrow] (player.east)--node[midway,above=4pt,rlabel] {Player\\actions} (engine.west);
    \draw[rarrow] (engine.east)--node[midway,above=4pt,rlabel] {RGB proxy\\frames} (renderer.west);
    \draw[rarrow] (renderer.east)--node[midway,above=4pt,rlabel] {Generated\\frames} (browser.west);
    \draw[rpinkarrow] (cond.south)--(renderer.north);
    \draw[rarrow] (browser.south)--(14.15,-.3)--(.75,-.3)--(player.south);
    \node[rlabel,fill=white,inner sep=3pt] at (7.5,-.3) {Visual feedback guides subsequent player actions};
    \node[rlabel,fill=white,inner sep=3pt] at (8.85,.7) {\textbf{Interaction Runtime}: input routing, rendering, delivery};
  \end{tikzpicture}}
  \endgroup
  \caption{Zing-1.0 system architecture, with the control side (the coding agent and WorldKit) and the rendering side (the streaming renderer) connected by proxy frames. Offline construction produces an executable proxy world. Online, player actions update WorldKit state; proxy frames convey its visible consequences to the streaming renderer, while text and references specify appearance. Generated video returns to the browser as feedback for subsequent actions. The runtime coordinates the input-to-display path without feeding generated frames directly into world state.}
  \label{fig:system-architecture}
\end{figure}

\figref{fig:system-architecture} illustrates how Zing-1.0 connects offline world construction to online execution and rendering. During construction, a coding agent uses WorldKit to define, test, and package an executable proxy world. During online play, WorldKit executes player commands under the world's rules and produces proxy RGB frames from the updated state. The streaming renderer combines these frames with text and appearance references to synthesize the player's view. The interaction runtime forwards player inputs to execution, coordinates capture, generation, delivery, and playback, and returns the generated video to the browser as feedback for subsequent actions.

\Needspace{9\baselineskip}
\subsection{WorldKit Engine}
\label{sec:worldkit}

WorldKit supports agent-based \emph{World Construction} and performs \emph{World Execution} in Zing-1.0. Its design follows three principles.

\paragraph{Interfaces designed for coding agents.}
WorldKit exposes authoring capabilities through structured interfaces that coding agents can query and inspect. Worlds are TypeScript programs built on Three.js~\citep{threejs} and Rapier~\citep{rapier} for scene graphs and rigid-body physics, allowing agents to use familiar programming patterns. Structured, versioned asset records support asset selection, while authoring tools return compilation diagnostics, preview images, and playtest traces in machine-readable form for behavioral verification and iterative revision.

\paragraph{Execution in the engine, appearance in the renderer.}
WorldKit maintains entity state, world rules, collision handling, and camera behavior while leaving final appearance to the renderer. Proxy authoring uses simple geometry and restrained, distinguishable colors that remain consistent for each subject during play. Shadows, reflections, detailed materials, and atmosphere are synthesized by the renderer described in \secref{sec:streaming-renderer}. This separation reduces the need for photorealistic asset production during world construction without changing the engine-defined rules and interaction outcomes.

\paragraph{Prebuilt Character, Camera, and Control.}
Coding agents can lay out scenes and implement world rules, but high-quality character animation, movement, and camera behavior require suitable assets and carefully integrated runtime components. These include rigged and skinned models, motion clips that transition smoothly between locomotion states, controllers tuned to each mode of movement, and cameras that follow a subject without being occluded. WorldKit provides these Character, Camera, and Control (3C) capabilities as reusable assets and runtime components, allowing agents to focus on world-specific content and interaction logic.

\begin{figure}[!t]
  \centering
  \begingroup
  \input{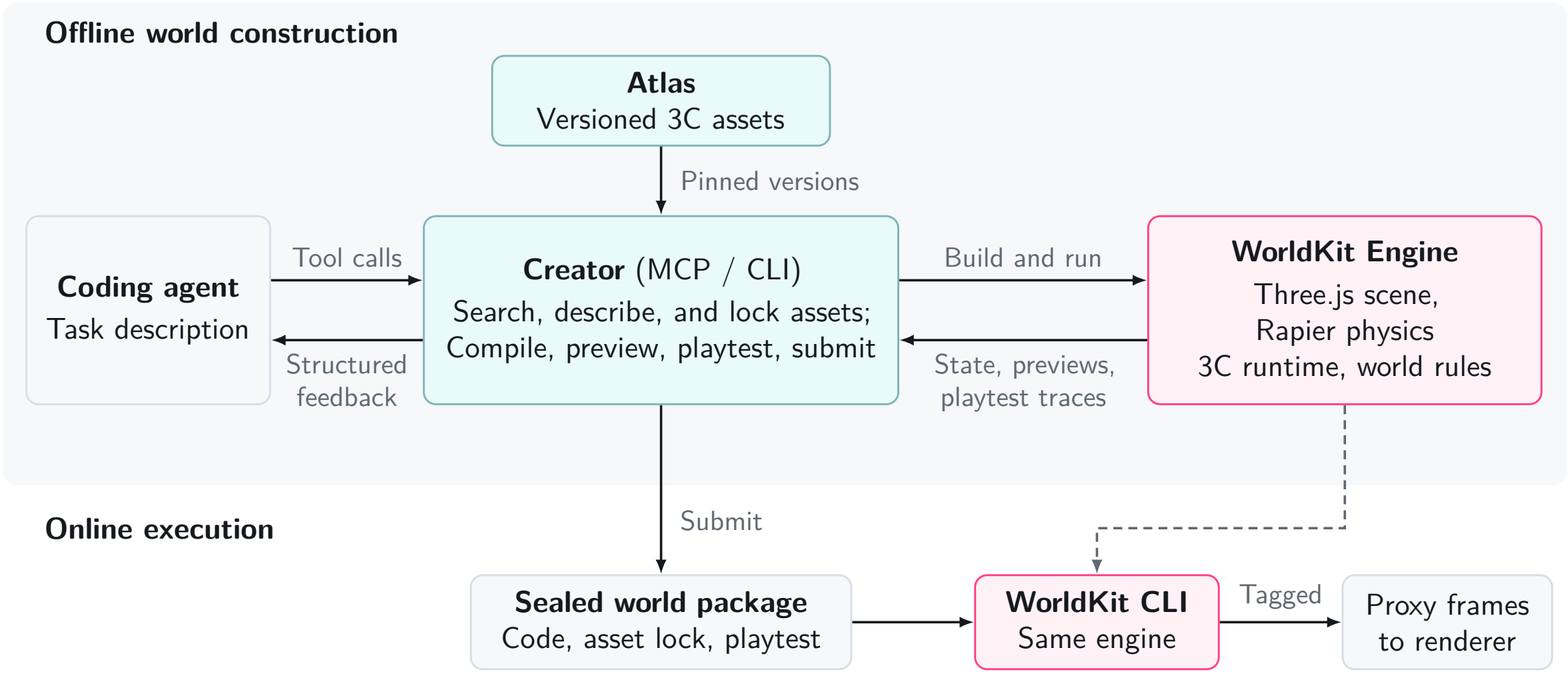}
  \resizebox{0.98\linewidth}{!}{%
  \begin{tikzpicture}[font=\sffamily\fontsize{10}{12}\selectfont,
    linklabel/.style={rlabel,font=\sffamily\fontsize{9.2}{11}\selectfont,inner sep=0pt}]
    \path[use as bounding box] (-.25,-.2) rectangle (18.15,8.4);
    \fill[zingSoft,rounded corners=5pt] (-.2,2.7) rectangle (18.1,8.35);
    \node[rtitle] at (.15,8.0) {Offline world construction};
    \node[rstate,text width=2.5cm,minimum height=2.2cm] (agent) at (1.5,4.75) {\textbf{Coding agent}\\[2pt]Task description};
    \node[rsource,text width=5.2cm,minimum height=2.2cm] (creator) at (7.5,4.75) {\textbf{Creator} (MCP / CLI)\\[2pt]Search, describe, and lock assets;\\Compile, preview, playtest, submit};
    \node[rsource,text width=3.6cm,minimum height=1.05cm] (atlas) at (7.5,7.2) {\textbf{Atlas}\\Versioned 3C assets};
    \node[rpink,text width=4.25cm,minimum height=2.2cm] (engine) at (15.5,4.75) {\textbf{WorldKit Engine}\\[2pt]Three.js scene, Rapier physics\\3C runtime, world rules};
    \draw[rarrow] ([yshift=.35cm]agent.east)--node[midway,above=4pt,linklabel] {Tool calls} ([yshift=.35cm]creator.west);
    \draw[rarrow] ([yshift=-.35cm]creator.west)--node[midway,below=4pt,linklabel,text width=1.6cm] {Structured\\feedback} ([yshift=-.35cm]agent.east);
    \draw[rarrow] (atlas.south)--node[midway,right=6pt,linklabel] {Pinned versions} (creator.north);
    \draw[rarrow] ([yshift=.35cm]creator.east)--node[midway,above=4pt,linklabel] {Build and run} ([yshift=.35cm]engine.west);
    \draw[rarrow] ([yshift=-.35cm]engine.west)--node[midway,below=4pt,linklabel,text width=2.65cm] {State, previews,\\playtest traces} ([yshift=-.35cm]creator.east);
    \node[rstate,text width=4.1cm,minimum height=1.1cm] (package) at (7.5,1.1) {\textbf{Sealed world package}\\Code, asset lock, playtest};
    \draw[rarrow] (creator.south)--node[pos=.68,right=6pt,linklabel] {Submit} (package.north);
    \node[rtitle] at (.15,2.2) {Online execution};
    \node[rpink,text width=2.5cm,minimum height=1.1cm] (cli) at (12.6,1.1) {\textbf{WorldKit CLI}\\Same engine};
    \node[rstate,text width=2.1cm,minimum height=1.1cm] (proxy) at (16.7,1.1) {Proxy frames\\to renderer};
    \draw[rarrow] (package.east)--(cli.west);
    \draw[rarrow] (cli.east)--node[midway,above=4pt,linklabel] {Tagged} (proxy.west);
    \draw[rguide] (engine.south)--(15.5,2.2)--(12.6,2.2)--(cli.north);
  \end{tikzpicture}}
  \endgroup
  \caption{WorldKit components. A coding agent authors a world through Creator tools, which retrieve pinned 3C assets from Atlas and build, run, and observe the world in the WorldKit Engine. Each tool returns structured feedback that the agent can read. The sealed package is executed online by the WorldKit CLI on the same engine, which emits proxy frames tagged with their simulation step.}
  \label{fig:worldkit-components}
\end{figure}

\figref{fig:worldkit-components} shows the resulting components. \emph{Atlas}, our asset library for WorldKit, stores versioned 3C assets. \emph{Creator} exposes authoring, testing, and packaging tools to coding agents through the Model Context Protocol (MCP)~\citep{mcp} and a JSON command-line interface. The \emph{WorldKit Engine} executes scene updates, physics, 3C, and world rules during both authoring and online play. The \emph{WorldKit CLI} loads a sealed world package for an online session.

\Needspace{6\baselineskip}
\subsubsection{Offline Proxy World Authoring}
\label{sec:world-authoring}

\paragraph{3C assets.}
\emph{Character} assets are versioned records that bind a model to its skeleton, motion clips, attachment sockets, and declared capabilities. Clips are referenced through semantic aliases, such as an idle loop or a standing jump, and through a declared skeleton type, so that motion packs can be shared by every subject with a compatible skeleton. For example, version 0.1.8 of the humanoid asset, used for the engine checks in \secref{sec:engine-results}, has a 101-joint skeleton and draws its motions from three version-locked motion packs. \emph{Control} is organized into seven motion families: humanoids, ground vehicles, watercraft, aircraft, flying creatures, underwater subjects, and spacecraft. A controller converts player input into motion intent, and physics and world rules determine the result. \emph{Camera} provides first-person, third-person, and over-the-shoulder modes, together with named views and scripted shots defined by each world. A follow camera mounted on a spring arm shortens the arm when geometry blocks the line of sight and extends it again once the view is clear.

Atlas distinguishes automated asset checks from per-version runtime acceptance. Upload checks verify byte integrity, content hashes, and file formats, and detect skeletons and motion clips where possible. Runtime acceptance requires a reviewer to check skeleton compatibility, wheel and seat placement, and controller behavior in the runtime before releasing an asset version for production.

\paragraph{Authoring workflow.}
Creator supports a structured authoring workflow in which each stage returns feedback that the agent can inspect. The agent first queries the authoring schema and minimal examples, then searches Atlas, inspects candidate assets, and seals an asset plan that pins exact versions and their dependencies. Based on this plan, it defines the scene, entities, cameras, and rules, compiles the program, and validates the world. Verification combines previews from the opening view, from above, and around key objects with playtests that replay scripted player inputs. Playtest traces record positions, actions, animation clips, camera state, and input events at each sampled simulation step, allowing the agent to compare execution with the intended behavior. After validation, submission produces a sealed package containing the executable world, frozen runtime, asset lock, source code, previews, and playtest record.

\subsubsection{Online World Execution}
\label{sec:world-execution}

During online play, WorldKit advances the world at a fixed simulation step of 1/60~s. At each step, the engine processes player input, executes world rules, converts input into motion intent through the active controllers, solves physics with Rapier, updates animation and camera, and publishes the new state. Presentation runs independently of this schedule and samples the latest available state. Captured proxy frames carry the simulation step, world revision, source-frame identifier, and capture time, enabling the Interaction Runtime to associate generated observations with their source world updates.

The WorldKit CLI loads a sealed package, receives player input, and emits raw RGB proxy frames in one of two output modes, 480p or 720p. Game commands, such as entering a vehicle or picking up an object, are routed to WorldKit through the input channel, whereas text instructions and appearance references are supplied only to the renderer as conditioning inputs. Generated frames are never read back into the engine, so position, collision, and scoring are always determined by WorldKit state.

\Needspace{7\baselineskip}
\subsection{Rendering Side: Streaming Renderer}
\label{sec:streaming-renderer}

The streaming renderer is the rendering side of Zing-1.0 and performs \emph{Observation Synthesis}. Controllable Rendering sets its task: follow the structure and motion of the proxy, take appearance from references and text, and do both causally as the proxy stream arrives. We describe the model, the data, its training, and the decoder that meet these requirements.

\FloatBarrier
\subsubsection{Model Architecture}
\label{sec:render-model}

\begin{figure}[H]
  \centering
  \setlength{\parskip}{0pt}
  \captionsetup{skip=4pt}
  \input{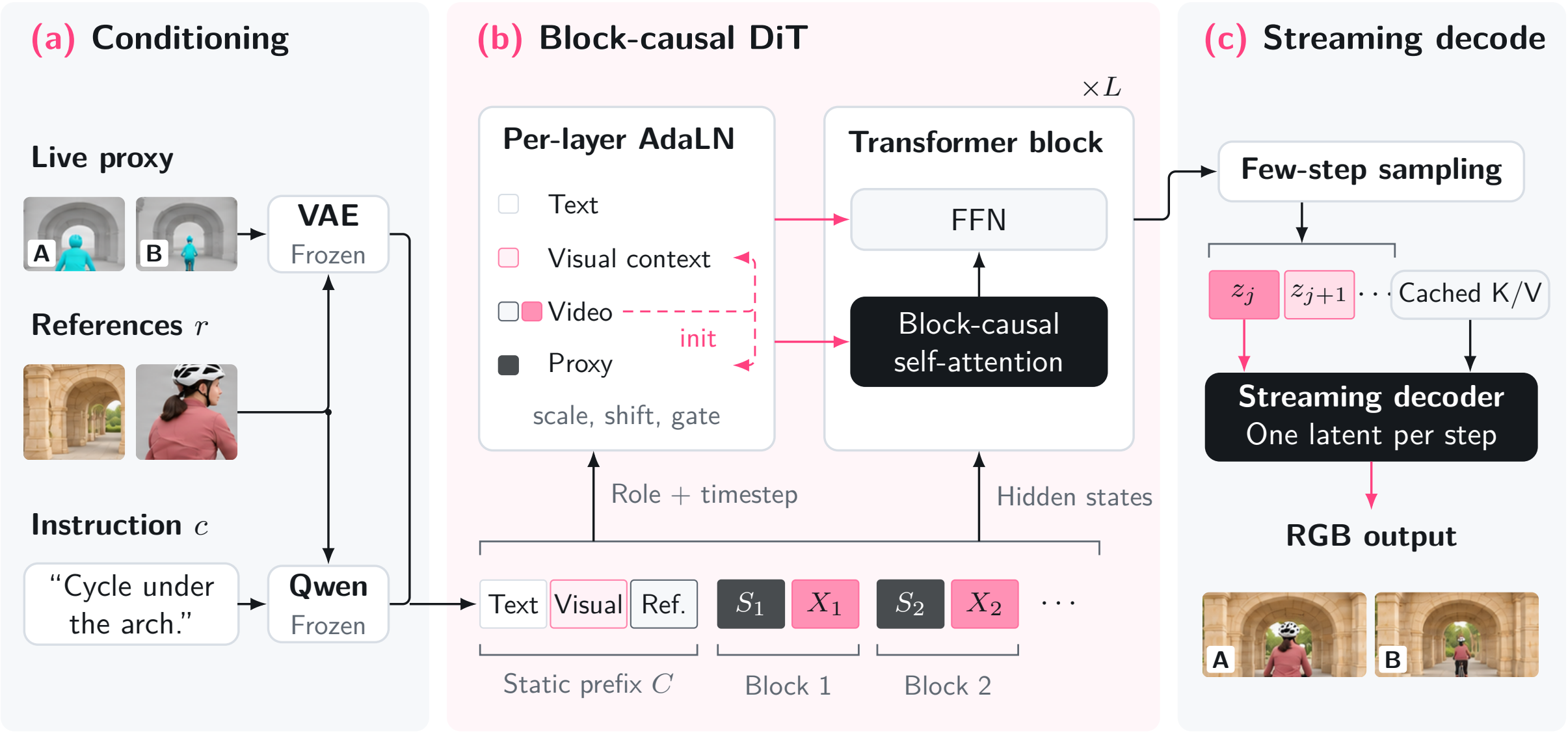}\par
  \caption{\textbf{Architecture of the Zing-1.0 streaming renderer.}
    (a) Frozen Qwen encodes the instruction and references, while the frozen visual VAE encodes the references and live proxy video.
    (b) The block-causal DiT processes interleaved proxy and target blocks after a static conditioning prefix. Role-specific AdaLN modulates both self-attention and the FFN. Dashed arrows indicate initialization from H3's video AdaLN branch.
    (c) The streaming decoder uses cached K/V to decode one latent at a time. Matching labels A and B identify corresponding proxy and rendered frames.}
  \label{fig:renderer-overview}
\end{figure}

We obtain the Zing-1.0 streaming renderer by post-training MiniMax-H3 (H3)~\citep{minimax2026h3}, starting from its H3-Base-Ref2VA checkpoint. The renderer comprises a block-causal diffusion transformer (DiT) and a streaming decoder. The DiT generates short video-latent blocks from the proxy stream, appearance references, text, and generated history. Within a fixed-condition segment, text and reference conditions are shared across blocks, while each new proxy block supplies updated scene structure and motion. The decoder emits RGB frames from each completed latent without waiting to decode the entire block. \figref{fig:renderer-overview} traces this path from input encoding to RGB output.

\Needspace{7\baselineskip}
\paragraph{Rendering formulation.}
Let $c$ be the instruction, $r$ the appearance references, $s_k$ the encoded proxy for block $k$, and $x_k$ the corresponding target video latents. Within a fixed-condition segment, we model
\begin{equation}
  p_\theta(x_{1:K}\mid c,r,s_{1:K})
  =\prod_{k=1}^{K}p_\theta(x_k\mid c,r,s_{\leq k},x_{<k}).
  \label{eq:renderer-factorization}
\end{equation}

Text specifies the rendering goal and links references to elements in the proxy. References provide scene and subject appearance independently of the proxy pose. The proxy specifies layout, camera motion, and visible object motion. No player command appears in this factorization: the renderer observes an action only through its executed result $s_k$, which is condition C2 of \secref{sec:cr-spec}.

\paragraph{Conditioning representations.}
The renderer combines static conditioning from the instruction and appearance references with dynamic conditioning from live proxy video. As in the original H3, Qwen jointly encodes the instruction $c$ and appearance references $r$ into semantic context, while the visual VAE preserves the appearance details of $r$ in reference latents. Together, these representations form the static prefix $C$ for each fixed-condition segment, and both encoders remain frozen in Zing-1.0. Qwen derives its visual context only from reference frames, without access to future proxy or target frames. For dynamic conditioning, Zing-1.0 adds a live proxy input, using the same visual VAE to encode each incoming proxy block as $s_k$ in Equation~\eqref{eq:renderer-factorization}. To align $s_k$ with the corresponding target block $x_k$, the two streams share block boundaries and temporal coordinates in H3's multimodal rotary position embedding (MM-RoPE), while separate spatial grids support different proxy and target resolutions.

\begin{figure}[!b]
  \centering
  \setlength{\parskip}{0pt}
  \captionsetup{skip=4pt}
  \input{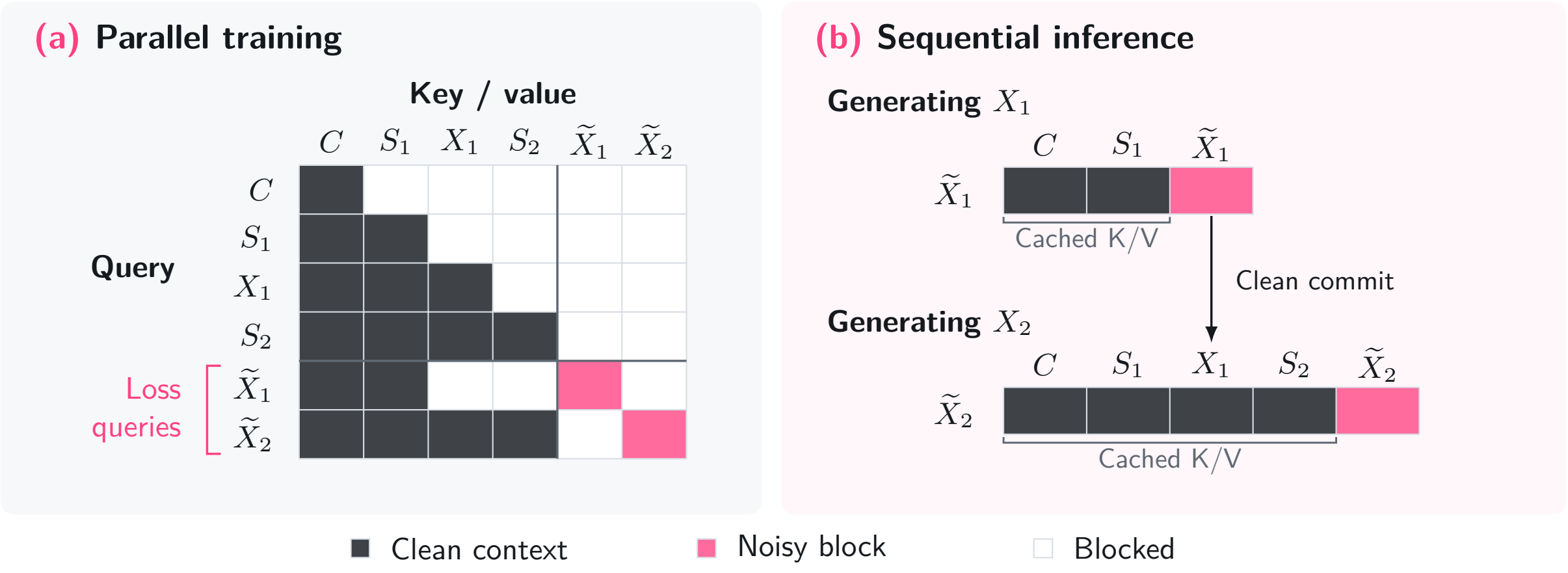}
\resizebox{\linewidth}{!}{%
\begin{tikzpicture}[x=1cm,y=1cm,font=\sffamily\fontsize{9}{11}\selectfont,text=zingInk,
  axis label/.style={font=\sffamily\bfseries\fontsize{9}{11}\selectfont},
  mask grid/.style={draw=zingRule,line width=.4pt},
  cache bracket/.style={draw=zingMuted,line width=.55pt}]
  \path[use as bounding box] (0,0) rectangle (16,5.82);
  \fill[zingSoft,rounded corners=4pt] (0,.58) rectangle (7.78,5.82);
  \fill[zingPinkWash!60,rounded corners=4pt] (7.98,.58) rectangle (16,5.82);
  \rpanel{a}{Parallel training}{.20}{5.43}
  \rpanel{b}{Sequential inference}{8.18}{5.43}

  \node[axis label] at (5.03,4.86) {Key / value};
  \node[axis label] at (1.35,3.09) {Query};
  \begin{scope}[shift={(3.05,1.15)},x=.66cm,y=.50cm]
    \fill[white] (0,0) rectangle (6,6);
    \foreach \r in {0,...,3}{
      \foreach \c in {0,...,\r}{\fill[zingInk!82] (\c,5-\r) rectangle ++(1,1);}
    }
    \foreach \c in {0,1}{\fill[zingInk!82] (\c,1) rectangle ++(1,1);}
    \foreach \c in {0,...,3}{\fill[zingInk!82] (\c,0) rectangle ++(1,1);}
    \fill[zingPink!78] (4,1) rectangle ++(1,1);
    \fill[zingPink!78] (5,0) rectangle ++(1,1);
    \draw[mask grid,step=1] (0,0) grid (6,6);
    \draw[zingMuted,line width=.65pt] (0,2)--(6,2) (4,0)--(4,6);
    \foreach \i/\key in {0/C,1/S_1,2/X_1,3/S_2,4/\widetilde X_1,5/\widetilde X_2}{
      \node at (\i+.5,6.48) {$\key$};
      \node[anchor=east] at (-.19,5.5-\i) {$\key$};
    }
  \end{scope}
  \node[anchor=east,align=right,text=zingPink] at (1.98,1.65) {Loss\\queries};
  \draw[zingPink,line width=.65pt] (2.25,2.10)--(2.11,2.10)--(2.11,1.20)--(2.25,1.20);

  \foreach \block/\cached/\y in {1/2/3.65,2/4/1.40}{
    \node[axis label,anchor=west] at (8.31,\y+1.13) {Generating $X_{\block}$};
    \begin{scope}[shift={(10.25,\y)},x=.85cm,y=.48cm]
      \foreach \c/\key in {0/C,1/S_1,2/X_1,3/S_2}{
        \ifnum\c<\cached\relax
          \fill[zingInk!82] (\c,0) rectangle ++(1,1);
          \node at (\c+.5,1.48) {$\key$};
        \fi
      }
      \fill[zingPink!78] (\cached,0) rectangle ++(1,1);
      \draw[mask grid,step=1] (0,0) grid (\cached+1,1);
      \node at (\cached+.5,1.48) {$\widetilde X_{\block}$};
      \node[anchor=east] at (-.19,.5) {$\widetilde X_{\block}$};
      \draw[cache bracket] (0,-.06)--(0,-.18)--(\cached,-.18)--(\cached,-.06);
      \node[font=\sffamily\fontsize{8}{10}\selectfont,text=zingMuted] at (\cached/2,-.56) {Cached K/V};
      \ifnum\block=1\relax
        \coordinate (noisy-first) at (2.5,-.04);
      \else
        \coordinate (cached-first) at (2.5,1.94);
      \fi
    \end{scope}
  }
  \draw[rarrow,line width=.65pt] (noisy-first)--node[rsmall,right=3pt] {Clean commit} (cached-first);

  \foreach \x/\swatch/\legendtext in {3.58/zingInk!82/Clean context,7.12/zingPink!78/Noisy block,10.56/white/Blocked}{
    \draw[mask grid,fill=\swatch] (\x,.14) rectangle ++(.19,.19);
    \node[anchor=west] at (\x+.27,.235) {\legendtext};
  }
\end{tikzpicture}%
}\par
  \caption{\textbf{Block-causal attention in training and inference.} Rows denote query blocks and columns denote K/V blocks. (a) Parallel teacher forcing evaluates multiple noisy target queries against clean context. Only the two bracketed target rows contribute to the loss in this example. (b) The DiT evaluates one noisy target query per denoising step. After denoising the first target, the DiT caches the K/V of clean $X_1$. The second target reads them alongside the static prefix $C$ and the proxy blocks $S_1$ and $S_2$.}
  \label{fig:renderer-causal}
\end{figure}

The original H3 DiT uses one video branch of adaptive layer normalization (AdaLN) for Qwen visual features and the visual latents of the references and target. With the addition of the live proxy, the renderer processes three types of visual tokens in the same sequence: (i) reference and target latents that encode the intended appearance and the video being generated, respectively; (ii) Qwen visual features that help interpret the references in the context of the instruction; and (iii) proxy latents that specify the layout and motion of the current block. As shown in \figref{fig:renderer-overview}, each layer retains H3's video AdaLN for reference and target latents and uses its weights to initialize separate branches for Qwen visual features and proxy latents. Qwen text retains its original branch. The token role and timestep determine the scale, shift, and gate for each sublayer, while the attention and feed-forward weights are shared across token roles.

\paragraph{Block-causal streaming.}
To implement Equation~\eqref{eq:renderer-factorization}, the DiT arranges its inputs in the sequence shown in \figref{fig:renderer-overview}. After the static prefix $C$, each proxy block $S_k$ precedes its aligned target block $X_k$. Attention is full within a block and causal across blocks. Training and inference share this rule but differ in how they obtain the target history.

\emph{Training.} Clean targets $X_j$ are available from the training video, so we can use parallel teacher forcing to predict multiple blocks without generating their history sequentially. As shown in \figpanelref{fig:renderer-causal}{a}, each noisy target query $\widetilde X_k$ attends only to itself, the static prefix $C$, and the visible blocks among the proxies $S_{\leq k}$ and clean targets $X_{<k}$. The loss is computed only for the noisy target queries during training.

\Needspace{5\baselineskip}
\emph{Inference.} No clean targets are supplied during inference, so the DiT must generate $X_{<k}$ before it can generate $X_k$. Once the incoming proxy block $S_k$ is processed, the DiT operates only on the current noisy block $\widetilde X_k$ at each denoising step while reading the other context blocks from the K/V cache, as shown in \figpanelref{fig:renderer-causal}{b}. After the final denoising step, the DiT runs another forward pass on the clean block $X_k$ and caches the resulting K/V for generating subsequent blocks.

\Needspace{6\baselineskip}
\paragraph{In-play prompt updates.}
The cached K/V for generated history depend on the conditioning prefix $C$. When an in-play event prompt updates the conditioning at a block boundary, the renderer retains recent generated latents as a visual prefix $P$ and recomputes their K/V under the updated conditioning prefix $C'$. Subsequent blocks are generated using the refreshed cache and the incoming proxy stream. To train the model to continue from a visual prefix, autoregressive (AR) adaptation keeps an observed target prefix fixed and computes the loss only on the remaining target latents.

\subsubsection{Training Data}
\label{sec:render-training-data}

\begingroup
\setlength{\intextsep}{4pt}
\begin{figure}[H]
  \centering
  \captionsetup{skip=4pt}
  \tikz[baseline=(data.base)]{
    \node[fill=zingSoft!85,rounded corners=4pt,minimum width=\linewidth,inner sep=0pt,outer sep=0pt] (data)
      {\includegraphics[width=0.84\linewidth]{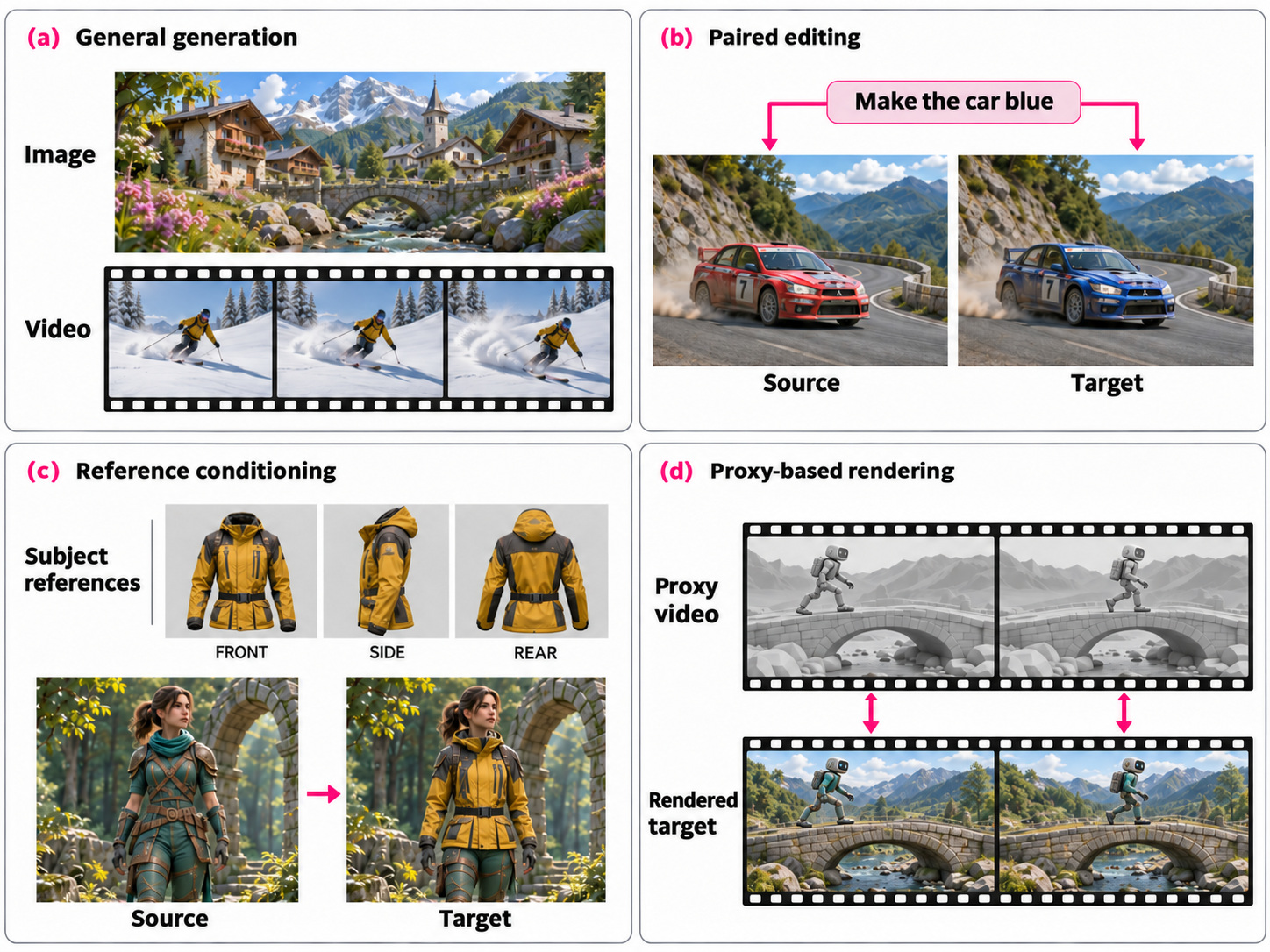}};
  }
  \caption{Conceptual examples of the four training data categories. (a) Standalone image and video generation. (b) Paired editing from a source and a text instruction. (c) Subject appearance specified by multi-view reference images. (d) Temporally aligned proxy and target video frames.}
  \label{fig:distinctive-training-data}
\end{figure}
\endgroup

To synthesize coherent video that follows the proxy's structure and motion while matching the specified appearance, we train the renderer with four complementary forms of supervision. General generation data supports the model's visual and motion capabilities, paired editing data teaches it to transform a source while retaining content outside the requested edit, and reference-conditioned examples teach it to use images as appearance controls. Training examples for proxy-based rendering combine these forms of supervision under the input conditions used during play. \figref{fig:distinctive-training-data} illustrates selected examples of the data design.

\paragraph{General generation.}
To preserve the general generation capabilities of the base MiniMax-H3 model during adaptation to proxy-conditioned rendering, we include image and video generation data covering a range of appearances, compositions, and motion. The image mixture includes quality-filtered images, game artwork, synthetic examples for text rendering and facial detail, and images synthesized using compositions from selected open-source examples. Video sources include open-source footage, film and television, game cinematics, and sports clips that cover large and rapid motion. The mixture includes text-conditioned audiovisual and image-conditioned video generation. Selected single-shot clips retain audio, and multi-shot clips are segmented using audio and subtitles to reduce cuts in speech and other continuous audio.

\paragraph{Paired editing.}
Paired editing examples consist of a source image or video, an edit instruction, and the corresponding target. The target provides supervision both for applying the requested change and for preserving source content outside the edit. We filter open-source image and video editing examples for visual quality and edit correctness, and reannotate video editing examples in the MiniMax-H3 format.

\begin{figure}[!b]
  \centering
  \includegraphics[width=0.94\linewidth]{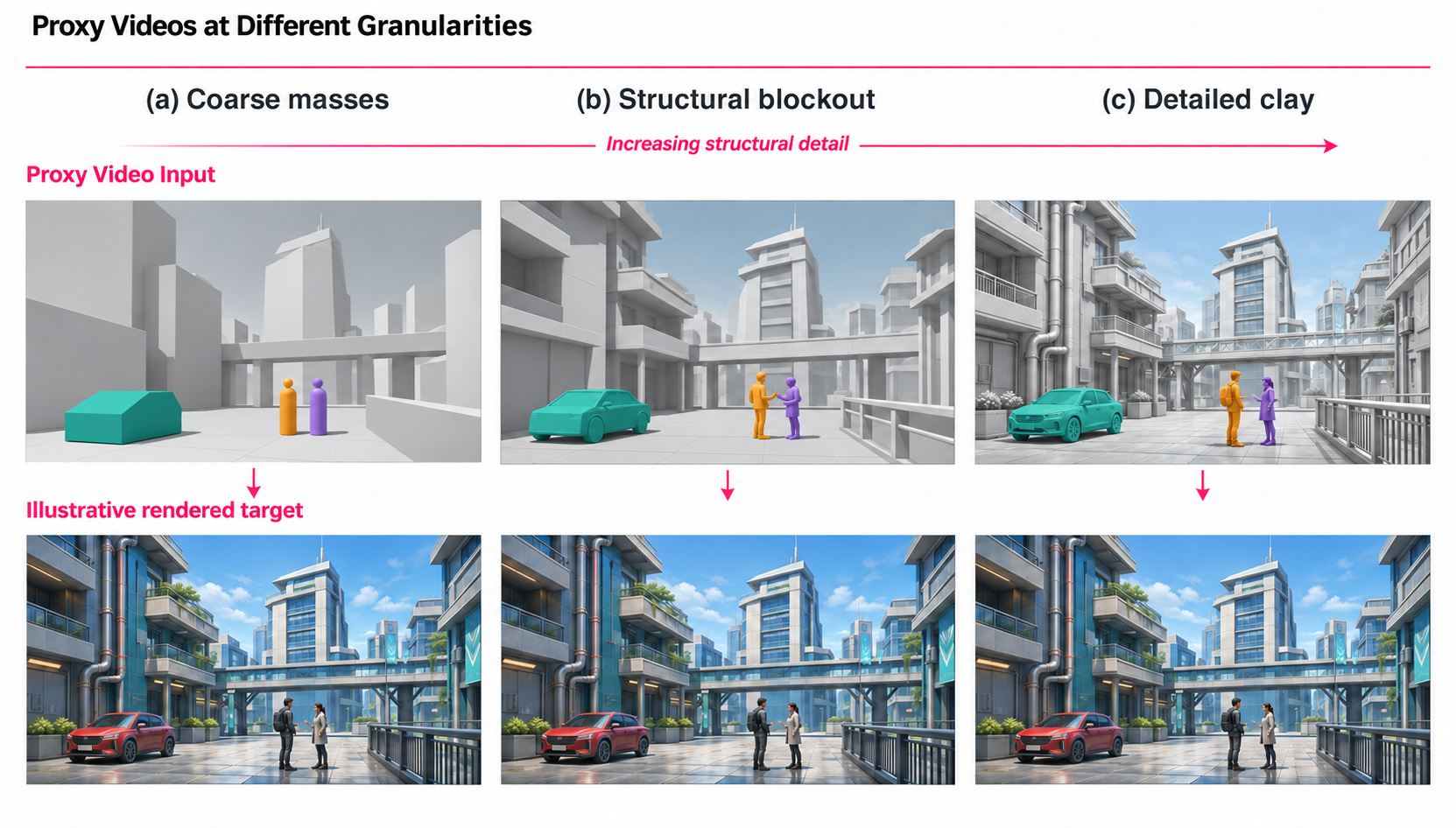}
  \caption{Proxy videos at different granularities: (a) coarse masses, (b) structural blockout, and (c) detailed clay geometry. Consistent proxy colors distinguish the vehicle and the two people across all three inputs, which share the same illustrative rendered target.}
  \label{fig:proxy-granularity}
\end{figure}

\paragraph{Reference conditioning.}
Reference-conditioned generation and editing data provide supervision for reproducing scene or subject appearance from one or more reference images. In editing, text identifies the element to modify, the reference specifies its appearance, and the source constrains what remains unchanged. Some image and video pairs provide three views of the same subject (\figpanelref{fig:distinctive-training-data}{c}), supplying appearance cues across viewpoints without fixing the output pose or position.

\paragraph{Proxy-based rendering.}
Proxy--target pairs combine the preceding forms of supervision to align training with the rendering task during play. The goal is coherent video that follows proxy structure and motion while matching the specified appearance. We construct proxy--target pairs from game imagery, film and television, gameplay recordings, and targeted animal and vehicle footage. For images, we abstract appearance while retaining composition and use the original image as the target. For video, the proxy and target are aligned in time (\figpanelref{fig:distinctive-training-data}{d}) to supervise the preservation of camera and subject motion and the timing of occlusions and visibility changes. Some pairs use proxy video produced by WorldKit, matching the type of input used during play. Proxy inputs range from coarse masses through structural blockouts to detailed clay representations (\figref{fig:proxy-granularity}).

\subsubsection{Training and Distillation}
\label{sec:render-training}

\begin{figure}[!ht]
  \centering
  \setlength{\parskip}{0pt}
  \captionsetup{skip=4pt}
  \input{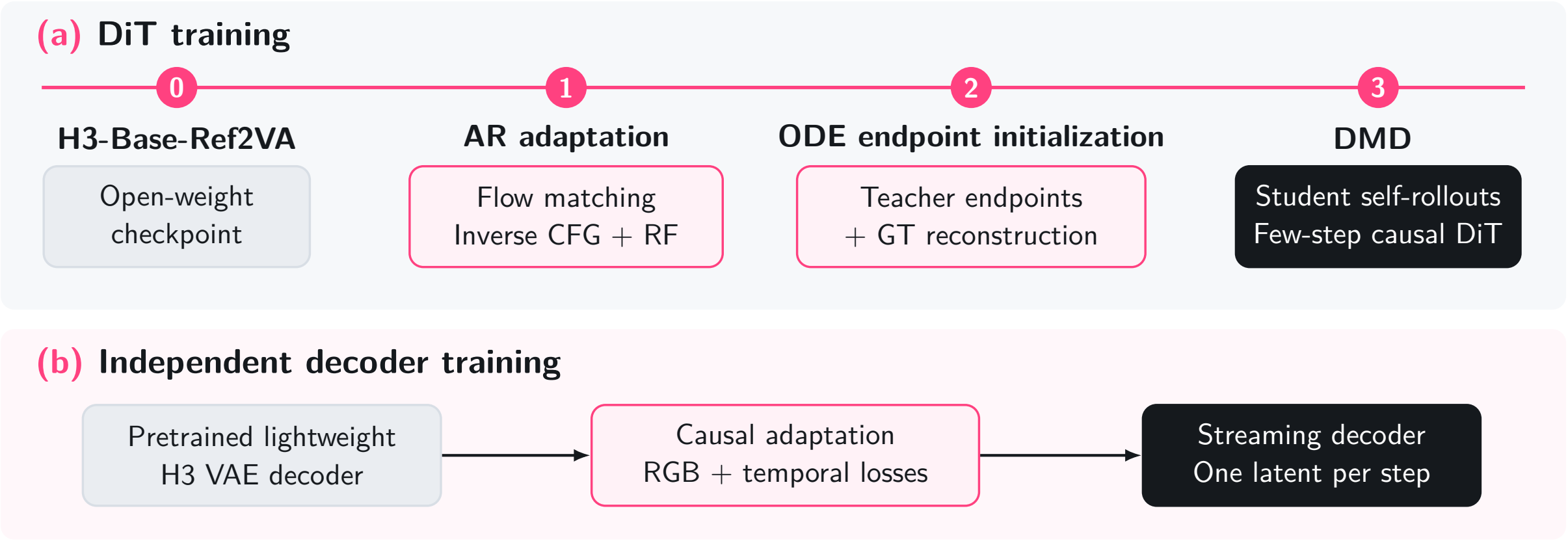}
\resizebox{\linewidth}{!}{%
\begin{tikzpicture}[x=1cm,y=1cm,text=zingInk,
  stage title/.style={font=\sffamily\bfseries\fontsize{9}{11}\selectfont,align=center},
  stage box/.style={font=\sffamily\fontsize{8.5}{11}\selectfont,minimum height=1.03cm}]
  \path[use as bounding box] (0,0) rectangle (16,5.50);
  \fill[zingSoft,rounded corners=4pt] (0,2.35) rectangle (16,5.50);
  \fill[zingPinkWash!60,rounded corners=4pt] (0,0) rectangle (16,2.15);
  \rpanel{a}{DiT training}{.22}{5.14}
  \rpanel{b}{Independent decoder training}{.22}{1.79}

  \draw[zingPink,line width=1pt] (.42,4.62)--(15.58,4.62);
  \foreach \stage/\x/\title in {
    0/1.80/H3-Base-Ref2VA,
    1/5.78/AR adaptation,
    2/9.92/ODE endpoint initialization,
    3/14.08/DMD
  }{
    \node[rbadge] at (\x,4.62) {\stage};
    \node[stage title] at (\x,4.10) {\title};
  }
  \node[rstate,fill=zingRule!55,stage box,minimum width=2.72cm] at (1.80,3.30) {Open-weight\\checkpoint};
  \node[rpink,stage box,minimum width=3.20cm] at (5.78,3.30) {Flow matching\\Inverse CFG + RF};
  \node[rpink,stage box,minimum width=3.56cm] at (9.92,3.30) {Teacher endpoints\\+ GT reconstruction};
  \node[rdark,stage box,minimum width=2.88cm] at (14.08,3.30) {Student self-rollouts\\Few-step causal DiT};

  \node[rstate,fill=zingRule!55,stage box,minimum width=3.66cm] (light) at (2.67,.86)
    {Pretrained lightweight\\H3 VAE decoder};
  \node[rpink,stage box,minimum width=3.96cm] (adapt) at (8.02,.86)
    {Causal adaptation\\RGB + temporal losses};
  \node[rdark,stage box,minimum width=3.45cm] (dec) at (13.40,.86)
    {Streaming decoder\\One latent per step};
  \draw[rarrow] (light.east)--(adapt.west);
  \draw[rarrow] (adapt.east)--(dec.west);
\end{tikzpicture}%
}\par
  \caption{\textbf{Independent training paths for the DiT and decoder.} (a) The DiT starts from H3-Base-Ref2VA and is trained in three successive stages. In autoregressive adaptation, RF changes the history seen by the model without changing the supervised target. Endpoint initialization prepares the student for larger denoising updates, and DMD refines few-step outputs using student rollouts. (b) The decoder is adapted independently on paired latents and RGB frames with reconstruction and temporal losses.}
  \label{fig:renderer-training}
\end{figure}

To support low-latency streaming during play, we train a block-causal generator that produces each video block in a few denoising steps. Starting from the same H3 checkpoint, training proceeds through AR adaptation, ODE endpoint initialization, and distribution matching distillation (DMD). \figref{fig:renderer-training} summarizes these DiT training stages and the separate adaptation of the decoder to process one latent frame at a time.

\paragraph{Autoregressive adaptation.}
This stage adapts the DiT to generate one target block at a time during inference. Training and inference use the same block layout, alternating between full blocks of three and two latent frames ($3,2,3,2,\ldots$) to align each block pair with one five-latent temporal cycle of the original H3. The flow-matching objective~\citep{flowmatching} follows H3's convention that the velocity points from noise toward data. Given clean latents $x_0$, Gaussian noise $\epsilon$, and noise fraction $\sigma$, the noisy latents, velocity target, and basic regression objective are
\begin{equation}
  x_\sigma=(1-\sigma)x_0+\sigma\epsilon,\qquad
  u^\star=x_0-\epsilon,\qquad
  \mathcal{L}_{\mathrm{FM}}=\mathbb{E}\!\left[\lVert\hat u_\theta-u^\star\rVert_2^2\right].
  \label{eq:h3-flow}
\end{equation}
Here $\hat u_\theta$ is the velocity prediction used for regression. The model timestep is $t=1-\sigma$, with $t=1$ denoting clean media. Different target blocks are trained at independently sampled noise levels. The flow regression loss is computed only on noisy target queries, as shown in \figpanelref{fig:renderer-causal}{a}.

\begin{figure}[!ht]
  \centering
  \input{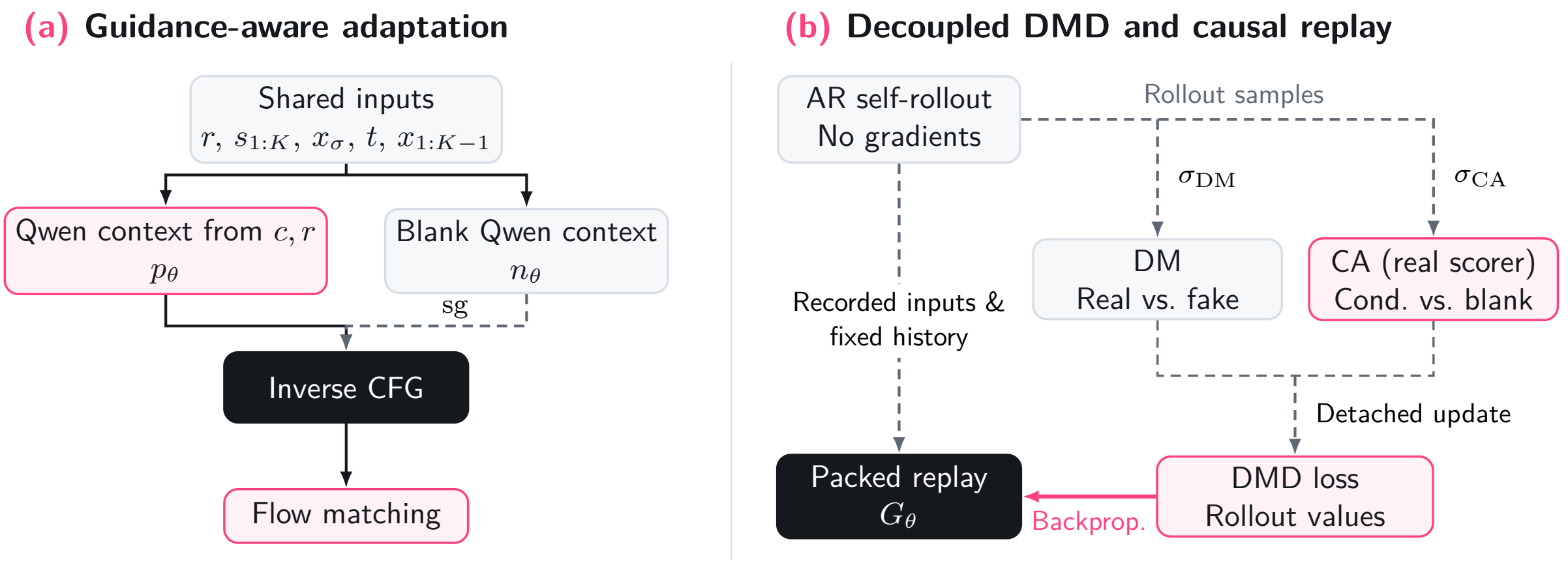}
\resizebox{\linewidth}{!}{%
\begin{tikzpicture}[x=1cm,y=1cm,font=\sffamily\fontsize{9}{11}\selectfont]
  \path[use as bounding box] (-.1,0) rectangle (16,5.95);
  \rpanel{a}{Guidance-aware adaptation}{0}{5.6}
  \rpanel{b}{Decoupled DMD and causal replay}{7.8}{5.6}
  \draw[zingRule] (7.4,.15)--(7.4,5.25);
  \node[rstate,minimum width=2.9cm,minimum height=.72cm,inner sep=3pt] (input) at (3.45,4.67)
    {Shared inputs\\$r,\,s_{1:K},\,x_\sigma,\,t,\,x_{1:K-1}$};
  \node[rpink,minimum width=2.75cm,minimum height=.81cm,inner sep=3pt] (cond) at (1.6,3.32) {Qwen context from $c,r$\\$p_\theta$};
  \node[rstate,minimum width=2.75cm,minimum height=.81cm,inner sep=3pt] (blank) at (5.3,3.32) {Blank Qwen context\\$n_\theta$};
  \draw[rarrow] (input.south)--(3.45,4.1)-|(cond.north);
  \draw[rarrow] (3.45,4.1)-|(blank.north);
  \node[rdark,minimum width=2.5cm,minimum height=.72cm,inner sep=3pt] (inverse) at (3.45,1.92) {Inverse CFG};
  \draw[rarrow] (cond.south)--(1.6,2.55)-|(inverse.north);
  \draw[rguide] (blank.south)--(5.3,2.55)-|(inverse.north);
  \node[rsmall,anchor=south,inner sep=2pt] at (4.57,2.55) {$\operatorname{sg}$};
  \node[rpink,minimum width=2.5cm,minimum height=.55cm,inner sep=3pt] (fm) at (3.45,.6) {Flow matching};
  \draw[rarrow] (inverse)--(fm);

  \node[rstate,minimum width=2.48cm,minimum height=.81cm,inner sep=3pt] (rollout) at (9.12,4.67) {AR self-rollout\\No gradients};
  \node[rstate,minimum width=2.55cm,minimum height=.81cm,inner sep=3pt] (dm) at (11.77,3.03) {DM\\Real vs.\ fake};
  \node[rpink,minimum width=2.55cm,minimum height=.81cm,inner sep=3pt] (ca) at (14.60,3.03) {CA (real scorer)\\Cond.\ vs.\ blank};
  \node[rlabel] at (12.55,4.91) {Rollout samples};
  \draw[rguide] (rollout.east)--(14.60,4.67)--node[rsmall,right=2pt] {$\sigma_{\mathrm{CA}}$}(ca.north);
  \draw[rguide] (11.77,4.67)--node[rsmall,right=2pt] {$\sigma_{\mathrm{DM}}$}(dm.north);
  \node[rdark,minimum width=2.50cm,minimum height=.79cm,inner sep=3pt] (replay) at (9.12,.80) {Packed replay\\$G_\theta$};
  \draw[rguide] (rollout.south)--(replay.north);
  \node[rsmall,fill=white,inner sep=2pt] at (9.12,2.60) {Recorded inputs \&\\fixed history};

  \node[rpink,minimum width=2.83cm,minimum height=.79cm,inner sep=3pt] (loss) at (13.18,.80) {DMD loss\\Rollout values};
  \draw[draw=zingMuted,densely dashed,line width=.65pt]
    (dm.south)--(11.77,2.04)--(14.60,2.04)--(ca.south);
  \draw[rguide] (13.18,2.04)--node[rsmall,right=2pt] {Detached update}(loss.north);
  \draw[rgrad] (loss.west)--node[rsmall,below,text=zingPink] {Backprop.}(replay.east);
\end{tikzpicture}%
}
  \caption{\textbf{Guidance-aware adaptation and DMD gradients.} (a) Both branches share the VAE encoding of $r$, proxy latents $s_{1:K}$, noisy targets $x_\sigma$, timesteps $t$, and target history $x_{1:K-1}$. Each block $k$ can attend only to $x_{<k}$ in this history. The blank branch replaces only the Qwen context with a blank-prompt embedding, and $\operatorname{sg}$ stops gradients through its prediction $n_\theta$. (b) DM compares the conditional real and fake scores, whereas CA compares conditional and blank real scores at a separately sampled noise level. The DMD loss uses rollout values, but gradients pass through the packed replay with generated history latents held fixed.}
  \label{fig:renderer-guidance}
\end{figure}

\begin{itemize}[label={},leftmargin=0.75em,labelsep=0pt,topsep=0pt,partopsep=0pt,itemsep=0pt,parsep=\parskip]
\item \textbf{\emph{Guidance-aware adaptation.}} The released H3 checkpoint~\citep{minimax2026h3} is CFG-distilled, so we use inverse classifier-free guidance (CFG)~\citep{cfg} regression, related to the H3 guidance loss in Musubi Tuner~\citep{musubiH3Guidance}, to retain the learned guidance during autoregressive adaptation. As shown in \figpanelref{fig:renderer-guidance}{a}, the conditional branch uses all inputs, while the unconditional branch replaces only the Qwen context with a blank-prompt embedding. For the conditional prediction $p_\theta$ and unconditional prediction $n_\theta$, inverse CFG gives the corrected velocity
\begin{equation}
  \hat u_\theta = \frac{p_\theta + (s-1)\operatorname{sg}(n_\theta)}{s},
  \label{eq:inverse-cfg}
\end{equation}
where $s$ is the guidance scale and $\operatorname{sg}$ denotes stop-gradient. We regress $\hat u_\theta$ against the flow-matching target $u^\star$ in Equation~\eqref{eq:h3-flow}.

\item \textbf{\emph{Resampling forcing.}} Teacher forcing uses ground-truth history, whereas at inference the model conditions on its own past predictions. Resampling Forcing (RF)~\citep{resamplingforcing} adds noise to selected ground-truth history blocks, then uses the current model without gradient tracking to denoise each block in one step, with earlier blocks resampled in the same rollout as history (\figpanelref{fig:renderer-training}{a}). This exposes the model to its own prediction errors without requiring a full denoising schedule for each history block.
\end{itemize}

\Needspace{7\baselineskip}
\paragraph{ODE endpoint initialization.}
After autoregressive adaptation, the model still relies on many-step sampling to generate each block. Reducing this to a few steps requires larger updates, over which the velocity can change along the denoising trajectory. Flow matching supervises the instantaneous velocity at each sampled noise level rather than the total change over an interval, so extrapolating a single prediction over a large interval can produce an inaccurate endpoint.

Following the mid-training principle of CMT~\citep{cmt}, we train the student to predict the clean endpoint of a frozen causal AR teacher's multi-step denoising trajectory directly from an intermediate state, with both models conditioned on the same ground-truth history. In addition, we apply an auxiliary reconstruction loss with the original clean video latents as targets to avoid relying solely on teacher-generated endpoints.

\paragraph{Distribution matching distillation.}
After ODE endpoint initialization, the student is trained with distribution matching distillation (DMD)~\citep{dmd,dmd2} to improve the quality of few-step generation by matching the output distribution of a frozen multi-step teacher. A frozen real scorer and a separately trained fake scorer estimate the scores of the teacher and student distributions, respectively. Following Self Forcing~\citep{huang2025selfforcing}, we generate samples via autoregressive self-rollout of the current student and use them to train the fake scorer with flow matching.

\begin{itemize}[label={},leftmargin=0.75em,labelsep=0pt,topsep=0pt,partopsep=0pt,itemsep=0pt,parsep=\parskip]
\item \textbf{\emph{Decoupled DMD.}} In practice, DMD often applies CFG to the teacher score but not to the student score, combining distribution matching and guidance in a single update. Decoupled DMD~\citep{decoupleddmd} decomposes this update into distribution matching (DM) and CFG augmentation (CA); we evaluate the two terms at independently sampled noise levels. For each term, the teacher's conditional score is recovered from the real scorer's guided and blank-context predictions using the inverse CFG transformation in Equation~\eqref{eq:inverse-cfg}. DM aligns the student and teacher distributions by comparing their conditional scores, whereas CA strengthens adherence to the conditioning inputs by comparing the teacher's conditional and blank-context scores.

\item \textbf{\emph{Data-Forcing Distillation.}} To preserve sample diversity and mitigate oversaturation during few-step distillation, we apply Data-Forcing Distillation~\citep{dfd} to a subset of samples by evaluating the real scorer on noised ground-truth latents instead of noised student outputs.

\item \textbf{\emph{Self Gradient Forcing.}} When historical K/V is detached during self-rollout training, gradients from later blocks cannot reach the computation that encodes earlier blocks into the cache. Self Gradient Forcing~\citep{sgf} restores this gradient path with a two-pass procedure (\figpanelref{fig:renderer-guidance}{b}). First, an autoregressive rollout without gradient tracking records each block's noisy input, timestep, and clean prediction at the selected denoising step. Second, a packed causal replay recomputes the predictions and history K/V, treating the generated history latents as fixed inputs. DMD gradients computed on the original rollout predictions are then backpropagated through the replay.
\end{itemize}

\subsubsection{Streaming Decoder}
\label{sec:render-decoder}

Jointly decoding a DiT block preserves temporal context but postpones the first RGB output until the block has been decoded. Decoding each latent independently allows earlier output but discards the history needed for coherent transitions. To reduce first-frame latency without discarding temporal context, we adapt the pretrained lightweight H3 VAE decoder from LynnReal-Omni~\citep{mao2026lynnrealomni} for causal decoding. The adapted decoder emits RGB frames one latent at a time using the current latent and cached history.

\begin{figure}[!ht]
  \centering
  \input{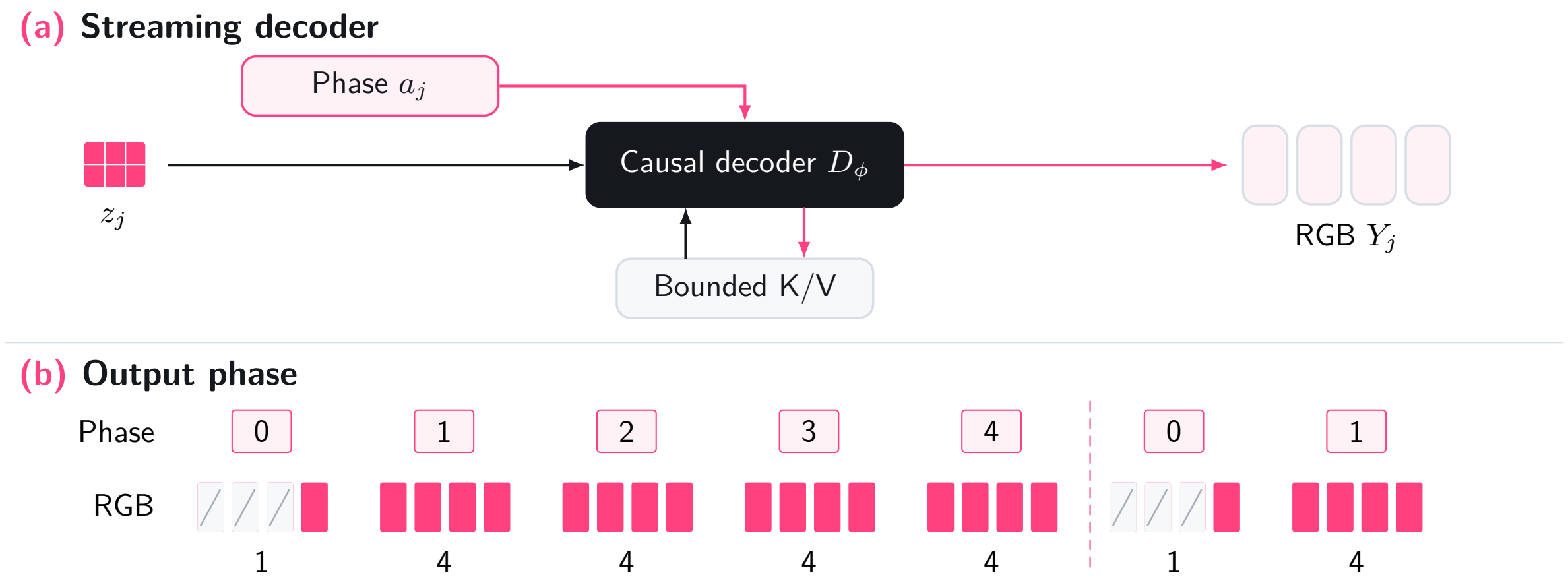}
\resizebox{\linewidth}{!}{%
\begin{tikzpicture}[x=1cm,y=1cm,font=\sffamily\fontsize{9}{11}\selectfont]
  \path[use as bounding box] (-.1,-.1) rectangle (16,5.8);
  \rpanel{a}{Streaming decoder}{0}{5.48}
  \rtile{.8}{3.88}{zingPink}
  \node at (1.1,3.55) {$z_j$};
  \node[rdark,text width=3.0cm,minimum height=.85cm,inner sep=3pt] (decoder) at (7.5,4.1) {Causal decoder $D_\phi$};
  \draw[rarrow] (1.65,4.1)--(decoder.west);
  \node[rpink,minimum width=2.6cm,minimum height=.6cm,inner sep=3pt] (phase) at (3.7,4.9) {Phase $a_j$};
  \draw[rpinkarrow] (phase.east)--(7.5,4.9)--(decoder.north);
  \node[rstate,minimum width=2.6cm,minimum height=.6cm,inner sep=3pt] (kv) at (7.5,2.85) {Bounded K/V};
  \draw[rarrow] (6.9,3.15)--(6.9,3.67);
  \draw[rpinkarrow] (8.1,3.67)--(8.1,3.15);
  \draw[rpinkarrow] (decoder.east)--(12.4,4.1);
  \foreach \i in {0,1,2,3}{\draw[rframe,fill=zingPinkWash] (12.55+.55*\i,3.7) rectangle ++(.45,.8);}
  \node at (13.6,3.36) {RGB $Y_j$};
  \draw[zingRule] (0,2.3)--(15.8,2.3);
  \rpanel{b}{Output phase}{0}{1.96}
  \node[anchor=east] at (1.65,1.4) {Phase};
  \node[anchor=east] at (1.65,.65) {RGB};
  \foreach \i/\ph/\count in {0/0/1,1/1/4,2/2/4,3/3/4,4/4/4,5/0/1,6/1/4}{
    \pgfmathsetmacro{\xx}{2.6+1.85*\i}
    \node[rtoken,fill=zingPinkWash,draw=zingPink,minimum width=.6cm] at (\xx,1.4) {\ph};
    \foreach \slot in {0,1,2,3}{\fill[zingPink,rounded corners=.7pt] (\xx-.65+.35*\slot,.38) rectangle ++(.27,.5);}
    \ifnum\ph=0
      \foreach \slot in {0,1,2}{
        \fill[zingSoft,rounded corners=.7pt] (\xx-.65+.35*\slot,.38) rectangle ++(.27,.5);
        \draw[renderMid,line width=.5pt] (\xx-.62+.35*\slot,.44)--++(.2,.37);
      }
    \fi
    \node at (\xx,.08) {\count};
  }
  \draw[zingPink,densely dashed] (11,.02)--(11,1.72);
\end{tikzpicture}%
}
  \caption{\textbf{Single-latent streaming decoder.} (a) A phase embedding and bounded historical K/V support incremental RGB decoding. (b) H3's inherited phase cycle yields one RGB frame at phase zero and four at each other phase.}
  \label{fig:renderer-decoder}
\end{figure}

\Needspace{5\baselineskip}
\paragraph{Phase embedding.}
The number of RGB frames emitted per decoding step depends on the latent's position in H3's temporal cycle. Each cycle maps five latents to 17 RGB frames in the pattern $1,4,4,4,4$, as shown in \figpanelref{fig:renderer-decoder}{b}. The phase $a_j=j\bmod 5$ uses the zero-based latent index $j$ across the full stream, rather than an index reset for each DiT block. A learned phase embedding conditions the decoder on $a_j$ to account for this periodic pattern in RGB frame counts.

\paragraph{Causal decoding with bounded history.}
As in the original H3 VAE decoder, decoding uses overlapping spatial tiles whose outputs are blended to form full frames. Within each tile, the decoder attends to all positions in the current latent and to the cached K/V $q_{j-1}$ from a fixed window of preceding latents. Each step of our causal decoder $D_\phi$ processes a single latent $z_j$, emits RGB frames $Y_j$ according to the output phase $a_j$, and updates the cache to $q_j$ (\figpanelref{fig:renderer-decoder}{a}):
\begin{equation}
  (Y_j,q_j)=D_\phi(z_j,q_{j-1},a_j).
  \label{eq:renderer-decoder-step}
\end{equation}

\paragraph{Decoder training.}
As shown in \figpanelref{fig:renderer-training}{b}, the decoder is trained separately from the DiT to reconstruct the original RGB frames from latents produced by the frozen H3 encoder. To process latent steps in parallel during training, we use a causal attention mask that reproduces the history window used at inference and prevents attention across samples and spatial tiles. For a clip of $T$ frames, the training objective combines a Charbonnier loss~\citep{charbonnier1997deterministic} for RGB reconstruction with temporal difference matching:
\begin{equation}
  \mathcal{L}_{\mathrm{dec}} = \frac{1}{N}\left[
  \sum_{t=1}^{T}\left\|\sqrt{(\hat{y}_t-y_t)^2+\epsilon^2}\right\|_1
  + \lambda_{\mathrm{temp}}\sum_{t=2}^{T}
  \left\| (\hat{y}_t-\hat{y}_{t-1}) - (y_t-y_{t-1}) \right\|_1
  \right].
  \label{eq:renderer-decoder-loss}
\end{equation}
Here $\hat{y}_t$ and $y_t$ are the predicted and target RGB frames, respectively, and $N$ is the total number of RGB values in the clip. The parameter $\epsilon>0$ is the Charbonnier smoothing constant, and $\lambda_{\mathrm{temp}}$ is the coefficient of the temporal difference loss.

\section{Interaction Infrastructure}
\label{sec:streaming-infrastructure}

The \emph{Interaction Runtime} integrates the streaming renderer in \secref{sec:streaming-renderer} with world execution and output delivery to provide timely visual feedback as the world advances and rendering conditions change. Within this runtime, Zing-1.0 builds its model inference infrastructure on SGLang~\citep{zheng2024sglang} and optimizes generation and condition updates for low-latency streaming.

\subsection{Streaming Runtime}
\label{sec:infra-cache}
\label{sec:infra-pipeline}

World execution, video generation, and delivery need not advance in lockstep.
The streaming runtime coordinates these stages by bounding queued work and tracking the proxy frames and rendering conditions used to generate each output batch.

\paragraph{Source admission and flow control.}
As described in \secref{sec:world-execution}, WorldKit executes gameplay commands and supplies proxy frames sampled from the evolving world state. Text and reference updates reach the renderer as separate conditions. The rendering service accepts ordered batches of RGB proxy frames under a credit-based protocol that limits the number of outstanding frames and bytes. Receipt acknowledgements confirm acceptance of submitted frames, but only consumption acknowledgements restore the corresponding credits. This limits input buildup when proxy frames arrive faster than the renderer can process them.

\paragraph{Pipelined generation and session state.}
While the DiT generates the current target block, the Encoder prepares inputs for subsequent blocks and the streaming Decoder converts completed latents into RGB frames. Bounded buffers limit queued work between stages. The evaluated serving group admits one active session at a time. Within a session, window attention and designated sink blocks bound the retained history. A preallocated KV pool holds the conditioning prefix, retained window, and temporary working rows. The pool is updated in place, and fixed-address metadata supports CUDA Graph replay as the window advances. Condition-encoding and Block0 caches reuse initial computations when inputs and model configuration match. Stage overlap improves sustained throughput. The joint execution scheme in \secref{sec:infra-gpu} further reduces backbone work per block, while the sparse attention and compressed KV techniques in \secref{sec:infra-sparse-kv} reduce the cost of reading and storing history.

\paragraph{Output delivery and provenance.}
Session and generation identifiers distinguish streams, and each output batch identifies its aligned proxy frame spans and the condition event used to generate it. The proxy frame tags relate those spans to the sampled world updates. Generating frames under an updated condition requires the state refresh described in \secref{sec:infra-prompt-switch}, so acknowledging the update is distinct from delivering those frames. Decoded RGB frames are published at the serving gateway, and the resulting video stream is delivered to the browser for playback.

\subsection{Cross-Block Joint Execution}
\label{sec:infra-gpu}

As described in \secref{sec:render-model}, an additional DiT forward pass computes the K/V of the clean target block $X_{k-1}$ after denoising. The decoder consumes the clean latents directly, so this pass can be deferred without delaying decoding. Once the next proxy block $S_k$ is available, a single DiT forward computes the K/V for $X_{k-1}$, processes $S_k$, and performs the first denoising step on $\widetilde X_k$. The joint forward batches the three branches for backbone projections and feed-forward layers while retaining branch-specific modulation and attention visibility.

The joint forward preserves the causal dependencies among these branches. The previous clean target does not attend to the current block, and the current proxy does not attend to the noisy target. The noisy target can attend to both, but its K/V are not retained as history. With two denoising steps, a baseline that already combines proxy processing with the first step still needs separate passes for the second step and the clean target. Cross-block joint execution reduces the steady-state schedule from three to two DiT forwards per eligible block without changing the number of denoising steps. Stream boundaries are handled separately, and the runtime falls back to separate passes when joint execution is not applicable.

\subsection{Frame-Aligned Sparse Attention and Compressed KV}
\label{sec:infra-sparse-kv}

Joint execution in \secref{sec:infra-gpu} reduces the number of DiT calls without changing the context visible to each denoising step. Even with the bounded window in \secref{sec:infra-pipeline}, the retained history occupies GPU memory and is read repeatedly during denoising. We compress the keys and values when they enter the cache and use the same cached representation to attend to different parts of the history at different resolutions.

\paragraph{Frame-aligned KV compression.}
We partition each latent frame into near-square spatial tiles of at most 64 tokens. Keeping a tile within one frame ensures that its tokens share a temporal RoPE position, and keeping it spatially compact limits the variation a pooled summary must represent. Within each tile, the pooling layout groups tokens into four spatial quadrants on a $2\times2$ grid. When writing a tile to the cache, we store its key and value means in BF16 and quantize the centered token-level residuals to FP8. Subtracting the mean allows the residual quantizer to represent differences among tokens without allocating range to their shared offset. We also store FP8 means for the tile and for each quadrant, together with quantization scales and routing statistics, so these quantities can be reused across denoising steps.

For a tile before any window-induced position shift, its unscaled token-level attention logit is approximated by
\begin{equation}
  \langle\mathbf{q}_i,\mathbf{k}_{m,j}\rangle
  \approx \langle\mathbf{q}_i,\bar{\mathbf{k}}_m\rangle
  + s_q s_{k,m}\langle\mathbf{q}_i^{\mathrm{FP8}},\Delta\mathbf{k}_{m,j}^{\mathrm{FP8}}\rangle,
  \label{eq:refine-dc-res}
\end{equation}
where $\mathbf{q}_i$ is the query at token $i$, $\mathbf{k}_{m,j}$ is the key at token $j$ in tile $m$, and $\bar{\mathbf{k}}_m$ is the stored BF16 key mean. The FP8 query and key residual have scales $s_q$ and $s_{k,m}$, respectively. For a given query, the first term is shared by every token in the tile, while the second recovers token-specific differences. Values use the same mean--residual representation with their own scale.

\paragraph{Query-adaptive three-tier attention.}
We retain all visible tokens in the current block and apply three-tier routing only to eligible historical tiles. For the retained history, \texttt{Refine} reads the token-level representation, \texttt{Skim} uses four quadrant means, and \texttt{Skip} uses the whole-tile mean. The router uses both the relevance of a tile to the current query and the spatial distribution of its key variance. A tile whose variation lies mainly between quadrants can retain that structure through four means, whereas substantial variation within quadrants makes token-level attention more useful.

For each query tile and attention head, let $\mathcal{H}$ contain all history tiles eligible for routing, and let $\mathcal{C}$ be the relevance-ranked list of candidates drawn from $\mathcal{H}$. The router combines relevance and cached tile statistics into a base utility and weights it by the variance distribution to obtain the promotion utilities $u_m^{\mathrm{refine}}$ and $u_m^{\mathrm{skim}}$ for candidate $m$. In the budget-anchored routing mode, it derives a shared price $\pi^*$ from the order statistic of the base utilities at the \texttt{Refine} quota and the relative cost $c_{\mathrm{rf}}$ of token-level attention. The price sets the two promotion thresholds $\tau_{\mathrm{rf}}$ and $\tau_{\mathrm{sk}}$. Given these thresholds and quotas $B_{\mathrm{rf}}$ and $B_{\mathrm{sk}}$, the router selects \texttt{Refine} first and \texttt{Skim} second, assigning all remaining eligible tiles to \texttt{Skip}:
\begin{equation}
  \begin{aligned}
    \mathcal{T}_{\mathrm{refine}} &= \operatorname{First}_{B_{\mathrm{rf}}}
      \bigl(\{m\in\mathcal{C}: u_m^{\mathrm{refine}} > \tau_{\mathrm{rf}}\}\bigr),\\
    \mathcal{T}_{\mathrm{skim}} &= \operatorname{First}_{B_{\mathrm{sk}}}
      \bigl(\{m\in\mathcal{C}\setminus\mathcal{T}_{\mathrm{refine}}: u_m^{\mathrm{skim}} > \tau_{\mathrm{sk}}\}\bigr),\\
    \mathcal{T}_{\mathrm{skip}} &= \mathcal{H}\setminus
      \bigl(\mathcal{T}_{\mathrm{refine}}\cup\mathcal{T}_{\mathrm{skim}}\bigr).
  \end{aligned}
  \label{eq:shadow-price}
\end{equation}
Here $\operatorname{First}_B$ keeps at most $B$ qualifying tiles in candidate order. The default serving budget is $16\%$ Refine, $32\%$ Skim, and $52\%$ Skip, with $c_{\mathrm{rf}}=12$. The \texttt{Skip} set is taken over $\mathcal{H}$, so tiles outside the candidate list still contribute to attention through their means. This reduces the number of keys and values evaluated without dropping eligible history tiles. The logits for pooled summaries include a log-cardinality correction for the number of represented tokens, with an additional variance correction for \texttt{Skip}. The kernel packs summaries from multiple history tiles into aligned compute tiles and evaluates all three tiers in one online-softmax pass.

\paragraph{Sliding-window RoPE updates.}
As the window advances, retained tiles can shift to new logical temporal positions. For the residual term in Equation~\eqref{eq:refine-dc-res}, forward rotation of a key residual is equivalent to inverse rotation of the query in the dot product before quantization. Applying the relative shift on the query side lets the runtime reuse the cached FP8 key residuals. It does not need a separate unrotated copy of the token-level keys or repeated dequantization, rotation, and requantization of the residuals.

\subsection{Low-Latency Condition Updates}
\label{sec:infra-prompt-switch}

An in-play update to the prompt or reference media changes the conditioning for subsequent blocks. The renderer retains recent generated latents but rebuilds their DiT K/V under the updated condition before generation resumes. To reduce the cost of an update, we reuse Qwen prefix computations when possible and combine the DiT cache rebuild with the first denoising step under the new condition.

\paragraph{Incremental Qwen conditioning.}
Within each session, we cache Qwen attention K/V and output features. For a text update with unchanged reference media, token IDs and multimodal position IDs identify the longest common prefix, so Qwen encodes only the portion after that prefix. The suffix queries attend to the cached prefix K/V, and the resulting features are concatenated with the cached prefix features. This supports appended events and edits to existing descriptions. A change to the reference media invalidates the Qwen prefix cache and requires a full Qwen encoding pass.

\paragraph{Joint state rebuilding.}
We extend the joint execution in \secref{sec:infra-gpu} to evaluate the conditioning prefix, recent clean target history, current source, and first target denoising step in one forward pass. The same branch-specific attention, modulation, and commit ordering preserve serial dependencies. If the DiT prefix K/V for the updated condition are already cached, the runtime reuses them instead of recomputing the prefix in the joint forward. The stream then continues under the updated condition.

\section{Results}
\label{sec:results}

We assess reproducible world execution and agent-based world construction in \secref{sec:engine-results}, video editing and controllable world rendering in \secref{sec:renderer-results}, and the inference infrastructure's streaming throughput, latency, and memory efficiency in \secref{sec:system-results}.

\subsection{Engine Capabilities}
\label{sec:engine-results}

Controllable Rendering places two demands on the control side. It must decide every outcome by executing commands under the world's rules (C1), so that the outcome and magnitude of an action are fixed before rendering begins. It must also be inexpensive to build, so that a coding agent can produce a new executable world for each form of play. We first examine execution in offline proxy worlds, reading WorldKit state at each 60~Hz simulation step without the renderer. We then follow a coding agent through a complete authoring session, and finally examine gameplay execution in an agent-authored world.

\subsubsection{Rule Execution and Prebuilt 3C}
\label{sec:engine-execution}

\begin{figure}[H]
  \centering
  \begin{subfigure}[t]{0.98\linewidth}
    \centering
    \includegraphics[width=0.325\linewidth]{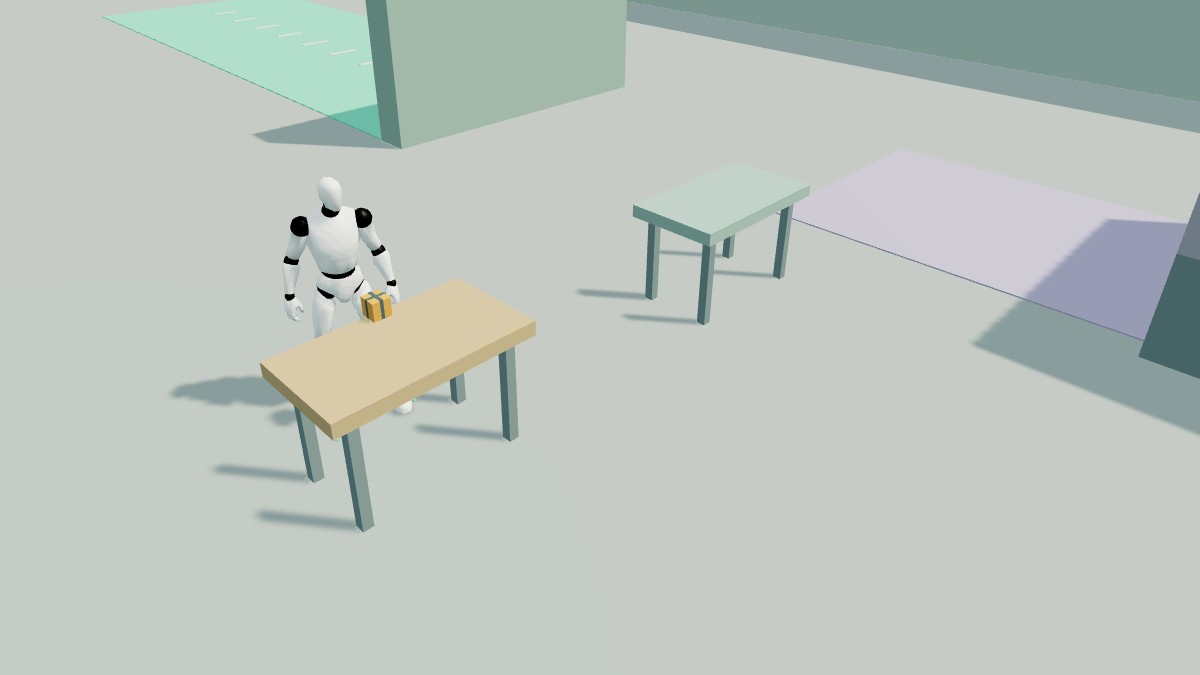}\hfill
    \includegraphics[width=0.325\linewidth]{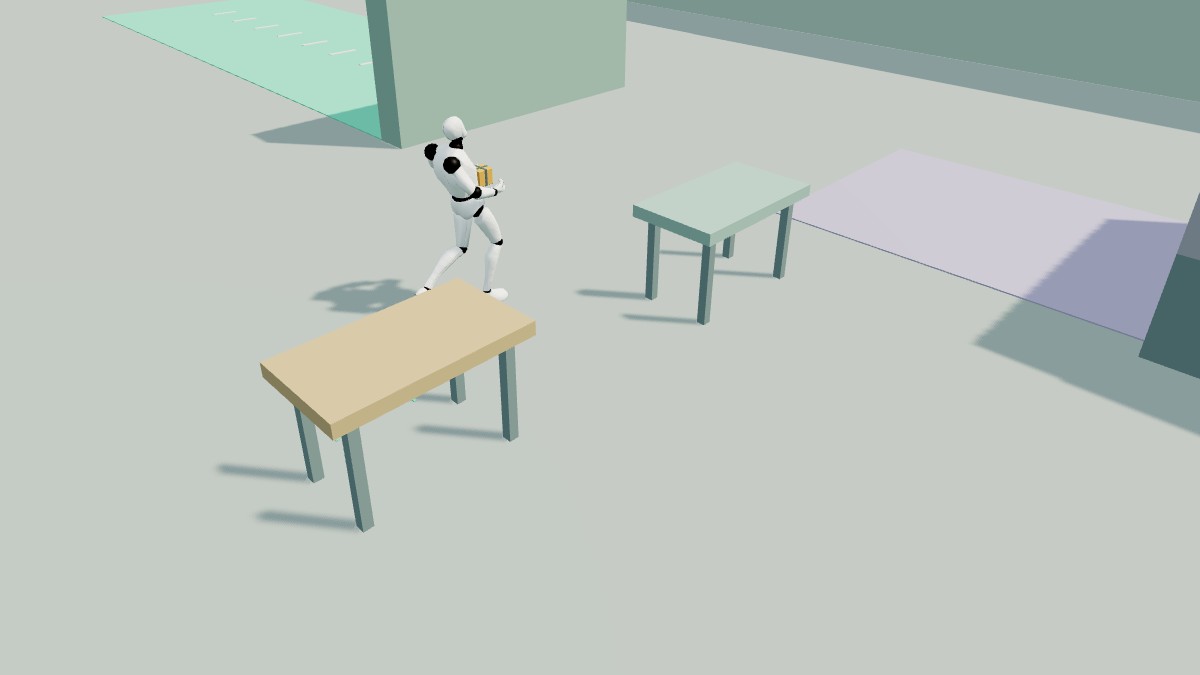}\hfill
    \includegraphics[width=0.325\linewidth]{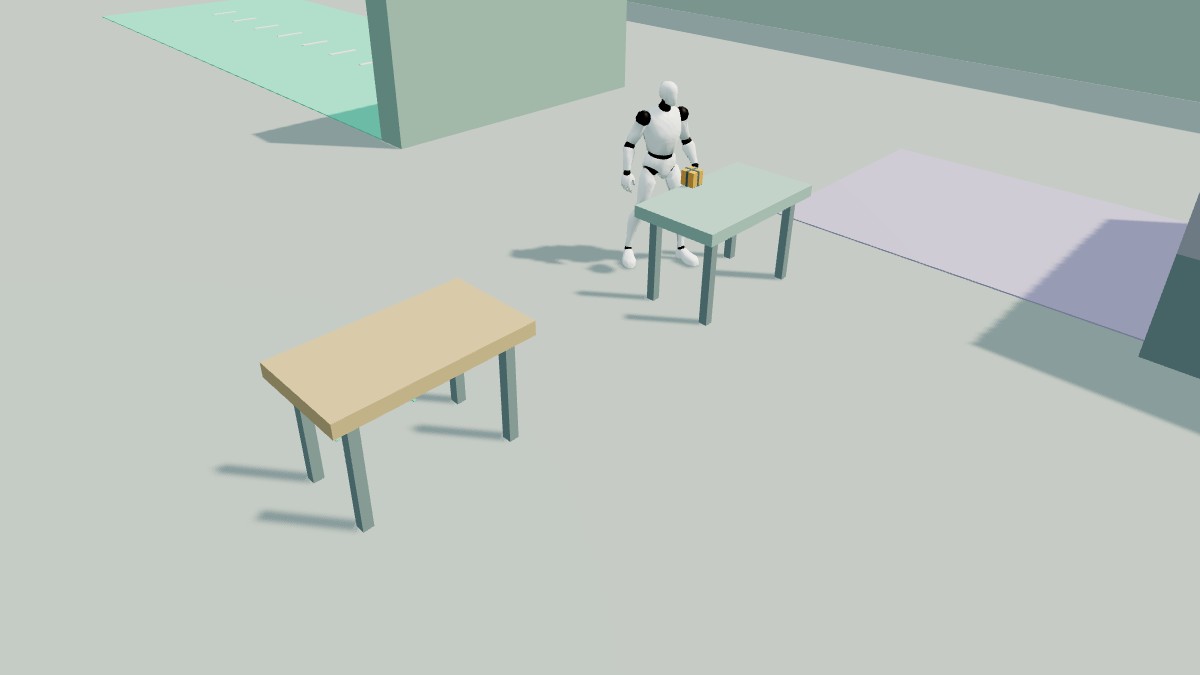}
    \caption{Object transfer: before pickup, carrying, and placed on the second table}
  \end{subfigure}\\[6pt]
  \begin{subfigure}[t]{0.98\linewidth}
    \centering
    \includegraphics[width=0.325\linewidth]{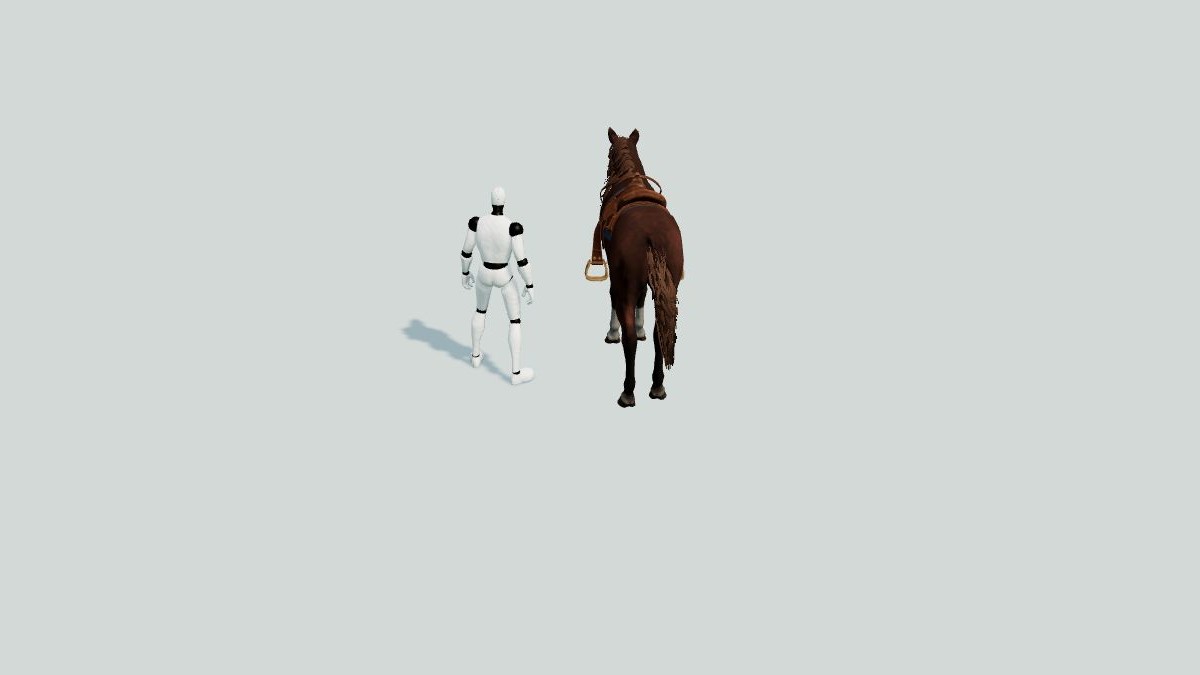}\hfill
    \includegraphics[width=0.325\linewidth]{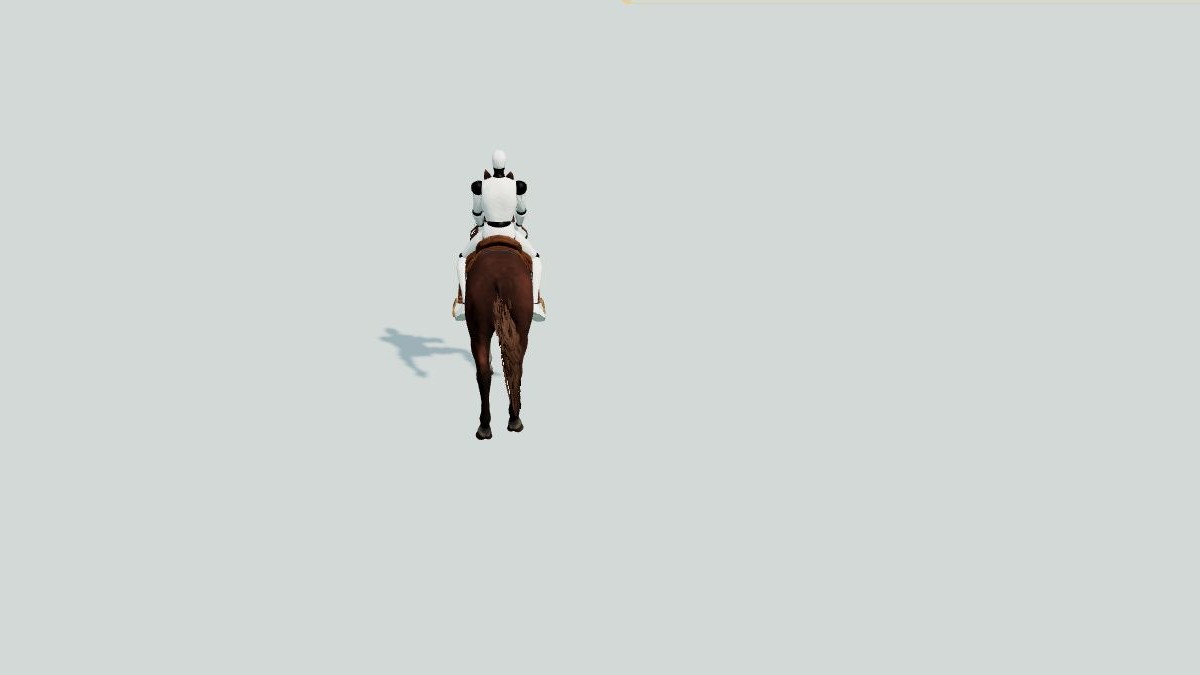}\hfill
    \includegraphics[width=0.325\linewidth]{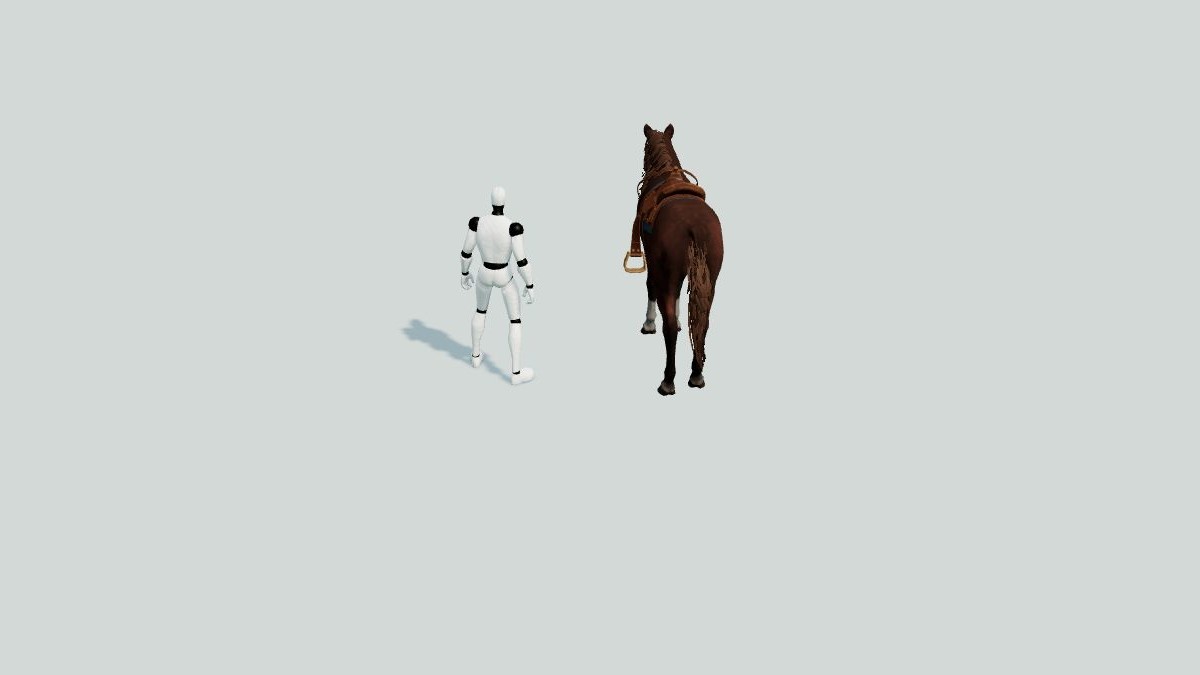}
    \caption{Mount: approach, riding after mounting, and dismounted}
  \end{subfigure}\\[6pt]
  \begin{subfigure}[t]{0.98\linewidth}
    \centering
    \includegraphics[width=0.325\linewidth]{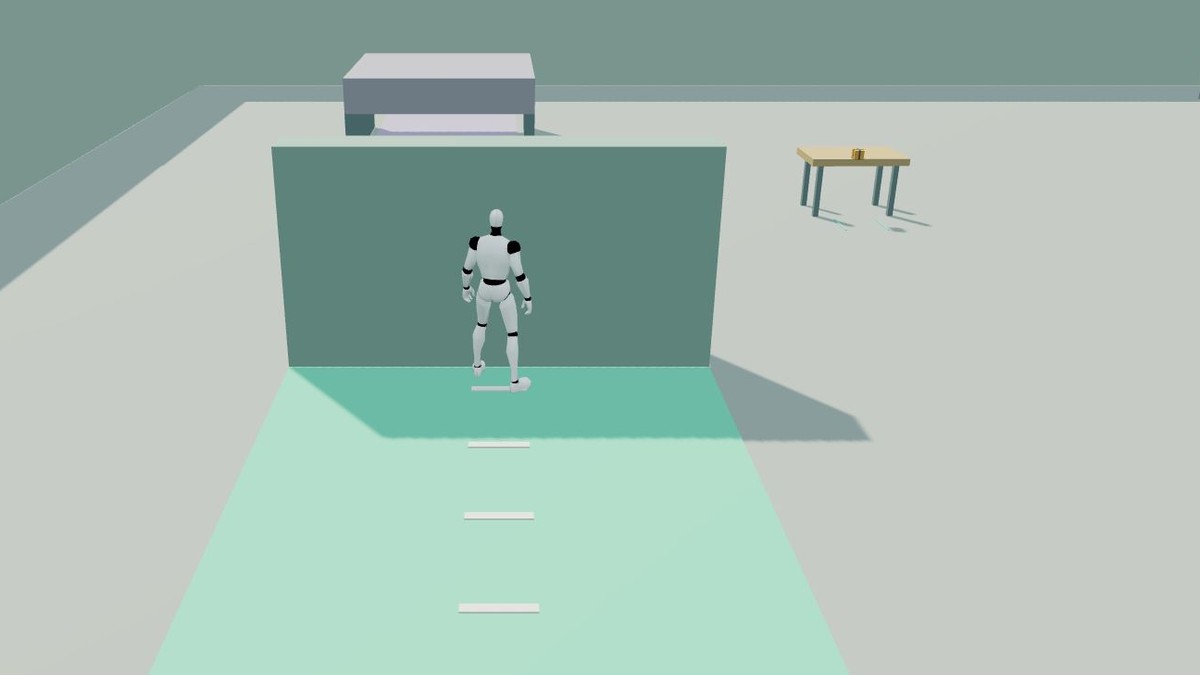}\hfill
    \includegraphics[width=0.325\linewidth]{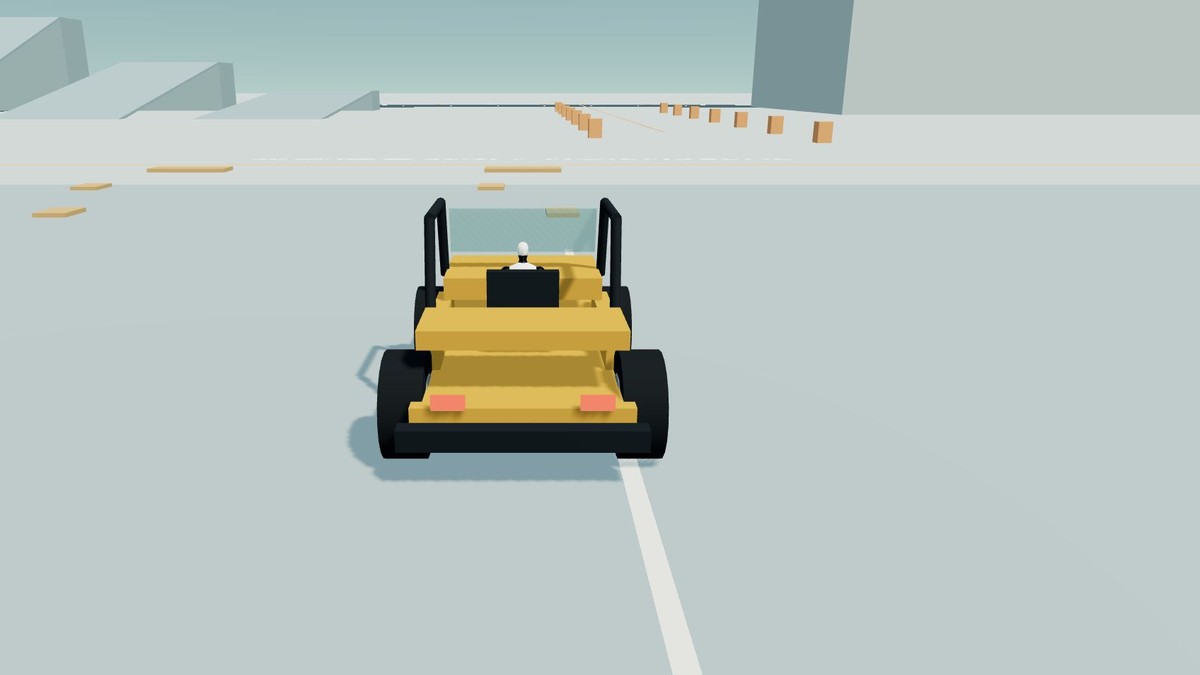}\hfill
    \includegraphics[width=0.325\linewidth]{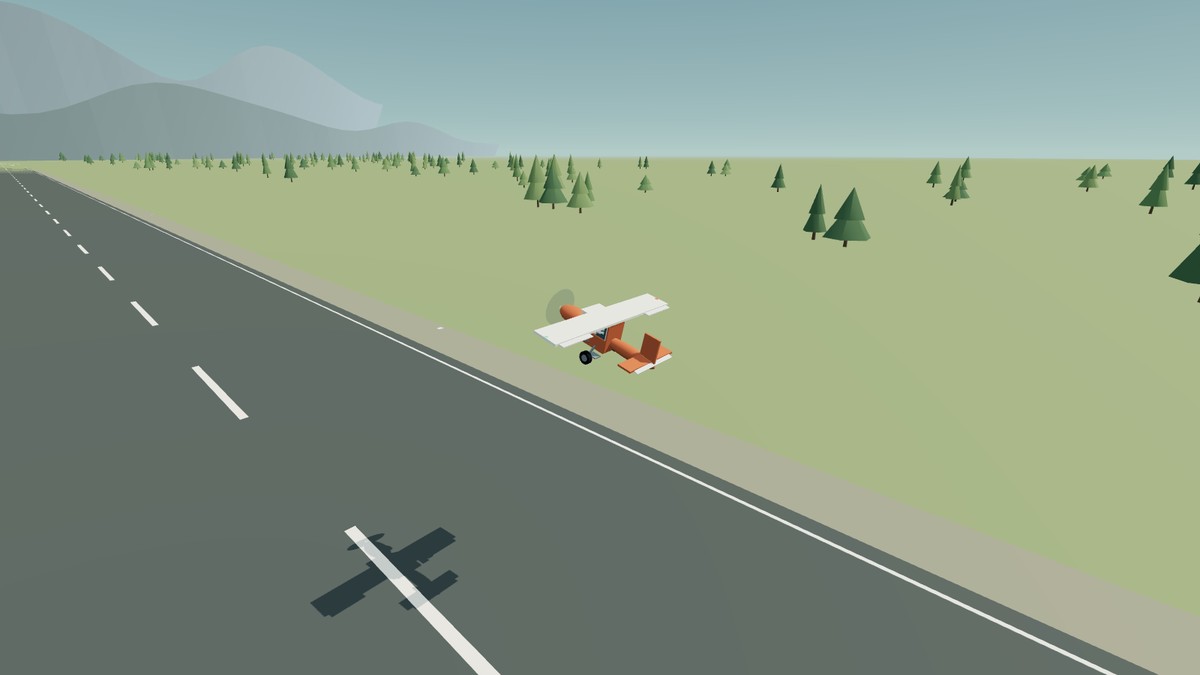}
    \caption{Wall contact, vehicle driver, and fixed-wing takeoff}
  \end{subfigure}
  \caption{\textbf{Interactions and locomotion with prebuilt 3C.} Rows (a) and (b) show interaction sequences read from left to right. Row (c) presents independent examples of wall contact, vehicle driving, and fixed-wing takeoff. Examples were recorded in separate offline proxy sessions using two WorldKit builds.}
  \label{fig:engine-3c}
\end{figure}

WorldKit's prebuilt 3C supplies reusable assets and controllers for the interactions and locomotion in \figref{fig:engine-3c} (\secref{sec:world-authoring}). Object pickup, carrying, and placement establish and release an attachment between the object and character (\figpanelref{fig:engine-3c}{a}), while mounting and dismounting establish and clear the rider relationship (\figpanelref{fig:engine-3c}{b}). Vehicle driving and fixed-wing flight in \figpanelref{fig:engine-3c}{c} illustrate continuous motion governed by subject-specific controllers and engine physics. In separate humanoid and vehicle tests, fixed inputs lasting 0.5, 1, and 2~s were each repeated five times from identical initial states, producing matching displacements and headings up to floating-point round-off. Environment geometry constrains this motion, as illustrated by the humanoid stopped at a wall under continued forward input in the left image of \figpanelref{fig:engine-3c}{c}. A 90-step collision check recorded no collision-capsule penetration and less than $10^{-5}$~m of position variation after contact. Together, these interaction sequences and motion checks illustrate the extensible interactions and geometric motion control compared in \tabref{tab:playable-world-comparison}. Changes in pose, attachment, and control are determined by the engine from world state, rules, and controllers rather than inferred by the video model.

\begin{figure}[H]
  \centering
  \begin{subfigure}[t]{0.98\linewidth}
    \centering
    \includegraphics[width=0.325\linewidth]{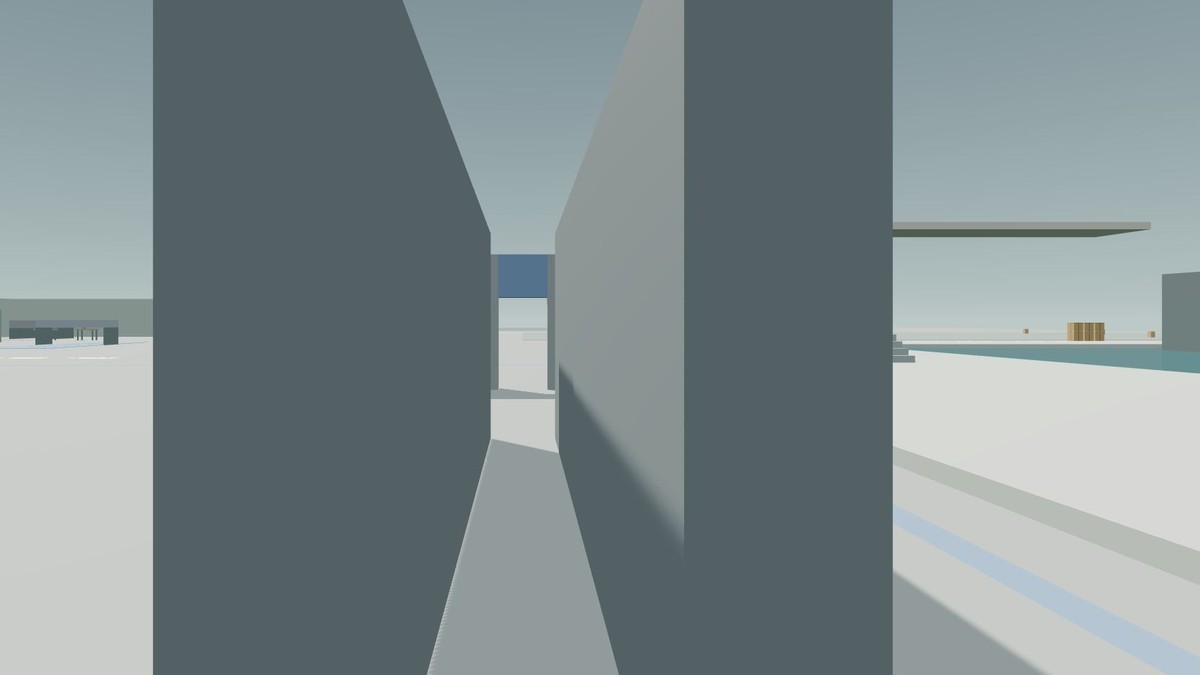}\hfill
    \includegraphics[width=0.325\linewidth]{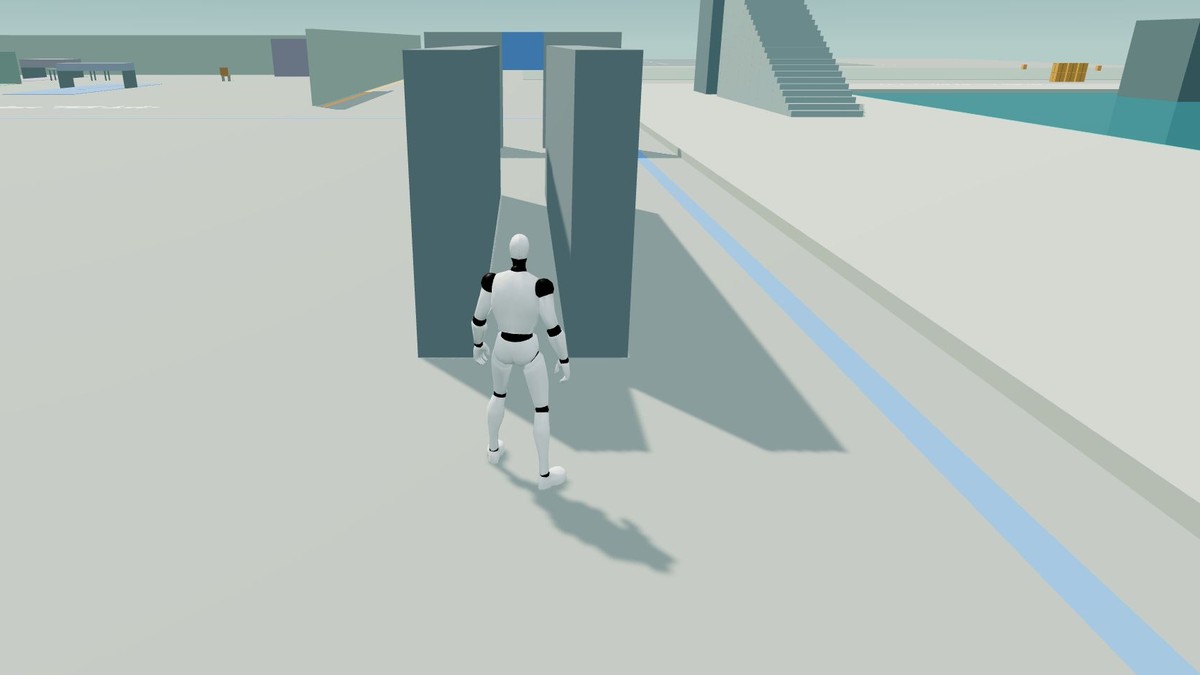}\hfill
    \includegraphics[width=0.325\linewidth]{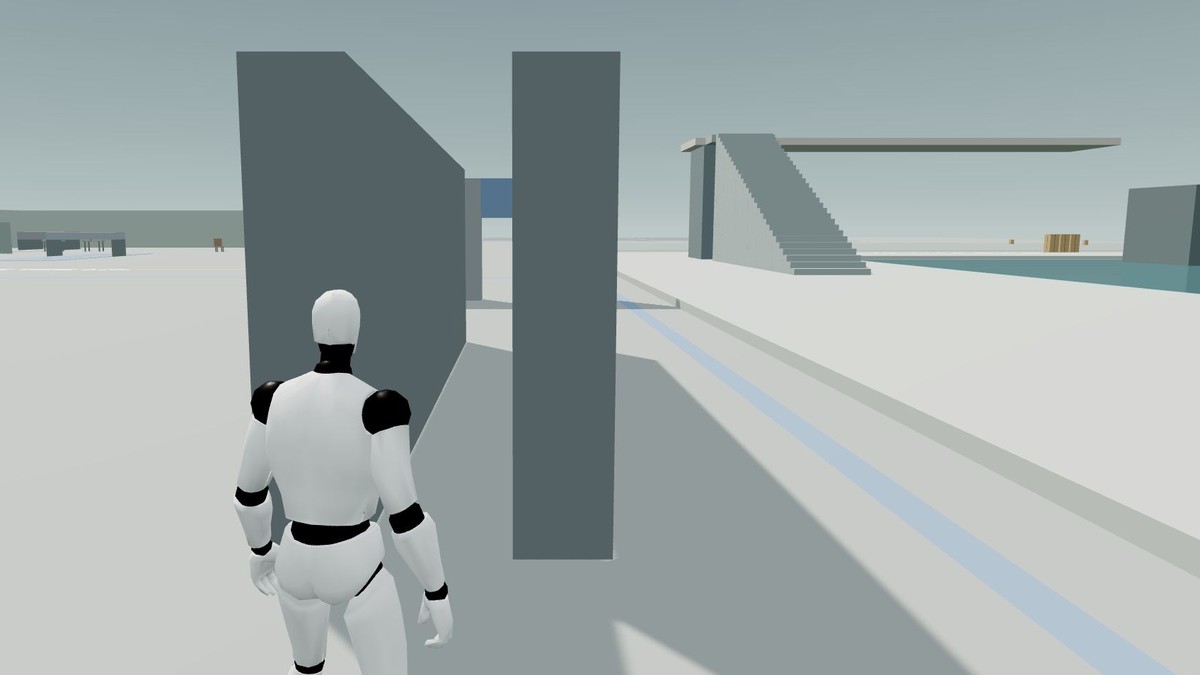}
    \caption{Camera modes: first person, third person, and over the shoulder}
  \end{subfigure}\\[6pt]
  \begin{subfigure}[t]{0.98\linewidth}
    \centering
    \begin{minipage}[t]{0.78\linewidth}
      \vspace{0pt}
      \includegraphics[width=\linewidth,trim=0bp 37bp 0bp 0bp,clip]{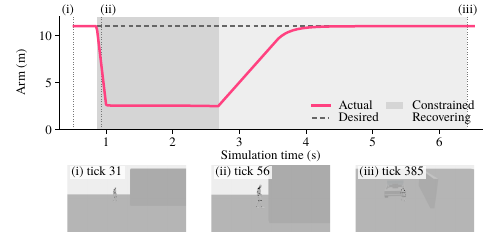}
    \end{minipage}\hfill
    \begin{minipage}[t]{0.185\linewidth}
      \vspace{0pt}
      \includegraphics[width=\linewidth,trim=32.4bp 12bp 148.3bp 79.2bp,clip]{figures/engine-camera-arm-serif.pdf}\\[2pt]
      \includegraphics[width=\linewidth,trim=101.6bp 12bp 79.1bp 79.2bp,clip]{figures/engine-camera-arm-serif.pdf}\\[2pt]
      \includegraphics[width=\linewidth,trim=170.6bp 12bp 10.1bp 79.2bp,clip]{figures/engine-camera-arm-serif.pdf}
    \end{minipage}
    \caption{Follow-camera arm and corresponding proxy frames}
  \end{subfigure}
  \caption{\textbf{Camera control in the engine.} (a) Three views of the same paused world. (b) Recorded follow-camera arm length and proxy frames at steps 31, 56, and 385. The arm shortens to 2.50~m when the wall blocks the view. After the camera turns away at step 162, the arm returns to the desired 11.0~m by step 386.}
  \label{fig:engine-camera}
\end{figure}

The camera is the last part of the observation that the engine fixes before rendering. The three prebuilt camera modes observe the same subject from different viewpoints (\figpanelref{fig:engine-camera}{a}). Switching among them at one simulation step left every entity pose unchanged, so the choice of view changes what is observed but not what happens. The follow camera keeps an explicit arm length in the engine state. When a wall enters its line of sight, the engine shortens the arm so that the subject stays visible, and it extends the arm again once the view is clear (\figpanelref{fig:engine-camera}{b}). The renderer therefore receives a proxy view in which viewpoint and occlusion have already been resolved, as C2 requires.

\subsubsection{Authoring by a Coding Agent}
\label{sec:engine-authoring}

\begin{figure}[H]
  \centering
  \begingroup
  \input{figures/renderer-figure-styles}
  \definecolor{engineFailure}{HTML}{BF3E48}
  \definecolor{engineSuccess}{HTML}{267848}
  \tikzset{
    enginefail/.pic={
      \fill[engineFailure!12] (0,0) circle[radius=.15];
      \draw[engineFailure,line width=.8pt,line cap=round] (-.075,-.075)--(.075,.075) (-.075,.075)--(.075,-.075);
    },
    enginepass/.pic={
      \fill[engineSuccess!12] (0,0) circle[radius=.15];
      \draw[engineSuccess,line width=.9pt,line cap=round,line join=round,sharp corners] (-.09,0)--(-.025,-.065)--(.10,.09);
    }
  }
  \resizebox{0.98\linewidth}{!}{%
  \begin{tikzpicture}[font=\sffamily\fontsize{10.25}{12.5}\selectfont]
    \path[use as bounding box] (-.05,0) rectangle (17.75,9.0);
    \fill[zingSoft,rounded corners=4pt] (0,1.2) rectangle (4.65,8.9);
    \fill[zingPinkWash,rounded corners=4pt] (5.2,1.2) rectangle (10.45,8.9);
    \fill[zingSoft,rounded corners=4pt] (11,1.2) rectangle (17.7,8.9);
    \node[rtitle,anchor=north west,inner sep=0pt,align=left,text width=4.05cm,font=\sffamily\bfseries\fontsize{11.25}{13.5}\selectfont] at (.2,8.55) {\textcolor{zingPink}{(a)} Construction\\\hspace*{1.4em}and repair};
    \node[rtitle,anchor=north west,inner sep=0pt,font=\sffamily\bfseries\fontsize{11.25}{13.5}\selectfont] at (5.4,8.55) {\textcolor{zingPink}{(b)} World preview};
    \node[rtitle,anchor=north west,inner sep=0pt,font=\sffamily\bfseries\fontsize{11.25}{13.5}\selectfont] at (11.2,8.55) {\textcolor{zingPink}{(c)} Recorded playtest};
    \draw[rarrow] (4.65,4.95)--(5.2,4.95);
    \draw[rarrow] (10.45,4.95)--(11,4.95);
    \node[rbox,text width=3.75cm,minimum height=.65cm] (task) at (2.325,7.0) {Short task description};
    \node[rbox,text width=3.75cm,minimum height=1.35cm] (assets) at (2.325,5.45) {\textbf{Asset selection}\\[5pt]\tikz[baseline=-.75ex]{\pic{enginefail};}\hspace{3pt}\textcolor{engineFailure}{Invalid response}\\[2pt]{\fontsize{9.25}{11.25}\selectfont\textrightarrow\ Retry with defaults}};
    \node[rbox,text width=3.75cm,minimum height=1.45cm] (compile) at (2.325,3.2) {\textbf{World compilation}\\[5pt]\tikz[baseline=-.75ex]{\pic{enginefail};}\hspace{3pt}\textcolor{engineFailure}{Asset loading failed}\\[2pt]{\fontsize{9.25}{11.25}\selectfont\textrightarrow\ Use asset manifest}};
    \draw[rarrow] (task.south)--(assets.north);
    \draw[rarrow] (assets.south)--(compile.north);
    \draw[rarrow] (compile.south)--(2.325,1.95);
    \node[rlabel,text=engineSuccess,font=\sffamily\fontsize{9.25}{11.25}\selectfont] at (2.325,1.65) {\tikz[baseline=-.75ex]{\pic{enginepass};}\hspace{3pt}Compiled world};
    \node[inner sep=0pt] at (7.825,4.95) {\includegraphics[width=4.95cm]{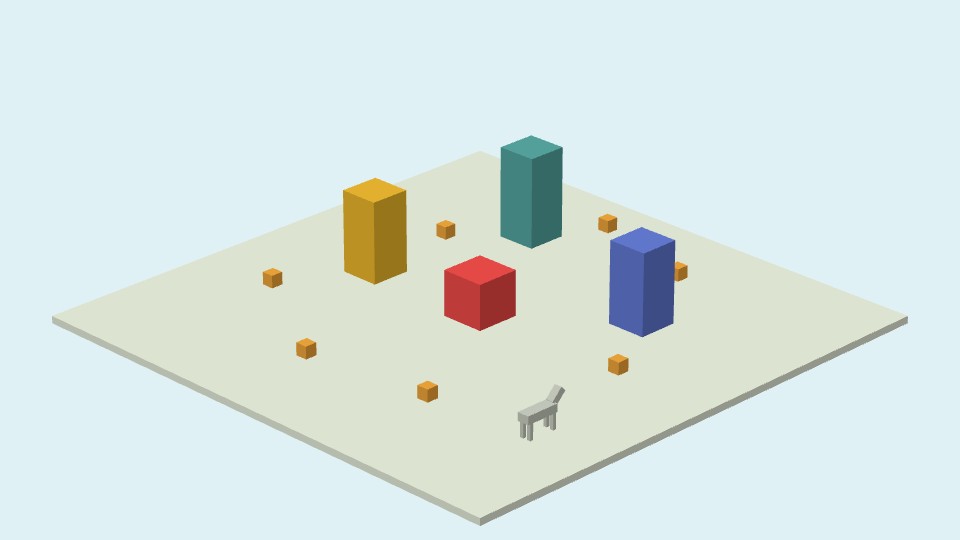}};
    \node[rsmall,text width=4.8cm,font=\sffamily\fontsize{10.25}{12.5}\selectfont] at (7.825,3.1) {Executable collection course};
    \node[inner sep=0pt] at (14.35,4.9) {\begin{tikzpicture}[x=1cm,y=1cm,font=\sffamily\fontsize{9.25}{11.25}\selectfont,text=zingInk,sharp corners]
  \path[use as bounding box] (0,0) rectangle (6.35,6.15);
  \path[fill=black!2] (1.17926,5.32576) rectangle (5.47074,1.02424);
  \draw[black!47,line width=.5pt] (1.17926,5.32576) rectangle (5.47074,1.02424);
  \path[fill=black!25,draw=black!47,line width=.5pt] (2.45041,2.08056) rectangle (2.76548,1.77020);
  \path[fill=black!25,draw=black!47,line width=.5pt] (2.09731,3.33290) rectangle (2.40695,3.01710);
  \path[fill=black!25,draw=black!47,line width=.5pt] (4.05835,2.61417) rectangle (4.36799,2.30381);
  \path[fill=zingPink,draw=none,even odd rule]
    (1.90718,4.43823)
    --(2.93931,4.43823)
    --(2.98277,4.43279)
    --(2.99907,4.43823)
    --(3.97687,4.43823)
    --(3.96600,4.40556)
    --(1.92891,4.40556)
    --(1.92891,3.96996)
    --(1.89632,3.95907)
    --(1.89632,4.42734)
    --cycle;
  \path[fill=zingPink,draw=none,even odd rule]
    (4.00946,4.31300)
    --(4.04206,4.30211)
    --(4.04749,3.84473)
    --(4.06379,3.81206)
    --(4.62874,3.24578)
    --(4.60701,3.22400)
    --(4.56898,3.25667)
    --(4.03119,3.79573)
    --(4.03119,3.80662)
    --(4.01489,3.80117)
    --(3.98773,3.82840)
    --(3.98773,3.86107)
    --(4.00403,3.87740)
    --cycle;
  \path[fill=zingPink,draw=none,even odd rule]
    (2.28201,4.11153)
    --(3.46081,4.11153)
    --(3.46081,3.28390)
    --(3.42821,3.28390)
    --(3.42821,4.07886)
    --(2.29287,4.07886)
    --(2.27114,4.06797)
    --(2.26571,3.61060)
    --(2.25485,3.57793)
    --(2.23312,3.57793)
    --(2.22769,3.58882)
    --(2.22225,3.87740)
    --(2.20596,3.89374)
    --(2.01040,3.89918)
    --(2.00496,3.92641)
    --(2.22225,3.92641)
    --(2.23312,3.91552)
    --(2.23855,4.08431)
    --cycle;
  \path[fill=zingPink,draw=none,even odd rule]
    (4.73195,3.21856)
    --(4.93294,3.21856)
    --(4.93294,1.70486)
    --(4.88949,1.67219)
    --(4.87862,1.66674)
    --(3.96600,1.66674)
    --(3.97687,1.68852)
    --(3.98230,1.69941)
    --(4.87862,1.70486)
    --(4.89492,1.71575)
    --(4.89492,3.16411)
    --(4.88949,3.18044)
    --(4.74282,3.18589)
    --cycle;
  \path[fill=zingPink,draw=none,even odd rule]
    (3.23808,3.10966)
    --(3.26525,3.08788)
    --(2.93388,2.75574)
    --(2.90672,2.76663)
    --(2.91758,2.78841)
    --cycle;
  \path[fill=zingPink,draw=none,even odd rule]
    (2.21682,2.69584)
    --(2.78178,2.69584)
    --(2.79264,2.66317)
    --(2.22769,2.65773)
    --(2.21682,2.63595)
    --(2.21139,2.48349)
    --(2.18423,2.47804)
    --(2.18423,2.65228)
    --(2.20596,2.69040)
    --cycle;
  \path[fill=zingPink,draw=none,even odd rule]
    (1.77138,2.39637)
    --(2.12447,2.39637)
    --(2.12991,2.36370)
    --(1.78224,2.35826)
    --(1.77138,2.33648)
    --(1.77138,1.46528)
    --(1.78768,1.44895)
    --(2.89585,1.44350)
    --(2.90672,1.41628)
    --(1.77138,1.41628)
    --(1.73879,1.44895)
    --(1.73879,2.34737)
    --(1.76051,2.39093)
    --cycle;
  \path[fill=zingPink,draw=none,even odd rule]
    (3.92798,1.66130)
    --(3.93884,1.66130)
    --(3.93884,1.45984)
    --(3.90625,1.41628)
    --(3.03166,1.41628)
    --(3.04796,1.44350)
    --(3.88995,1.44895)
    --(3.90625,1.47073)
    --(3.91168,1.65585)
    --cycle;
  \draw[fill=white,draw=black!80,line width=.65pt] (4.03934,4.38923) circle[radius=0.07741cm];
  \draw[fill=white,draw=black!80,line width=.65pt] (3.91168,1.74025) circle[radius=0.07741cm];
  \draw[fill=white,draw=black!80,line width=.65pt] (1.93163,3.88829) circle[radius=0.07741cm];
  \draw[fill=white,draw=black!80,line width=.65pt] (2.96919,1.47345) circle[radius=0.07605cm];
  \draw[fill=white,draw=black!80,line width=.65pt] (4.66133,3.17228) circle[radius=0.07741cm];
  \draw[fill=white,draw=black!80,line width=.65pt] (2.20324,2.40454) circle[radius=0.07605cm];
  \draw[fill=white,draw=black!80,line width=.65pt] (2.86054,2.70673) circle[radius=0.07741cm];
  \draw[fill=white,draw=black!80,line width=.65pt] (2.96919,4.51446) circle[radius=0.07469cm];
  \path[fill=black,draw=none] (3.24895,3.24578) rectangle (3.39562,3.09877);
  \draw[black,line width=.9pt] (3.37389,3.36013)--(3.51513,3.21856) (3.37389,3.21856)--(3.51513,3.36013);
  \node[font=\sffamily\bfseries\fontsize{9.25}{11.25}\selectfont,inner sep=0pt] at (2.96376,4.78399) {7};
  \node[font=\sffamily\bfseries\fontsize{9.25}{11.25}\selectfont,inner sep=0pt] at (4.25663,4.55258) {6};
  \node[font=\sffamily\bfseries\fontsize{9.25}{11.25}\selectfont,inner sep=0pt] at (1.68446,3.94002) {8};
  \node[font=\sffamily\bfseries\fontsize{9.25}{11.25}\selectfont,inner sep=0pt] at (4.68306,3.39008) {5};
  \node[font=\sffamily\bfseries\fontsize{9.25}{11.25}\selectfont,inner sep=0pt] at (3.13759,2.73124) {1};
  \node[font=\sffamily\bfseries\fontsize{9.25}{11.25}\selectfont,inner sep=0pt] at (1.97780,2.19218) {2};
  \node[font=\sffamily\bfseries\fontsize{9.25}{11.25}\selectfont,inner sep=0pt] at (4.18601,1.52790) {4};
  \node[font=\sffamily\bfseries\fontsize{9.25}{11.25}\selectfont,inner sep=0pt] at (2.90943,1.19576) {3};
  \draw[zingInk,line width=.55pt] (1,.85) rectangle (5.65,5.5);
  \draw[zingInk,line width=.45pt] (1.17885,.85)--(1.17885,.75);
  \node[anchor=north,inner sep=0pt] at (1.17885,.68) {-12};
  \draw[zingInk,line width=.45pt] (1,1.02885)--(.9,1.02885);
  \node[anchor=east,inner sep=0pt] at (.8,1.02885) {-12};
  \draw[zingInk,line width=.45pt] (2.25192,.85)--(2.25192,.75);
  \node[anchor=north,inner sep=0pt] at (2.25192,.68) {-6};
  \draw[zingInk,line width=.45pt] (1,2.10192)--(.9,2.10192);
  \node[anchor=east,inner sep=0pt] at (.8,2.10192) {-6};
  \draw[zingInk,line width=.45pt] (3.32500,.85)--(3.32500,.75);
  \node[anchor=north,inner sep=0pt] at (3.32500,.68) {0};
  \draw[zingInk,line width=.45pt] (1,3.17500)--(.9,3.17500);
  \node[anchor=east,inner sep=0pt] at (.8,3.17500) {0};
  \draw[zingInk,line width=.45pt] (4.39808,.85)--(4.39808,.75);
  \node[anchor=north,inner sep=0pt] at (4.39808,.68) {6};
  \draw[zingInk,line width=.45pt] (1,4.24808)--(.9,4.24808);
  \node[anchor=east,inner sep=0pt] at (.8,4.24808) {6};
  \draw[zingInk,line width=.45pt] (5.47115,.85)--(5.47115,.75);
  \node[anchor=north,inner sep=0pt] at (5.47115,.68) {12};
  \draw[zingInk,line width=.45pt] (1,5.32115)--(.9,5.32115);
  \node[anchor=east,inner sep=0pt] at (.8,5.32115) {12};
  \node[anchor=base,inner sep=0pt] at (3.325,.08) {X (m)};
  \node[rotate=90,inner sep=0pt] at (.40,3.175) {Z (m)};
  \draw[zingPink,line width=1pt] (.95,5.88)--(1.3,5.88);
  \node[anchor=west,inner sep=0pt] at (1.4,5.88) {Recorded path};
  \path[fill=black] (3.7,5.78) rectangle (3.9,5.98);
  \node[anchor=west,inner sep=0pt] at (4.02,5.88) {Start};
  \draw[black,line width=.9pt] (5.15,5.78)--(5.35,5.98) (5.15,5.98)--(5.35,5.78);
  \node[anchor=west,inner sep=0pt] at (5.45,5.88) {End};
\end{tikzpicture}};
    \node[rsmall,font=\sffamily\bfseries\fontsize{10.25}{12.5}\selectfont] at (14.35,1.5) {\textcolor{zingPink}{8/8} target regions reached};
    \node[rpink,text width=16.9cm,minimum height=.7cm] (package) at (8.85,.45) {\tikz[baseline=-.75ex]{\pic{enginepass};}\hspace{3pt}\textcolor{engineSuccess}{Technical verification passed}\hspace{.6cm}\textrightarrow\hspace{.6cm}\textbf{Submit world package}};
    \draw[rarrow,rounded corners=1pt] (14.35,1.2)--(14.35,1.0)--(5.7,1.0)--(5.7,.8);
  \end{tikzpicture}}
  \endgroup
  \caption{\textbf{Construction and verification of an agent-authored world.} (a) Two tool-reported failures and the corresponding repairs. (b) A Creator preview of the constructed course. (c) The recorded playtest path. Numbers 1--8 indicate the order of first entry into each target region, and grey squares mark pillars.}
  \label{fig:engine-authoring}
\end{figure}

Using WorldKit's prebuilt components and Creator tools, a coding agent constructed a small executable collection course from a one-paragraph task description. It selected and pinned assets, assembled the scene, and compiled the world. During construction, Creator returned structured feedback on an invalid asset-search response and an unresolved asset definition. The agent used these diagnostics to retry the search and correct the asset-loading configuration without manual intervention (\figpanelref{fig:engine-authoring}{a}). After compilation, it inspected Creator previews of the scene (\figpanelref{fig:engine-authoring}{b}) and ran a playtest that replayed player-input events. The recorded trajectory reached all eight target regions (\figpanelref{fig:engine-authoring}{c}), showing that the planned locations were reachable through the world's controls. The final submission passed technical verification and produced a sealed package containing the executable world, previews, and playtest record.

\subsubsection{Executing Playable Worlds}
\label{sec:engine-rendered-play}

\begin{figure}[!t]
  \centering
  \includegraphics[width=\linewidth]{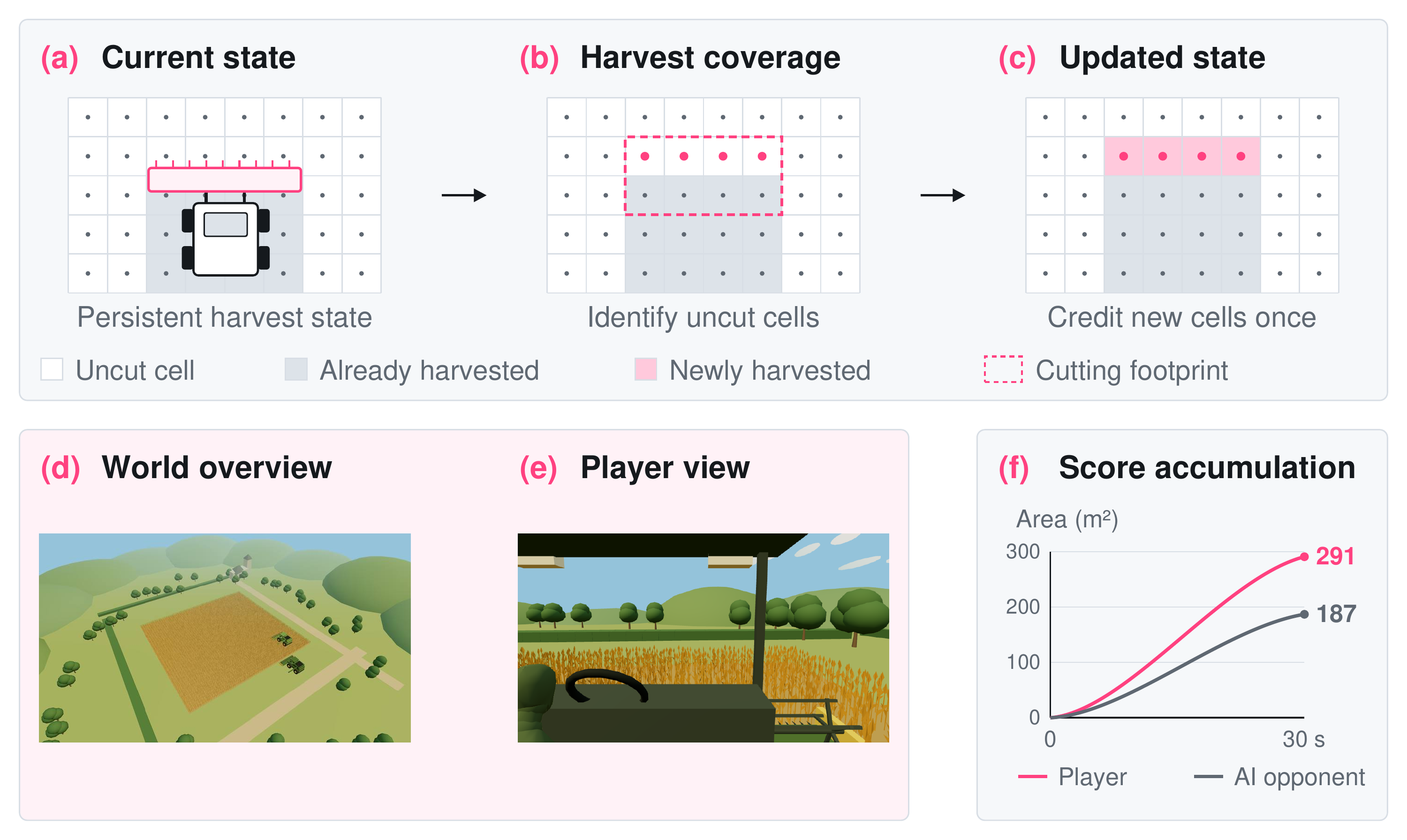}
  \caption{\textbf{World construction and rule execution in a harvest race.} (a--c) Only previously uncut 1~m$^2$ cells whose centers lie within the cutting footprint receive credit. (d--e) World overview and native player view. (f) Schematic score accumulation.}
  \label{fig:engine-farm-case}
\end{figure}

The harvest race in \figref{fig:engine-farm-case} illustrates how an agent-authored world combines reusable controls with persistent spatial state and executable game rules.

The player and an AI-controlled opponent operate combine harvesters in the same wheat field, competing to harvest the larger area within 30~s. Given a short task description, the coding agent assembled a combine harvester from custom geometry and WorldKit's wheeled-vehicle drive because Atlas contained no such asset. Both drivers reuse the same humanoid asset and version-locked motion packs. The world overview and player-view proxy in \figpanelref{fig:engine-farm-case}{d--e} show the resulting world at scene-wide and local scales.

WorldKit derives each vehicle's cutting footprint from its pose at every simulation step and identifies the cell centers it covers. A cell contributes to the vehicle's harvested area only if it was previously uncut, after which its harvested state persists. The update in \figpanelref{fig:engine-farm-case}{a--c} illustrates how new coverage changes this state without crediting already harvested cells again. Vehicle motion, the shared harvest state, and cumulative areas are therefore updated by the same executable world rules.

When the 30~s time limit expires, the accumulated areas are fixed. In a complete standard round, WorldKit computed final areas of 291~m$^2$ for the player and 187~m$^2$ for the AI opponent. These totals provide the endpoints of \figpanelref{fig:engine-farm-case}{f}. The example combines continuous vehicle control, persistent spatial state, and competitive scoring within one agent-authored world. More generally, the same construction and execution interfaces enable agents to build diverse worlds with explicitly defined rules and interaction mechanics, providing a reusable foundation for playable environments.

\subsection{Renderer Capabilities}
\label{sec:renderer-results}

Fixed-instruction and continuous-editing benchmarks quantitatively assess instruction following, reference adherence, content preservation, visual quality, and temporal stability. Qualitative examples of controllable world rendering illustrate reference-conditioned appearance variation, sustained rendering across scene transitions, and interaction outcomes retained across revisits.

\subsubsection{Fixed-Instruction Video Editing}
\label{sec:renderer-general-editing}

\paragraph{Evaluation setup.}
All editing benchmark evaluations use 480p source-aligned rendering with two sink blocks in a nine-block attention window. We evaluate 308 short videos from OpenVE-Bench~\citep{he2025openve} under the five-category protocol used by JoyAI-Video-Edit~\citep{xiao2026joyai}. We also evaluate LongV2VBench, introduced by the same work, which comprises 229 editing tasks on approximately one-minute videos. The instruction is fixed throughout each video. We follow the published scoring rubrics with Gemini 2.5 Pro and compute overall scores by macro-averaging the five categories.

\begin{table}[H]
  \centering
  \caption{\textbf{Fixed-instruction video editing.} Scores are on a 1--5 scale. Dashes indicate unreported results.}
  \label{tab:renderer-general-editing}
  \small
  \setlength{\tabcolsep}{8pt}
  \renewcommand{\arraystretch}{1.13}
  \setlength{\aboverulesep}{0pt}
  \setlength{\belowrulesep}{0pt}
  \begin{tabularx}{0.94\linewidth}{>{\raggedright\arraybackslash}Xlccc}
    \arrayrulecolor{black}\toprule[0.8pt]
    \textbf{Method} & \textbf{Setting} & \textbf{Resolution} & \textbf{OpenVE} $\uparrow$ & \textbf{LongV2V} $\uparrow$ \\
    \midrule[0.5pt]
    Kling-O1~\citep{kuaishou2025klingo1} & Offline & $1080\!\times\!1920$ & 3.62 & --- \\
    Kling-3.0 Omni~\citep{kuaishou2026kling3} & Offline & $1080\!\times\!1920$ & 3.64 & --- \\
    Bernini-R~\citep{liu2026bernini} & Offline & $480\!\times\!848$ & 3.72 & --- \\
    \midrule[0.5pt]
    StreamDiffusionV2~\citep{feng2025streamdiffusionv2} & Streaming & $480\!\times\!832$ & 1.23 & 1.21 \\
    XMax X2.0~\citep{xmax2026x2} & Streaming & $832\!\times\!1440$ & 1.87 & 1.71 \\
    LiveEdit~\citep{wang2026liveedit} & Streaming & $480\!\times\!832$ & 2.00 & 1.23 \\
    SANA-Streaming~\citep{zhao2026sanastreaming} & Streaming & $704\!\times\!1280$ & 2.62 & 1.64 \\
    JoyAI-Video-Edit~\citep{xiao2026joyai} & Streaming & $720\!\times\!1280$ & 3.60 & 3.30 \\
    \rowcolor{zingPinkWash}\textbf{Zing-1.0 (Ours)} & Streaming & $480\!\times\!832$ & \textbf{3.90} & \textbf{3.44} \\
    \bottomrule[0.6pt]
  \end{tabularx}
  \par\smallskip
  \parbox{0.94\linewidth}{\raggedright\footnotesize
    \textit{Note.} Baseline scores and resolutions are taken from JoyAI-Video-Edit~\citep{xiao2026joyai} rather than remeasured.}
\end{table}

Zing-1.0 achieves overall scores of 3.90 on OpenVE and 3.44 on LongV2VBench, outperforming all listed streaming editors on both benchmarks (\tabref{tab:renderer-general-editing}). JoyAI-Video-Edit, the strongest streaming baseline in this comparison, scores 3.60 and 3.30, respectively.

Style and local appearance changes are particularly effective, with scores above 4.0 on both benchmarks (\tabref{tab:renderer-edit-categories}). Background editing also improves over JoyAI-Video-Edit in both settings. These results support control over the visual style, subject appearance, and surroundings of an evolving scene. Object addition and removal remain more challenging on LongV2VBench, leaving room to improve persistent changes to scene content.

\begin{table}[H]
  \centering
  \caption{\textbf{Editing scores by task category.} Scores use the benchmark-specific rubrics on a 1--5 scale. JoyAI-Video-Edit results are taken from its published comparison~\citep{xiao2026joyai}. Bold indicates the higher score within each benchmark and category.}
  \label{tab:renderer-edit-categories}
  \small
  \setlength{\tabcolsep}{3.5pt}
  \renewcommand{\arraystretch}{1.13}
  \setlength{\aboverulesep}{0pt}
  \setlength{\belowrulesep}{0pt}
  \begin{tabularx}{0.94\linewidth}{l>{\raggedright\arraybackslash}Xccccc}
    \arrayrulecolor{black}\toprule[0.8pt]
    \textbf{Benchmark} & \textbf{Method} & \textbf{Global style} & \textbf{Local change} & \textbf{Background} & \textbf{Removal} & \textbf{Addition} \\
    \midrule[0.5pt]
    OpenVE & \mbox{JoyAI-Video-Edit~\citep{xiao2026joyai}} & 3.62 & \textbf{4.47} & 2.90 & 4.06 & 2.97 \\
    \rowcolor{zingPinkWash} & \textbf{Zing-1.0 (Ours)} & \textbf{4.28} & 4.42 & \textbf{3.18} & \textbf{4.24} & \textbf{3.38} \\
    \midrule[0.5pt]
    LongV2V & \mbox{JoyAI-Video-Edit~\citep{xiao2026joyai}} & 3.85 & 4.09 & 2.49 & \textbf{2.99} & \textbf{3.10} \\
    \rowcolor{zingPinkWash} & \textbf{Zing-1.0 (Ours)} & \textbf{4.20} & \textbf{4.13} & \textbf{2.90} & 2.93 & 3.01 \\
    \bottomrule[0.6pt]
  \end{tabularx}
\end{table}

\subsubsection{Continuous Editing in Streaming Sessions}
\label{sec:renderer-streaming-editing}

\paragraph{Evaluation setup.}
We evaluate all 60 cases in each of the two editing tracks of StreamEval~\citep{streameval2026}: Instruction Editing and Reference Editing. Successive requests require the system to add, replace, or selectively cancel edits while preserving those that remain active. The evaluated system combines the renderer with the online multimodal model that rewrites instructions and updates reference conditions.

We average three runs per case, then cases within each duration tier, and weight the five tiers equally (15, 30, 60, 120, and 300 seconds). Overall is the mean of the five metrics on a 0--100 scale. Edit Success assesses complete fulfillment of the requested change, while Preservation assesses retention of protected source content. Baseline scores are taken from the official leaderboard~\citep{streamevalleaderboard}.

The combined system obtains provisional overall lower bounds of 57.86 and 54.06 on the two tracks, exceeding the published baselines in \tabref{tab:renderer-streameval}.

\begin{table}[H]
  \centering
  \caption{\textbf{Continuous editing on StreamEval across the instruction and reference tracks.} Alignment reports Following and Reference, respectively. Bold marks the best score per column in each track.}
  \label{tab:renderer-streameval}
  \small
  \setlength{\tabcolsep}{5pt}
  \renewcommand{\arraystretch}{1.13}
  \setlength{\aboverulesep}{0pt}
  \setlength{\belowrulesep}{0pt}
  \begin{tabularx}{0.94\linewidth}{>{\raggedright\arraybackslash}Xccccc>{\hspace{0.45em}}c}
    \arrayrulecolor{black}\toprule[0.8pt]
    \textbf{Method} & \textbf{Quality} & \textbf{Stability} & \textbf{Alignment} & \textbf{Edit Success} & \textbf{Preservation} & \textbf{Overall} \\
    \midrule[0.5pt]
    \rowcolor[HTML]{EAF3FC}\multicolumn{7}{l}{\rule{0pt}{11.5pt}\textit{Instruction Editing}} \\
    XMax X2.0~\citep{xmax2026x2} & 5.14 & 0.97 & 11.30 & 2.59 & 28.52 & 9.70 \\
    EditaLive (14B)~\citep{li2026editalive} & 7.86 & 4.12 & 12.04 & 4.26 & 23.15 & 10.29 \\
    SANA-Streaming~\citep{zhao2026sanastreaming} & 15.15 & 9.22 & 11.48 & 4.63 & 44.26 & 16.95 \\
    Lucy 2.5~\citep{decart2026lucy25} & 19.94 & 9.40 & 36.48 & 12.12 & 65.00 & 28.59 \\
    JoyAI-Video-Edit~\citep{xiao2026joyai} & 11.63 & 9.57 & 63.89 & 43.52 & \textbf{83.33} & 42.39 \\
    Vidu S2 Editing~\citep{zhang2026vidus2} & 22.99 & 13.94 & 72.10 & 57.20 & 60.93 & 45.43 \\
    \rowcolor{zingPinkWash}\textbf{Zing-1.0 (Ours)}\textsuperscript{\textdagger} & \textbf{38.93} & \textbf{47.80} & \textbf{73.31} & \textbf{57.25} & 72.00 & \textbf{57.86} \\
    \midrule[0.5pt]
    \rowcolor[HTML]{EAF3FC}\multicolumn{7}{l}{\rule{0pt}{11.5pt}\textit{Reference Editing}} \\
    XMax X2.0~\citep{xmax2026x2} & 8.20 & 7.51 & 20.56 & 1.11 & 10.00 & 9.47 \\
    JoyAI-Video-Edit~\citep{xiao2026joyai} & 6.54 & 3.46 & 26.11 & 8.89 & 26.67 & 14.33 \\
    Lucy 2.5~\citep{decart2026lucy25} & 25.55 & 28.55 & 7.59 & 1.30 & 30.38 & 18.67 \\
    Vidu S2 Editing~\citep{zhang2026vidus2} & \textbf{37.72} & \textbf{46.74} & 31.20 & 15.00 & 64.44 & 39.02 \\
    \rowcolor{zingPinkWash}\textbf{Zing-1.0 (Ours)}\textsuperscript{\textdagger} & 36.15 & 40.11 & \textbf{70.19} & \textbf{51.22} & \textbf{72.63} & \textbf{54.06} \\
    \bottomrule[0.6pt]
  \end{tabularx}
  \par\smallskip
  \parbox{0.94\linewidth}{\raggedright\footnotesize
    \textsuperscript{\textdagger}\,The self-reported Zing-1.0 results have been submitted to StreamEval and are awaiting official verification.}
\end{table}

For instruction-driven editing, the principal gain is a more stable output stream during successive updates. Zing-1.0 reaches a Stability score of 47.80, compared with 13.94 for Vidu S2 Editing, while maintaining comparable Edit Success. This improvement is accompanied by higher visual quality, showing that effective instruction execution can be combined with cleaner, more stable rendering during continuous interaction.

For reference-driven editing, Zing-1.0 more reliably adopts the requested appearance and completes the transformation. Reference and Edit Success scores reach 70.19 and 51.22, compared with 31.20 and 15.00 for Vidu S2 Editing. Together with the highest Preservation score among the compared systems, these results support reference-conditioned appearance updates that retain protected source content. Visual refinement remains an opportunity, with Vidu scoring higher in Quality and Stability on this track.

\begin{figure}[H]
  \centering
  \includegraphics[width=\linewidth,trim=170bp 6bp 17bp 16bp,clip]{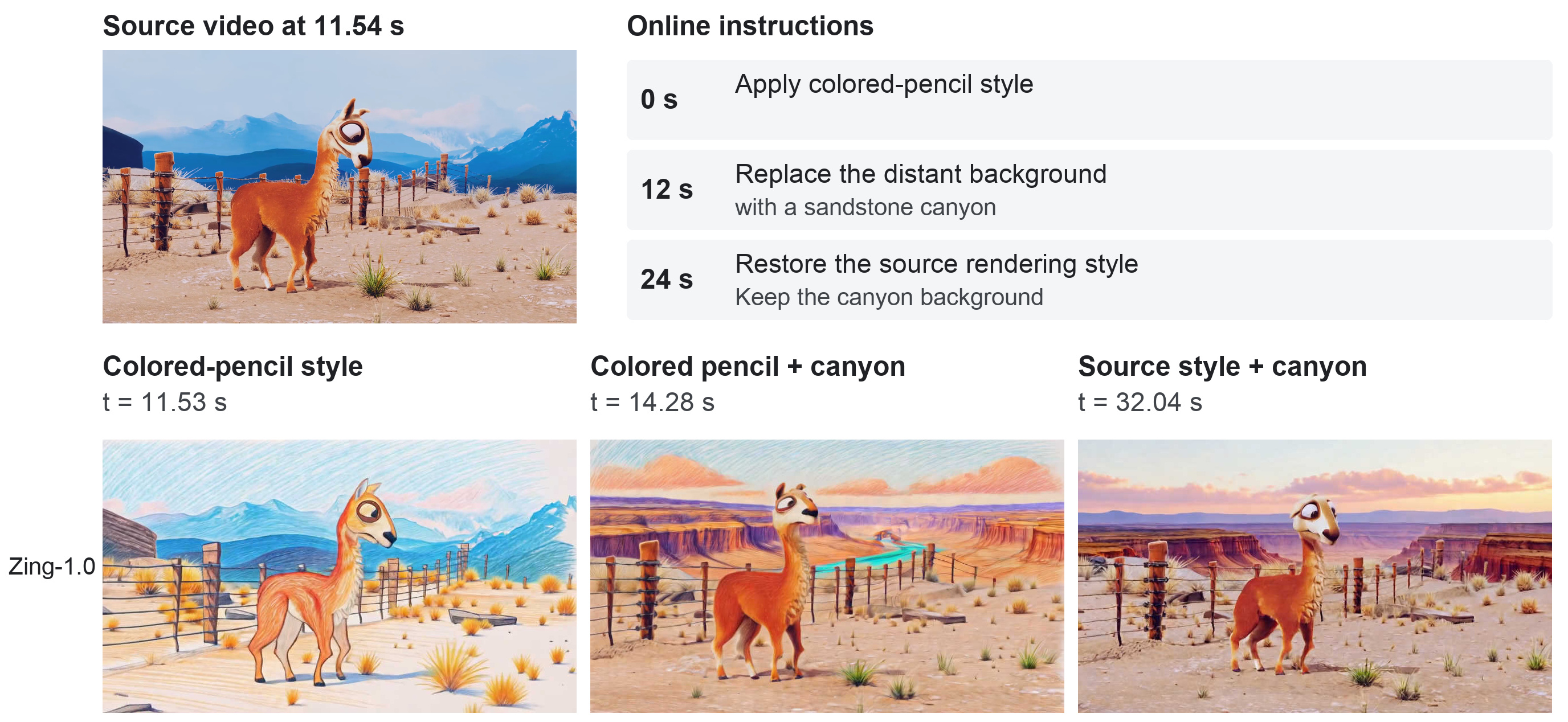}
  \caption{\textbf{Composing and selectively cancelling edits in a continuous stream.} Recorded outputs from the first three stages of a 60-second StreamEval session, case 009, Instruction Editing, repeat 2~\citep{streameval2026}. Requests at 0, 12, and 24 seconds apply a colored-pencil style, add a canyon while retaining the style, and restore the source style while retaining the canyon. The source frame at 11.54 seconds provides scene context. Output samples at 11.53, 14.28, and 32.04 seconds are timed by reception relative to session start.}
  \label{fig:renderer-continuous-editing}
\end{figure}

The recorded sequence in \figref{fig:renderer-continuous-editing} illustrates selective control over accumulated edits. Later requests revise selected appearance attributes while preserving earlier edits that remain active, allowing users to compose and partially cancel changes within the same ongoing session.

\subsubsection{Controllable World Rendering}
\label{sec:renderer-world-rendering}

\begin{figure}[H]
  \centering
  \includegraphics[width=\linewidth]{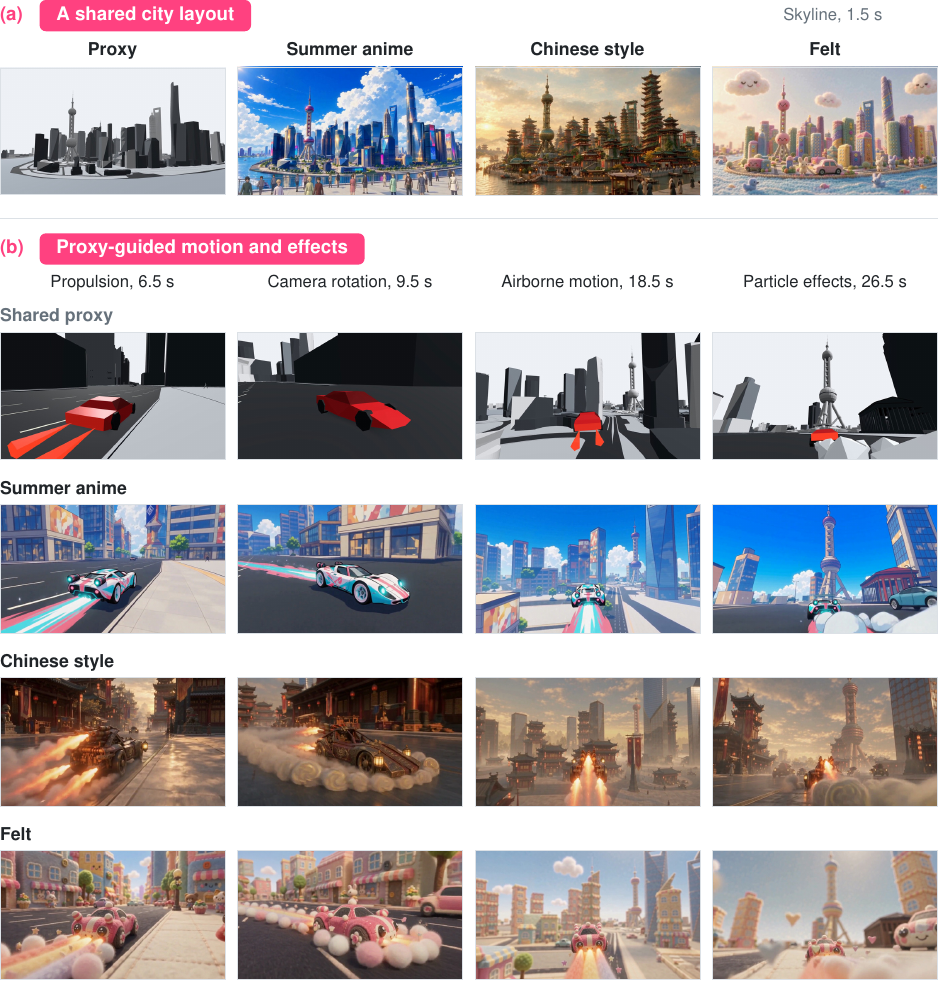}
  \caption{\textbf{Reference-conditioned rendering of a shared Shanghai proxy.} (a) Skyline views in summer-anime, Chinese-inspired, and felt styles. (b) Corresponding proxy and rendered frames at representative moments of propulsion, camera rotation, airborne motion, and particle effects. Each column uses the same elapsed timestamp across rows.}
  \label{fig:renderer-proxy-correspondence}
\end{figure}

\paragraph{Reference-conditioned appearance and proxy-guided dynamics.}
\figref{fig:renderer-proxy-correspondence} shows renderings of a shared Shanghai proxy sequence conditioned on scene and vehicle references in summer-anime, Chinese-inspired, and felt styles. The opening shot presents the city skyline with distinct materials, lighting, and architectural treatments of the same layout across the three styles. The sequence then cuts to a third-person driving view, where aligned proxy and rendered frames show propulsion, camera rotation, airborne travel, and particle effects. The renderer produces distinct vehicle designs, building appearances, and dynamic effects for each reference style while following the viewpoint and motion supplied by the proxy. These examples illustrate appearance variation and dynamic rendering without introducing action-specific control channels to the renderer.

\begin{figure}[H]
  \centering
  \includegraphics[width=\linewidth]{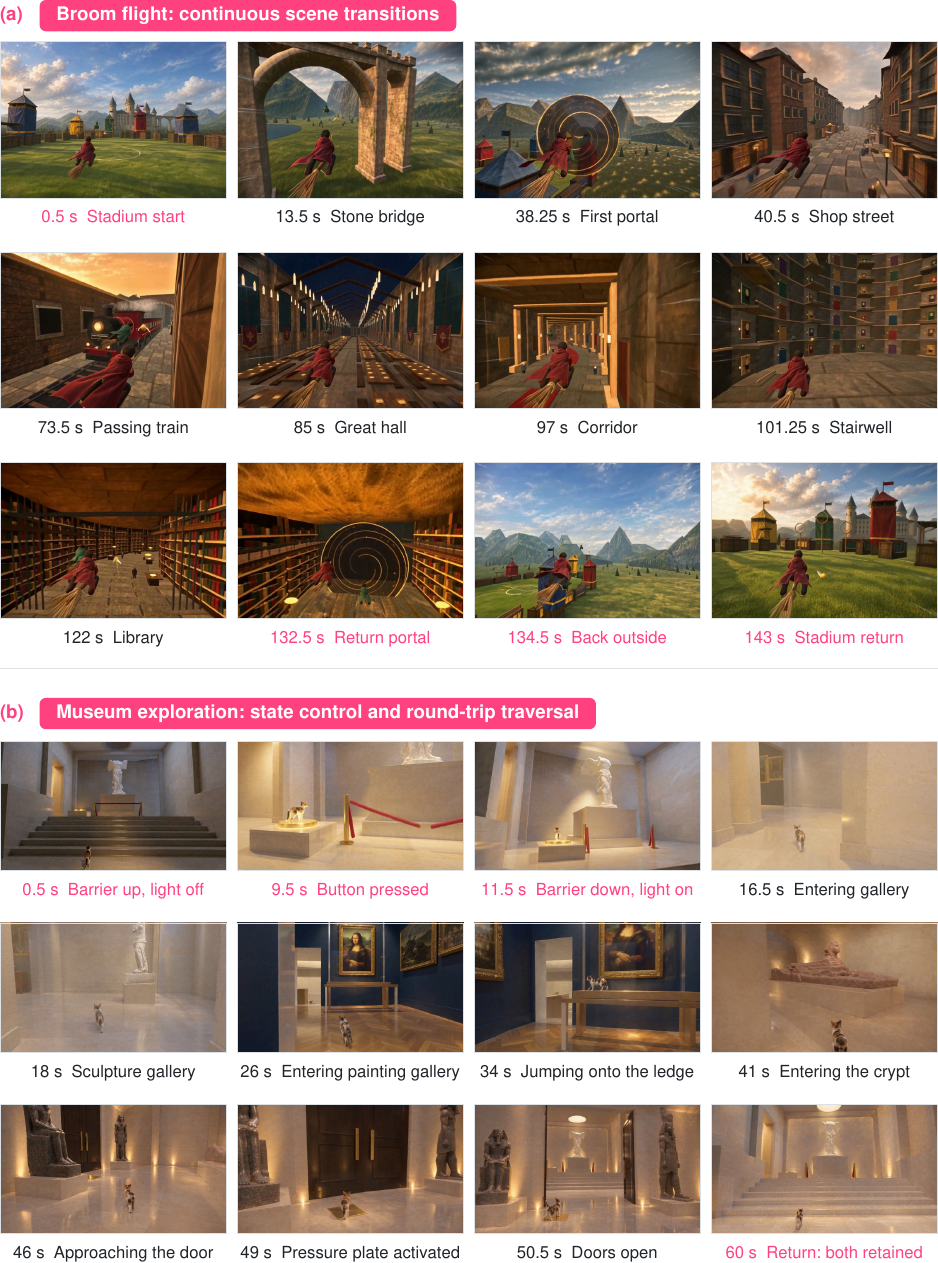}
  \caption{\textbf{Continuous scene transitions and persistent interaction state.} (a) Broom flight across successive outdoor and interior scenes, including portals into the shop street and back to the stadium. (b) Museum exploration through button and pressure-plate interactions. The barrier remains lowered and the hall light remains on when the cat returns. Each panel contains twelve observations, read left to right and top to bottom.}
  \label{fig:renderer-world-sequences}
\end{figure}

\paragraph{World traversal and persistent interaction state.}
\figref{fig:renderer-world-sequences} presents two exploration sequences. \textbf{Broom~flight} in panel (a) follows a broom rider from the stadium, beneath a stone bridge, and through the first portal into a shop street. After passing a train and traversing the great hall, corridor, stairwell, and library, the broom rider enters the final portal and returns to the stadium. This sequence illustrates sustained rendering through successive scene transitions. \textbf{Museum~exploration} in panel (b) shows a cat pressing a button to lower the hall barrier and switch on the overhead light. The cat traverses the galleries and crypt, then steps on a pressure plate to open the return doors. On return, the barrier remains lowered and the light remains on, accounting for the brighter final view. This round trip illustrates state-controlled interactions and the persistence of their visible effects across revisits.

\subsection{Inference Efficiency and Streaming Performance}
\label{sec:system-results}

We evaluate four $832\times480$ serving workloads on one H200 (140\,GB) for the DiT and two H100s, one each for encoding and decoding. The proxy input is configured at 24 FPS. In steady state, the DiT alternates between target blocks of three and two latents and uses two denoising steps per block ($\sigma\in\{1.0,0.5\}$). Each comparison holds hardware, parallel layout, inputs, and the denoising schedule fixed. The serving results are from a warmed run. Session timing runs from request arrival to decoded RGB publication at the serving gateway and excludes subsequent transport and browser presentation.

\paragraph{Streaming throughput and latency.}
\tabref{tab:e2e-stepwise-ablation} measures how the optimizations affect serving performance. Starting from Dense BF16, Dense FP8 adds W8A8 projections in the attention and feed-forward layers and enables fused quantization, while retaining the BF16 token refiner, dense FlashAttention, and an uncompressed KV pool. The full stack then adds three-tier attention and compressed KV. All configurations share offline AdaLN preparation, 50-layer CUDA Graph replay, and the fused decoder.

\begin{table}[!htbp]
  \centering
  \caption{\textbf{Streaming performance and output fidelity across four workloads.} All configurations are evaluated in warmed runs at $832\times480$ with the DiT on one H200. Peak DiT memory is reported by the GPU driver. Fidelity is measured relative to Dense BF16. Bold marks the best result where a comparison is available. Dashes indicate unavailable comparisons.}
  \label{tab:e2e-stepwise-ablation}
  \begingroup
  \small
  \setlength{\tabcolsep}{0pt}
  \renewcommand{\arraystretch}{1.13}
  \setlength{\aboverulesep}{0pt}
  \setlength{\belowrulesep}{0pt}
  \newcommand{\ablationheader}[2]{\begin{tabular}[c]{@{}c@{}}\textbf{#1}\\#2\end{tabular}}
  \begin{tabular*}{\linewidth}{@{\extracolsep{\fill}}lccccccccc@{}}
    \toprule
    \textbf{Configuration} &
    \ablationheader{DiT latency}{(ms/block) $\downarrow$} &
    \ablationheader{Peak memory}{(GiB) $\downarrow$} &
    \ablationheader{KV pool}{(GiB) $\downarrow$} &
    \ablationheader{TTFF}{(s) $\downarrow$} &
    \ablationheader{Total time}{(s) $\downarrow$} &
    \ablationheader{Post-TTFF}{FPS $\uparrow$} &
    \ablationheader{PSNR}{(dB) $\uparrow$} &
    \textbf{SSIM} $\uparrow$ &
    \textbf{LPIPS} $\downarrow$ \\
    \midrule
    \rowcolor[HTML]{EAF3FC}\multicolumn{10}{@{}l@{}}{\textit{Scenario 1: Short sequence (403 frames, 2 references)}} \\
    \quad Dense BF16 & 441.47 & 111.6 & 64.80 & 1.391 & 22.69 & 18.88 & --- & --- & --- \\
    \quad Dense FP8 & 359.17 & \textbf{93.5} & 64.80 & 0.738 & 18.11 & 23.15 & 17.63 & 0.5225 & 0.2623 \\
    \quad Full stack & \textbf{348.77} & 94.5 & \textbf{24.08} & \textbf{0.695} & \textbf{17.58} & \textbf{23.82} & \textbf{18.82} & \textbf{0.5996} & \textbf{0.2089} \\
    \midrule
    \rowcolor[HTML]{EAF3FC}\multicolumn{10}{@{}l@{}}{\textit{Scenario 2: Reference and prompt switching (764 frames, 6 mid-stream switches)}} \\
    \quad Dense BF16 & 457.24 & 112.5 & 64.80 & 2.325 & 45.16 & 17.81 & --- & --- & --- \\
    \quad Dense FP8 & \textbf{360.21} & \textbf{94.2} & 64.80 & 1.956 & 35.93 & \textbf{22.46} & \textbf{24.02} & \textbf{0.7734} & \textbf{0.1449} \\
    \quad Full stack & 362.40 & 94.8 & \textbf{24.08} & \textbf{1.855} & \textbf{35.90} & 22.41 & 23.09 & 0.7445 & 0.1566 \\
    \midrule
    \rowcolor[HTML]{EAF3FC}\multicolumn{10}{@{}l@{}}{\textit{Scenario 3: Long-prompt session (6,247 frames, 4 mid-stream prompt updates)}} \\
    \quad Dense BF16 & 505.57 & 137.1 & 64.80 & 23.30 & 401.13 & 16.53 & --- & --- & --- \\
    \quad Dense FP8 & 417.37 & 119.0 & 64.80 & 10.06 & 323.51 & 19.93 & 15.92 & 0.3161 & 0.2956 \\
    \quad Full stack & \textbf{401.30} & \textbf{117.3} & \textbf{24.08} & \textbf{9.78} & \textbf{312.33} & \textbf{20.64} & \textbf{16.07} & --- & --- \\
    \midrule
    \rowcolor[HTML]{EAF3FC}\multicolumn{10}{@{}l@{}}{\textit{Scenario 4: Long interactive session (15,921 frames, 18 mid-stream reference switches)}} \\
    \quad Dense BF16 & 524.00 & 137.1 & 64.80 & 11.83 & 1,001.55 & 16.09 & --- & --- & --- \\
    \quad Dense FP8 & 436.59 & 119.0 & 64.80 & 11.49 & 837.61 & 19.27 & 21.47 & 0.7344 & 0.2204 \\
    \quad Full stack & \textbf{412.14} & \textbf{117.3} & \textbf{24.08} & \textbf{10.33} & \textbf{790.67} & \textbf{20.40} & \textbf{23.14} & \textbf{0.7710} & \textbf{0.1623} \\
    \bottomrule
  \end{tabular*}
  \endgroup
\end{table}

In the representative short-sequence workload (Scenario~1), post-TTFF throughput rises from $18.88$ to $23.15$ FPS with W8A8 and fused quantization and then to $23.82$ FPS with the full stack, a \textbf{$26\%$ improvement} over Dense BF16. Warmed time to first frame (TTFF) falls from $1.39$ to $0.70$ seconds. The full stack also improves throughput over Dense BF16 across the longer and condition-changing workloads, reaching $20.40$ FPS in the longest session compared with $16.09$ FPS for Dense BF16.

\paragraph{Output fidelity relative to Dense BF16.}
We compare generated streams with Dense BF16 using the same seeds and denoising schedule. \tabref{tab:e2e-stepwise-ablation} reports PSNR, SSIM, and LPIPS~\citep{lpips} as measures of agreement with that reference. The full stack improves PSNR over Dense FP8 in three of the four workloads, including gains of $1.19$ dB in the short sequence and $1.67$ dB in the longest session. The table presents the per-workload fidelity results alongside throughput and memory.

\Needspace{6\baselineskip}
\paragraph{Attention efficiency and KV footprint.}
\tabref{tab:sm90-kernel-kv-microbench} evaluates attention kernels and resident KV storage on H100 GPUs (80\,GB) at $832\times480$ and $1280\times736$. The query/KV lengths are $(780,9{,}360)$ and $(1{,}840,22{,}080)$. The three-tier configuration assigns $22/44/88$ and $48/96/186$ historical tiles to Refine/Skim/Skip at the two resolutions, respectively, while keeping the current block at token granularity.

Frame-aligned three-tier attention achieves approximately \textbf{$2.8\times$ kernel speedup} over Dense BF16 at both resolutions, while compressed KV reduces the resident pool by approximately \textbf{$63\%$}. Both FP8 configurations share the same compressed pool; evaluating selected history tiles at coarser resolutions provides a further $3.31\times$ and $3.24\times$ kernel speedup over token-level FP8 attention at the two resolutions. Attention output cosine similarity remains above $0.9998$ for the tested inputs. In the serving workloads, compressed KV reduces the reported pool from $64.80$ to $24.08$ GiB across all four scenarios.
\begin{table}[H]
  \centering
  \caption{\textbf{Attention kernel latency and resident KV pool size.} Isolated NVIDIA H100 microbenchmarks at two streaming resolutions. The two FP8 configurations share the same compressed KV pool. Cosine similarity is measured against the Dense BF16 attention output.}
  \label{tab:sm90-kernel-kv-microbench}
  \begingroup
  \small
  \setlength{\tabcolsep}{0pt}
  \renewcommand{\arraystretch}{1.13}
  \setlength{\aboverulesep}{0pt}
  \setlength{\belowrulesep}{0pt}
  \newcommand{\kvbenchheader}[2]{\begin{tabular}[c]{@{}c@{}}\textbf{#1}\\#2\end{tabular}}
  \begin{tabular*}{\linewidth}{@{\extracolsep{\fill}}lccccc@{}}
    \toprule
    \textbf{Configuration} &
    \kvbenchheader{Kernel latency}{($\mu$s) $\downarrow$} &
    \kvbenchheader{Speedup}{\textbf{vs.\ BF16} $\uparrow$} &
    \textbf{Cosine similarity} $\uparrow$ &
    \kvbenchheader{Bytes}{per tile per head} &
    \kvbenchheader{KV pool}{(GiB/rank) $\downarrow$} \\
    \midrule
    \rowcolor[HTML]{EAF3FC}\multicolumn{6}{@{}l@{}}{\textit{Resolution: 832\,$\times$\,480 (SP = 1)}} \\
    \quad Dense BF16 & 935.30 & 1.00$\times$ & 1.000000 & 49,152 & 40.83 \\
    \quad FP8 (100\% Refine) & 1,114.53 & 0.84$\times$ & 0.999986 & 18,248 & 15.18 ($-$62.82\%) \\
    \quad \textbf{FP8 (Three-Tier)} & \textbf{337.06} & \textbf{2.77}$\times$ & 0.999865 & 18,248 & 15.18 ($-$62.82\%) \\
    \midrule
    \rowcolor[HTML]{EAF3FC}\multicolumn{6}{@{}l@{}}{\textit{Resolution: 1280\,$\times$\,736 (SP = 2)}} \\
    \quad Dense BF16 & 2,492.16 & 1.00$\times$ & 1.000000 & 49,152 & 39.52 \\
    \quad FP8 (100\% Refine) & 2,837.18 & 0.88$\times$ & 0.999986 & 18,248 & 14.68 ($-$62.85\%) \\
    \quad \textbf{FP8 (Three-Tier)} & \textbf{875.52} & \textbf{2.85}$\times$ & 0.999856 & 18,248 & 14.68 ($-$62.85\%) \\
    \bottomrule
  \end{tabular*}
  \endgroup
\end{table}

\FloatBarrier

\section{Discussion and Limitations}
\label{sec:limitations}

Controllable Rendering makes world-state updates independent of generated video. Several design challenges remain in building playable world systems around this separation. Addressing them will be important for extending this approach to more diverse worlds and forms of play.

\paragraph{Long-horizon rendering fidelity.}
The streaming renderer conditions each block on the current proxy and recent generated history. A deviation in an earlier block can make these signals conflict, with the history favoring continuity with the generated layout or trajectory and the proxy specifying a different one. The model can then struggle to determine which signal to follow when generating the next block. Continuing from the erroneous history can compound the deviation and degrade visual quality over long rollouts, while an abrupt correction toward the proxy can disrupt temporal continuity.

\paragraph{The division of responsibilities between the engine and the renderer.}
Authority over world state does not determine how much of the player's view the engine should prescribe. For locomotion, the engine could supply a trajectory and leave the gait to the renderer, or specify additional details such as body pose and foot contacts. Similarly, a new object could be introduced into the executable world or appear only in the rendered scene. These choices become consequential when a footstep triggers a mechanism or a player expects to pick up a rendered object. Determining where to draw this boundary for different forms of play remains a central design question.

\paragraph{Richer conditioning interfaces.}
Once the system determines which execution details the renderer must preserve, it still needs to decide how to convey those details to the renderer. Depth and semantic maps could supplement RGB proxy video, while entity identifiers and event signals could make particular objects and interactions explicit. Image references could be combined in different ways, such as pairing a scene reference with multiple views of a subject or assigning separate references to several entities. Beyond still images, other reference formats such as video clips could convey changes in appearance over time. Combining these inputs requires maintaining their correspondence to the relevant entities and world updates.

\Needspace{7\baselineskip}
\paragraph{World construction and gameplay experience.}
A central challenge is to broaden the range of executable worlds that coding agents can build while reducing the time needed to author and validate them. Meeting both goals places greater demands on the agent harness and its supporting component libraries. The libraries need to offer a wider range of reusable interactions, such as driving, climbing, and object manipulation, while the harness needs to help agents discover these components, combine them with rules and objectives to create distinct forms of play, and playtest the resulting worlds. Overlays for inventory, progress, and scores could further improve the player experience by presenting engine state alongside the generated view.

\paragraph{Interaction latency.}
Further reducing the time from proxy input to rendered output is important for playability. Latency can determine whether timing-sensitive interactions and game mechanics are viable in practice. Rendering, however, accounts for only part of the delay between a player input and the first displayed frame that shows its effect. World execution, proxy capture, transport, and display also contribute. Shortening the full path while preserving visual quality and the correspondence between displayed frames and world updates remains an end-to-end systems challenge.

\section{Conclusion}
\label{sec:conclusion}

We presented Zing-1.0, a Playable World System built on Controllable Rendering. Its design follows a clear division of responsibilities: \textbf{determinism comes from control, and imagination comes from generation.} We formalized this division with an explicit criterion and specified four responsibilities and their interfaces. For \emph{World Construction}, a coding agent authored an executable world with WorldKit. During \emph{World Execution}, the engine applied the rules, computed all scores from its state, and supplied live proxy video throughout a full round of play. The renderer used that stream for \emph{Observation Synthesis}, and the \emph{Interaction Runtime} delivered the synthesized view to the player. In benchmark comparisons, the renderer leads the streaming editors evaluated on OpenVE and LongV2VBench, while the full editing system outperforms the reported baselines on both StreamEval editing tracks. Pipelined serving and DiT inference optimizations enable Zing-1.0 to sustain 23.82 FPS at $832\times480$ in steady state, using a three-GPU deployment comprising a single H200 for the DiT and one H100 each for the encoding and decoding services.

\clearpage
\section*{Contributions and Acknowledgments}

Zing-1.0 was developed by the SeedLeap.ai team. We thank the MiniMax-H3 team, the authors of the training and distillation methods used in this work, and the SGLang, OpenVE-Bench, LongV2VBench, and StreamEval teams for making their research and implementations available.

We are especially grateful to Loopit's engineering, product, and operations teams for their substantial support in preparing and launching Zing-1.0, including improving the user experience, showcasing model capabilities, and selecting demos.

\subsection*{Contributors}

Contributors are listed in reverse alphabetical order by surname, not in order of contribution.

Wenbo Zhang, Guoli Zhang, Xuewen Yu, Xueliang Yang, Jiangqiao Yan, Teng Xia, Xinmiao Wu, Zeyang Tian, Haoze Sun, Jinghui Niu, Hongli Ning, Zhiyuan Liu, Xinna Liu, Weimin Liu, Jialin Liu, Tianpeng Li, Kejun Li, Jun Li, Jiawen Li, Yirong He, Haoyuan Guo, Bosheng Gong, Junhao Gao, Xiaoxiao Fu, Xuezhi Chen, Shengdong Chen, Mingyang Chen, Diying Chen, Yinglun Chai, Yiwen Cai.

\clearpage
\begingroup
\small
\setlength{\bibsep}{2pt}
\bibliographystyle{abbrvnat}
\bibliography{refs}

\begin{thebibliography}{80}
\providecommand{\natexlab}[1]{#1}
\providecommand{\url}[1]{\texttt{#1}}
\expandafter\ifx\csname urlstyle\endcsname\relax
  \providecommand{\doi}[1]{doi: #1}\else
  \providecommand{\doi}{doi: \begingroup \urlstyle{rm}\Url}\fi

\bibitem[{Alibaba Cloud Community}(2026)]{alibaba2026happyoyster}
{Alibaba Cloud Community}.
\newblock {Alibaba Launches HappyOyster, a World Model Product for Real-Time
  Immersive Creation and Interaction}.
\newblock
  \href{https://www.alibabacloud.com/blog/alibaba-launches-happyoyster-a-world-model-product-for-real-time-immersive-creation-and-interaction_603048}{Official
  announcement}, Apr. 2026.

\bibitem[{Anthropic}(2024)]{mcp}
{Anthropic}.
\newblock {Model Context Protocol Specification}.
\newblock \href{https://modelcontextprotocol.io/}{modelcontextprotocol.io},
  2024.

\bibitem[Charbonnier et~al.(1997)Charbonnier, Blanc-F{\'{e}}raud, Aubert, and
  Barlaud]{charbonnier1997deterministic}
P.~Charbonnier, L.~Blanc-F{\'{e}}raud, G.~Aubert, and M.~Barlaud.
\newblock {Deterministic Edge-Preserving Regularization in Computed Imaging}.
\newblock \emph{IEEE Transactions on Image Processing}, 6\penalty0
  (2):\penalty0 298--311, 1997.
\newblock \doi{10.1109/83.551699}.

\bibitem[Chen et~al.(2026{\natexlab{a}})Chen, Chen, Fu, Gong, Guo, Li, Li, Li,
  Li, Liu, Sun, Tian, Wang, Wu, Yan, and Zhao]{chen2026zing05}
M.~Chen, S.~Chen, X.~Fu, B.~Gong, H.~Guo, B.~Li, J.~Li, K.~Li, T.~Li, Y.~Liu,
  H.~Sun, Z.~Tian, M.~Wang, X.~Wu, J.~Yan, and Z.~Zhao.
\newblock {Zing-0.5: Toward Playable Worlds with Real-Time Joint Action and
  Text Control}.
\newblock \emph{arXiv preprint
  \href{https://arxiv.org/abs/2609.17909}{arXiv:2609.17909}},
  2026{\natexlab{a}}.

\bibitem[Chen et~al.(2026{\natexlab{b}})Chen, Cheng, Deng, Hang, Hou, Li, Liu,
  Lu, Tang, Tang, Teng, Tian, Wang, Wang, Wang, Wu, Yang, Yu, Zhao, Cao, Chen,
  Duan, Han, Han, Jiang, Li, Liang, Ma, Wang, Wang, Wu, Xing, Ying, Yu, Zhang,
  Zhou, Zhou, and Zou]{chen2026hynomas}
Q.~Chen, J.~Cheng, X.~Deng, T.~Hang, J.~Hou, J.~Li, Y.~Liu, T.~Lu, J.~Tang,
  Y.~Tang, Y.~Teng, Y.~Tian, H.~Wang, W.~Wang, X.~Wang, X.~Wu, M.~Yang, Y.~Yu,
  R.~Zhao, L.~Cao, L.~Chen, H.~Duan, J.~Han, M.~Han, E.~Jiang, S.~Li, Y.~Liang,
  Q.~Ma, C.~Wang, L.~Wang, B.~Wu, K.~Xing, R.~Ying, Z.~Yu, K.~Zhang, Y.~Zhou,
  Z.~Zhou, and K.~Zou.
\newblock {Hy-Nomas: Towards Free-form Interactive World Generation}.
\newblock \emph{arXiv preprint
  \href{https://arxiv.org/abs/2610.36729}{arXiv:2610.36729}},
  2026{\natexlab{b}}.

\bibitem[Chen et~al.(2026{\natexlab{c}})Chen, Liu, Jia, Wang, Ling, Qu, and
  Gao]{dfd}
S.~Chen, S.~Liu, Y.~Jia, Z.~Wang, H.~Ling, Q.~Qu, and J.~Gao.
\newblock {Data-Forcing Distillation: Restoring Diversity and Fidelity in
  Few-Step Video Generation}.
\newblock \href{https://arxiv.org/abs/2606.18478}{arXiv:2606.18478},
  2026{\natexlab{c}}.

\bibitem[Chen et~al.(2026{\natexlab{d}})Chen, Lin, and
  Zhang]{chen2026codeworldmodel}
Y.~Chen, G.~Lin, and C.~Zhang.
\newblock {Code World Model: Coding Agent as World Brain}.
\newblock \emph{arXiv preprint
  \href{https://arxiv.org/abs/2608.25927}{arXiv:2608.25927}},
  2026{\natexlab{d}}.

\bibitem[Chi et~al.(2026)Chi, Miao, Shi, Wu, Wang, Chen, Dai, Li, Ren, Gao,
  Long, Duan, Lyu, Wu, and Liu]{chi2026agentgarten}
J.~Chi, S.~Miao, Z.~Shi, K.~Wu, H.~Wang, W.~Chen, Q.~Dai, C.~Li, J.~Ren,
  J.~Gao, M.~Long, Y.~Duan, J.~Lyu, J.~Wu, and F.~Liu.
\newblock {AgentGarten: Code Worlds for Evolving Agents}.
\newblock \emph{arXiv preprint
  \href{https://arxiv.org/abs/2610.12374}{arXiv:2610.12374}}, 2026.

\bibitem[{Decart}(2026)]{decart2026lucy25}
{Decart}.
\newblock {Lucy 2.5: Raising the Bar for Live AI}.
\newblock
  \href{https://decart.ai/publications/lucy-2-5-raising-the-bar-for-live-ai}{Official
  release}, July 2026.

\bibitem[{Decart} and {Etched}(2024)]{decart2024oasis}
{Decart} and {Etched}.
\newblock {Oasis: A Universe in a Transformer}.
\newblock \href{https://oasis-model.github.io/}{Project release}, 2024.

\bibitem[{Dimforge}(n.d.)]{rapier}
{Dimforge}.
\newblock {Rapier: Fast 2D and 3D Physics Engines}.
\newblock \href{https://rapier.rs/}{rapier.rs}, n.d.

\bibitem[Ding et~al.(2026)Ding, Chen, Cai, Xu, Wang, Lu, Li, Chen, Gao, Tao,
  Wan, and Zhao]{ding2026playworld}
K.~Ding, X.~Chen, M.~Cai, Z.~Xu, Y.~Wang, Y.~Lu, J.~Li, S.~Chen, Y.~Gao,
  X.~Tao, P.~Wan, and H.~Zhao.
\newblock {PlayWorld: Benchmarking World Models with Agent Players over
  Long-Horizon Objectives}.
\newblock \emph{arXiv preprint
  \href{https://arxiv.org/abs/2608.13552}{arXiv:2608.13552}}, 2026.

\bibitem[{Epic Games}(n.d.)]{epicunrealgameplay}
{Epic Games}.
\newblock {Gameplay Framework in Unreal Engine}.
\newblock
  \href{https://dev.epicgames.com/documentation/unreal-engine/gameplay-framework-in-unreal-engine}{Unreal
  Engine documentation}, n.d.

\bibitem[Fang et~al.(2026{\natexlab{a}})Fang, Qian, Lu, Huang, Jiang, Wang,
  Song, Jiang, and Shou]{fang2026solarrenderer}
G.~Fang, S.~Qian, D.~Lu, J.~Huang, L.~Jiang, X.~Wang, Y.~Song, L.~Jiang, and
  M.~Z. Shou.
\newblock {SolarRenderer: Few-Step Generative Reference Rendering}.
\newblock \emph{arXiv preprint
  \href{https://arxiv.org/abs/2610.28630}{arXiv:2610.28630}},
  2026{\natexlab{a}}.

\bibitem[Fang et~al.(2026{\natexlab{b}})Fang, Shao, Zhu, Xiao, Chen, Liu, Fu,
  Cao, Zhai, and Zha]{fang2026coder}
Z.~Fang, Y.~Shao, K.~Zhu, J.~Xiao, S.~Chen, Y.~Liu, X.~Fu, Y.~Cao, W.~Zhai, and
  Z.-J. Zha.
\newblock {Code Plans, Diffusion Renders: Open-Ended Generative World
  Modeling}.
\newblock \emph{arXiv preprint
  \href{https://arxiv.org/abs/2609.26458}{arXiv:2609.26458}},
  2026{\natexlab{b}}.

\bibitem[Feng et~al.(2025)Feng, Li, Yang, Xi, Li, Li, Zhang, Yang, Peng, Han,
  Agrawala, Keutzer, Kodaira, and Xu]{feng2025streamdiffusionv2}
T.~Feng, Z.~Li, S.~Yang, H.~Xi, M.~Li, X.~Li, L.~Zhang, K.~Yang, K.~Peng,
  S.~Han, M.~Agrawala, K.~Keutzer, A.~Kodaira, and C.~Xu.
\newblock {StreamDiffusionV2: A Streaming System for Dynamic and Interactive
  Video Generation}.
\newblock \emph{arXiv preprint
  \href{https://arxiv.org/abs/2511.07399}{arXiv:2511.07399}}, 2025.

\bibitem[Gao et~al.(2026)Gao, Wang, Zhu, Chen, Liu, Bai, Wang, Yuan, Wang, Lu,
  Cheng, Zhang, Gao, Feng, Liu, Yao, Xu, Zhu, Shen, and
  Ouyang]{gao2026lingbotinfinity}
Z.~Gao, Q.~Wang, J.~Zhu, J.~Chen, Z.~Liu, Q.~Bai, J.~Wang, Y.~Yuan, H.~Wang,
  Y.~Lu, K.~L. Cheng, H.~Zhang, J.~Gao, T.~Feng, Y.~Liu, Y.~Yao, Y.~Xu, X.~Zhu,
  Y.~Shen, and H.~Ouyang.
\newblock {Infinite Worlds with Versatile Interactions}.
\newblock \emph{arXiv preprint
  \href{https://arxiv.org/abs/2607.07534}{arXiv:2607.07534}}, 2026.

\bibitem[Guo et~al.(2026)Guo, Li, Li, and Huang]{guo2026worldclaw}
C.~Guo, J.~Li, Y.~Li, and Z.~Huang.
\newblock {WorldClaw: Agentic 3D Open-World Generation at Scale}.
\newblock \emph{arXiv preprint
  \href{https://arxiv.org/abs/2608.05248}{arXiv:2608.05248}}, 2026.

\bibitem[Guo et~al.(2025)Guo, Yang, He, Zhao, Wei, Yang, Huang, and
  Lin]{resamplingforcing}
Y.~Guo, C.~Yang, H.~He, Y.~Zhao, M.~Wei, Z.~Yang, W.~Huang, and D.~Lin.
\newblock {End-to-End Training for Autoregressive Video Diffusion via
  Self-Resampling}.
\newblock \href{https://arxiv.org/abs/2512.15702}{arXiv:2512.15702}, 2025.

\bibitem[Hadsell et~al.(2026)Hadsell, Rivas, and Qi]{hadsell2026genie4}
R.~Hadsell, D.~Rivas, and H.~Qi.
\newblock {Genie 4: A New Generation of World Simulation}.
\newblock
  \href{https://deepmind.google/blog/genie-4-a-new-generation-of-world-simulation/}{Google
  DeepMind research blog}, Oct. 2026.

\bibitem[He et~al.(2025{\natexlab{a}})He, Wang, Zhang, Xue, Bu, Yang, Wen, and
  Xie]{he2025openve}
H.~He, J.~Wang, J.~Zhang, Z.~Xue, X.~Bu, Q.~Yang, S.~Wen, and L.~Xie.
\newblock {OpenVE-3M: A Large-Scale High-Quality Dataset for Instruction-Guided
  Video Editing}.
\newblock \emph{arXiv preprint
  \href{https://arxiv.org/abs/2512.07826}{arXiv:2512.07826}},
  2025{\natexlab{a}}.

\bibitem[He et~al.(2025{\natexlab{b}})He, Yang, Lin, Xu, Wei, Gui, Zhao,
  Wetzstein, Jiang, and Li]{he2025cameractrlii}
H.~He, C.~Yang, S.~Lin, Y.~Xu, M.~Wei, L.~Gui, Q.~Zhao, G.~Wetzstein, L.~Jiang,
  and H.~Li.
\newblock {CameraCtrl II: Dynamic Scene Exploration via Camera-controlled Video
  Diffusion Models}.
\newblock In
  \emph{\href{https://openaccess.thecvf.com/content/ICCV2025/html/He_CameraCtrl_II_Dynamic_Scene_Exploration_via_Camera-controlled_Video_Diffusion_Models_ICCV_2025_paper.html}{IEEE/CVF
  International Conference on Computer Vision}}, pages 13416--13426,
  2025{\natexlab{b}}.

\bibitem[He et~al.(2026)He, Peng, Liu, Wang, Zhang, Cui, Kang, Jiang, An, Ren,
  Xu, Guo, Gong, Wu, Li, Song, Liu, Li, and Zhou]{he2026matrixgame2}
X.~He, C.~Peng, Z.~Liu, B.~Wang, Y.~Zhang, Q.~Cui, F.~Kang, B.~Jiang, M.~An,
  Y.~Ren, B.~Xu, H.-X. Guo, K.~Gong, S.~Wu, W.~Li, X.~Song, Y.~Liu, Y.~Li, and
  Y.~Zhou.
\newblock {Matrix-Game 2.0: An Open-Source, Real-Time, and Streaming
  Interactive World Model}.
\newblock \emph{arXiv preprint
  \href{https://arxiv.org/abs/2508.13009v5}{arXiv:2508.13009}}, 2026.

\bibitem[Ho and Salimans(2022)]{cfg}
J.~Ho and T.~Salimans.
\newblock {Classifier-Free Diffusion Guidance}.
\newblock \href{https://arxiv.org/abs/2207.12598}{arXiv:2207.12598}, 2022.

\bibitem[Hu et~al.(2025)Hu, Lai, Mitsufuji, and Ermon]{cmt}
Z.~Hu, C.-H. Lai, Y.~Mitsufuji, and S.~Ermon.
\newblock {CMT: Mid-Training for Efficient Learning of Consistency, Mean Flow,
  and Flow Map Models}.
\newblock \href{https://arxiv.org/abs/2509.24526}{arXiv:2509.24526}, 2025.

\bibitem[Huang et~al.(2026{\natexlab{a}})Huang, Wang, Li, Jiang, and
  Wu]{huang2026disco}
H.~Huang, J.~Wang, Q.~Li, Y.-G. Jiang, and Z.~Wu.
\newblock {DisCo: World Models with Discrete Camera Motion Control}.
\newblock \emph{arXiv preprint
  \href{https://arxiv.org/abs/2606.07967}{arXiv:2606.07967}},
  2026{\natexlab{a}}.

\bibitem[Huang et~al.(2025)Huang, Li, He, Zhou, and
  Shechtman]{huang2025selfforcing}
X.~Huang, Z.~Li, G.~He, M.~Zhou, and E.~Shechtman.
\newblock {Self Forcing: Bridging the Train-Test Gap in Autoregressive Video
  Diffusion}.
\newblock In \emph{Advances in Neural Information Processing Systems},
  volume~38, 2025.
\newblock \doi{10.52202/085713-5576}.

\bibitem[Huang et~al.(2026{\natexlab{b}})Huang, Xu, Peng, Zhang, Zheng, Song,
  Gu, Yu, Wang, and Cao]{edenlive2026}
Z.~Huang, T.~Xu, C.~Peng, C.~Zhang, Z.~Zheng, Y.~Song, Z.~Gu, Z.~Yu, P.~Wang,
  and Y.~Cao.
\newblock {EdenLive: Real-Time Generative Rendering from Any Visual Reference}.
\newblock \href{https://vast-eden.ai/?p=edenlive}{Technical report and project
  release}, 2026{\natexlab{b}}.

\bibitem[Huang et~al.(2026{\natexlab{c}})Huang, Lin, Lin, Huang, Yu, Niu, Yang,
  Liu, Chuang, Zhang, and Wang]{huang2026programmableworldmodel}
Z.-H. Huang, G.~Lin, J.~Lin, Y.-C. Huang, R.~Yu, M.~Niu, S.~Yang, Y.-L. Liu,
  Y.-Y. Chuang, K.~Zhang, and Z.~Wang.
\newblock {Programmable World Model}.
\newblock \emph{arXiv preprint
  \href{https://arxiv.org/abs/2609.10540}{arXiv:2609.10540}},
  2026{\natexlab{c}}.

\bibitem[Kaelbling et~al.(1998)Kaelbling, Littman, and
  Cassandra]{kaelbling1998pomdp}
L.~P. Kaelbling, M.~L. Littman, and A.~R. Cassandra.
\newblock {Planning and Acting in Partially Observable Stochastic Domains}.
\newblock
  \emph{\href{https://www.sciencedirect.com/science/article/pii/S000437029800023X}{Artificial
  Intelligence}}, 101\penalty0 (1--2):\penalty0 99--134, 1998.

\bibitem[{Kuaishou Technology}(2025)]{kuaishou2025klingo1}
{Kuaishou Technology}.
\newblock {Kling O1 Launches as the World's First Unified Multimodal Video
  Model}.
\newblock
  \href{https://ir.kuaishou.com/news-releases/news-release-details/kling-o1-launches-worlds-first-unified-multimodal-video-model-0/}{Official
  announcement}, Dec. 2025.

\bibitem[{Kuaishou Technology}(2026)]{kuaishou2026kling3}
{Kuaishou Technology}.
\newblock {Kling AI Launches 3.0 Model, Ushering in an Era Where Everyone Can
  Be a Director}.
\newblock
  \href{https://ir.kuaishou.com/news-releases/news-release-details/kling-ai-launches-30-model-ushering-era-where-everyone-can-be}{Official
  announcement}, Feb. 2026.

\bibitem[Li et~al.(2025)Li, Tang, Xu, Wu, Zhou, Shao, Yu, Cao, and
  Lu]{li2025gamecraft}
J.~Li, J.~Tang, Z.~Xu, L.~Wu, Y.~Zhou, S.~Shao, T.~Yu, Z.~Cao, and Q.~Lu.
\newblock {Hunyuan-GameCraft: High-dynamic Interactive Game Video Generation
  with Hybrid History Condition}.
\newblock \emph{arXiv preprint
  \href{https://arxiv.org/abs/2506.17201}{arXiv:2506.17201}}, 2025.

\bibitem[Li et~al.(2026)Li, Pun, Jiang, Li, and Cun]{li2026editalive}
Z.~Li, C.-M. Pun, P.-T. Jiang, B.~Li, and X.~Cun.
\newblock {EditaLive! Unified Character Video Editing for Live Streaming}.
\newblock \emph{arXiv preprint
  \href{https://arxiv.org/abs/2608.27123}{arXiv:2608.27123}}, 2026.

\bibitem[Lin et~al.(2026{\natexlab{a}})Lin, Huang, Yang, Yang, Zhang, and
  Wang]{lin2026alayaflash}
G.~Lin, Z.-H. Huang, S.~Yang, M.-H. Yang, K.~Zhang, and Z.~Wang.
\newblock {Generative World Renderer at the Speed of Play}.
\newblock \emph{arXiv preprint
  \href{https://arxiv.org/abs/2607.18703}{arXiv:2607.18703}},
  2026{\natexlab{a}}.

\bibitem[Lin et~al.(2026{\natexlab{b}})Lin, Deng, Wu, Wen, and
  Jin]{lin2026gamedirector}
Z.~Lin, Z.~Deng, Y.~Wu, B.~Wen, and Y.~Jin.
\newblock {GameDirector: Decoupling Gameplay Logic from Rendering for
  Player-Configurable Game World Models}.
\newblock \emph{arXiv preprint
  \href{https://arxiv.org/abs/2609.25652}{arXiv:2609.25652}},
  2026{\natexlab{b}}.

\bibitem[Lin et~al.(2026{\natexlab{c}})Lin, Wang, Tan, Wen, and
  Jin]{lin2026stateplay}
Z.~Lin, Z.~Wang, C.~Tan, B.~Wen, and Y.~Jin.
\newblock {StatePlay: State-Aware Game World Models for Mechanics-Consistent
  Generation}.
\newblock \emph{arXiv preprint
  \href{https://arxiv.org/abs/2607.26754}{arXiv:2607.26754}},
  2026{\natexlab{c}}.

\bibitem[Lipman et~al.(2022)Lipman, Chen, Ben-Hamu, Nickel, and
  Le]{flowmatching}
Y.~Lipman, R.~T.~Q. Chen, H.~Ben-Hamu, M.~Nickel, and M.~Le.
\newblock {Flow Matching for Generative Modeling}.
\newblock \emph{arXiv preprint
  \href{https://arxiv.org/abs/2210.02747v2}{arXiv:2210.02747}}, 2022.

\bibitem[Liu et~al.(2026{\natexlab{a}})Liu, Chen, Li, Chi, Sun, Li, Fu, Guo,
  Wu, Bai, and Yuan]{liu2026bernini}
C.~Liu, J.~Chen, L.~Li, L.~Chi, M.~Sun, Z.~Li, Y.~Fu, R.~Guo, Y.~Wu, G.~Bai,
  and Z.~Yuan.
\newblock {Bernini: Latent Semantic Planning for Video Diffusion}.
\newblock \emph{arXiv preprint
  \href{https://arxiv.org/abs/2605.22344}{arXiv:2605.22344}},
  2026{\natexlab{a}}.

\bibitem[Liu et~al.(2025)Liu, Gao, Liu, Du, Li, Wu, et~al.]{decoupleddmd}
D.~Liu, P.~Gao, D.~Liu, R.~Du, Z.~Li, Q.~Wu, et~al.
\newblock {Decoupled DMD: CFG Augmentation as the Spear, Distribution Matching
  as the Shield}.
\newblock \href{https://arxiv.org/abs/2511.22677}{arXiv:2511.22677}, 2025.

\bibitem[Liu et~al.(2026{\natexlab{b}})Liu, Ouyang, Lan, Wang, and
  Zhang]{liu2026anygamerenderer}
R.~Liu, L.~Ouyang, F.~Lan, Z.~Wang, and K.~Zhang.
\newblock {AnyGameRenderer: Entity-Consistent Generative Rendering Across Games
  and an Entity-Level Benchmark}.
\newblock \emph{arXiv preprint
  \href{https://arxiv.org/abs/2610.31466}{arXiv:2610.31466}},
  2026{\natexlab{b}}.

\bibitem[Lu et~al.(2026)Lu, Ye, Li, Xue, Liu, Han, and
  Xie]{lu2026streamrenderh3}
Y.~Lu, T.~Ye, Y.~Li, S.~Xue, H.~Liu, S.~Han, and E.~Xie.
\newblock {StreamRender-H3: Interactive Coding Worlds with Real-time Streaming
  Diffusion Render}.
\newblock
  \href{https://nvlabs.github.io/Sana/Sol-Engine/StreamRender-H3/}{Project
  page}, 2026.

\bibitem[Mao et~al.(2026)Mao, Lin, Rui, Zhang, Wan, and
  Ma]{mao2026lynnrealomni}
X.~Mao, P.~Lin, S.~Rui, Y.~Zhang, H.~Wan, and W.~Ma.
\newblock {LynnReal-Omni: Native multi-modal Video Generation for Agentic
  Visual Workflows}.
\newblock \emph{arXiv preprint
  \href{https://arxiv.org/abs/2609.15863}{arXiv:2609.15863}}, 2026.

\bibitem[{MiniMax}(2026)]{minimax2026h3}
{MiniMax}.
\newblock {MiniMax-H3}.
\newblock \href{https://huggingface.co/MiniMaxAI/MiniMax-H3}{Official model
  card}, 2026.

\bibitem[{Musubi Tuner Contributors}(n.d.)]{musubiH3Guidance}
{Musubi Tuner Contributors}.
\newblock {MiniMax-H3 Advanced: Guidance Loss}.
\newblock
  \url{https://github.com/kohya-ss/musubi-tuner/blob/main/docs/minimax_h3_advanced.md},
  n.d.
\newblock Software documentation, Guidance Loss section.

\bibitem[Nam et~al.(2026)Nam, Hong, Huang, Liu, Lee, Kim, Jin, Lee, Jung, Choi,
  Kim, and Zhou]{nam2026worldcam}
J.~Nam, Y.~Hong, C.-H.~P. Huang, F.~Liu, J.~Lee, J.~Kim, S.~Jin, Y.~Lee,
  J.~Jung, S.~Choi, S.~Kim, and Y.~Zhou.
\newblock {WorldCam: Interactive Autoregressive 3D Gaming Worlds with Camera
  Pose as a Unifying Geometric Representation}.
\newblock \emph{arXiv preprint
  \href{https://arxiv.org/abs/2603.16871}{arXiv:2603.16871}}, 2026.

\bibitem[Parker-Holder and Fruchter(2025)]{parkerholder2025genie3}
J.~Parker-Holder and S.~Fruchter.
\newblock {Genie 3: A New Frontier for World Models}.
\newblock
  \href{https://deepmind.google/blog/genie-3-a-new-frontier-for-world-models/}{Google
  DeepMind research blog}, Aug. 2025.

\bibitem[{Robbyant Team} et~al.(2026){Robbyant Team}, Gao, Wang, Zeng, Zhu,
  Cheng, Li, Wang, Xu, Ma, Chen, Liu, Cheng, Yao, Zhu, Meng, Zheng, Bai, Chen,
  Shen, Yu, Zhu, Shen, and Ouyang]{robbyant2026lingbotworld}
{Robbyant Team}, Z.~Gao, Q.~Wang, Y.~Zeng, J.~Zhu, K.~L. Cheng, Y.~Li, H.~Wang,
  Y.~Xu, S.~Ma, Y.~Chen, J.~Liu, Y.~Cheng, Y.~Yao, J.~Zhu, Y.~Meng, K.~Zheng,
  Q.~Bai, J.~Chen, Z.~Shen, Y.~Yu, X.~Zhu, Y.~Shen, and H.~Ouyang.
\newblock {Advancing Open-source World Models}.
\newblock \emph{arXiv preprint
  \href{https://arxiv.org/abs/2601.20540}{arXiv:2601.20540}}, 2026.

\bibitem[{StreamEval Team}(2026{\natexlab{a}})]{streameval2026}
{StreamEval Team}.
\newblock {StreamEval: Evaluating Interactive Streaming Video Generation
  Models}.
\newblock \emph{arXiv preprint
  \href{https://arxiv.org/abs/2610.28402}{arXiv:2610.28402}},
  2026{\natexlab{a}}.

\bibitem[{StreamEval Team}(2026{\natexlab{b}})]{streamevalleaderboard}
{StreamEval Team}.
\newblock {StreamEval Leaderboard}.
\newblock \url{https://streameval.ai/}, 2026{\natexlab{b}}.
\newblock Accessed November 11, 2026.

\bibitem[Sun et~al.(2026)Sun, Zhang, Wang, Wu, Wang, Wang, Wang, Zhang, Wang,
  and Guo]{sun2026worldplay}
W.~Sun, H.~Zhang, H.~Wang, J.~Wu, Z.~Wang, Z.~Wang, Y.~Wang, J.~Zhang, T.~Wang,
  and C.~Guo.
\newblock {WorldPlay: Towards Long-Term Geometric Consistency for Real-Time
  Interactive World Modeling}.
\newblock In
  \emph{\href{https://proceedings.mlr.press/v306/sun26az.html}{International
  Conference on Machine Learning}}, volume 306, pages 117501--117523. PMLR,
  2026.

\bibitem[Tang et~al.(2025)Tang, Liu, Li, Wu, Yang, Zhao, Gong, Yuan, Shao,
  Zhang, and Lu]{tang2025gamecraft2}
J.~Tang, J.~Liu, J.~Li, L.~Wu, H.~Yang, P.~Zhao, S.~Gong, X.~Yuan, S.~Shao,
  L.~Zhang, and Q.~Lu.
\newblock {Hunyuan-GameCraft-2: Instruction-following Interactive Game World
  Model}.
\newblock \emph{arXiv preprint
  \href{https://arxiv.org/abs/2511.23429}{arXiv:2511.23429}}, 2025.

\bibitem[{Tencent Multimodal RL Team} et~al.(2026){Tencent Multimodal RL Team},
  Zhou, Tang, Ma, Yang, Yao, Niu, Guo, Zhao, Yao, Jian, Li, Wu, Huang, Liu,
  Sun, Li, and Pang]{tencent2026gca}
{Tencent Multimodal RL Team}, X.~Zhou, J.~Tang, K.~Ma, W.~Yang, J.~Yao, B.~Niu,
  H.~Guo, Y.~Zhao, H.~Yao, P.~Jian, R.~Li, H.~Wu, Y.~Huang, X.~Liu, Y.~Sun,
  K.~Li, and T.~Pang.
\newblock {Game-Creation-Agents: Autonomous Game Production}.
\newblock \emph{arXiv preprint
  \href{https://arxiv.org/abs/2610.16776}{arXiv:2610.16776}}, 2026.

\bibitem[{three.js authors}(n.d.)]{threejs}
{three.js authors}.
\newblock {three.js: JavaScript 3D Library}.
\newblock \href{https://threejs.org/}{threejs.org}, n.d.

\bibitem[Valevski et~al.(2025)Valevski, Leviathan, Arar, and
  Fruchter]{valevski2025gamengen}
D.~Valevski, Y.~Leviathan, M.~Arar, and S.~Fruchter.
\newblock {Diffusion Models Are Real-Time Game Engines}.
\newblock In
  \emph{\href{https://openreview.net/forum?id=P8pqeEkn1H}{International
  Conference on Learning Representations}}, 2025.

\bibitem[{VAST AI Research}(2026)]{vast2026projecteden}
{VAST AI Research}.
\newblock {Project Eden}.
\newblock \url{https://www.tripo3d.ai/research/project-eden}, May 2026.
\newblock Research preview.

\bibitem[Wang et~al.(2026{\natexlab{a}})Wang, Zhao, Zhan, and
  Ma]{wang2026liveedit}
X.~Wang, C.~Zhao, F.~Zhan, and Y.~Ma.
\newblock {LiveEdit: Towards Real-Time Diffusion-Based Streaming Video
  Editing}.
\newblock \emph{arXiv preprint
  \href{https://arxiv.org/abs/2606.26740}{arXiv:2606.26740}},
  2026{\natexlab{a}}.

\bibitem[Wang et~al.(2026{\natexlab{b}})Wang, Liu, Li, Huang, Xu, Kang, An,
  Wang, Jiang, Wei, Xietian, Pei, Hu, Jiang, Xue, Wang, Sun, Li, Ouyang, He,
  Liu, Li, and Zhou]{wang2026matrixgame3}
Z.~Wang, Z.~Liu, J.~Li, K.~Huang, B.~Xu, F.~Kang, M.~An, P.~Wang, B.~Jiang,
  Y.~Wei, Y.~Xietian, J.~Pei, L.~Hu, B.~Jiang, H.~Xue, Z.~Wang, H.~Sun, W.~Li,
  W.~Ouyang, X.~He, Y.~Liu, Y.~Li, and Y.~Zhou.
\newblock {Matrix-Game 3.0: Real-Time and Streaming Interactive World Model
  with Long-Horizon Memory}.
\newblock \emph{arXiv preprint
  \href{https://arxiv.org/abs/2604.08995v3}{arXiv:2604.08995}},
  2026{\natexlab{b}}.

\bibitem[{World Labs}(2025)]{worldlabs2025marble}
{World Labs}.
\newblock {Marble: A Multimodal World Model}.
\newblock \href{https://worldlabs.ai/blog/marble-world-model}{Model
  announcement}, Nov. 2025.

\bibitem[{World Labs}(2026)]{worldlabs2026taxonomy}
{World Labs}.
\newblock {A Functional Taxonomy of World Models}.
\newblock \href{https://worldlabs.ai/blog/taxonomy-of-world-models}{Research
  essay}, June 2026.

\bibitem[{World Labs Team}(2026)]{worldlabs2026atlas}
{World Labs Team}.
\newblock {Atlas: A World Model for Spatial Intelligence}.
\newblock \emph{\href{https://www.worldlabs.ai/blog/atlas}{World Labs Blog}},
  2026.

\bibitem[Wu et~al.(2026{\natexlab{a}})Wu, Xu, Zhang, Zhao, and
  Tang]{wu2026code2games}
W.~Wu, Z.~Xu, Z.~Zhang, Y.~Zhao, and H.~Tang.
\newblock {Code2Games: Enabling Coding Agents for Gaming World Generation}.
\newblock \emph{arXiv preprint
  \href{https://arxiv.org/abs/2610.05033}{arXiv:2610.05033}},
  2026{\natexlab{a}}.

\bibitem[Wu et~al.(2026{\natexlab{b}})Wu, Lin, and Jin]{wu2026wizard}
Y.~Wu, Z.~Lin, and Y.~Jin.
\newblock {Wizard: Manipulate Any World in Real Time via Agentic Control}.
\newblock \emph{arXiv preprint
  \href{https://arxiv.org/abs/2610.40339}{arXiv:2610.40339}},
  2026{\natexlab{b}}.

\bibitem[Wu et~al.(2026{\natexlab{c}})Wu, Hatem, Trivedi, Lira, Qin, Wang, and
  Liao]{wu2026boxworld}
Z.~Wu, A.~Hatem, C.~Trivedi, W.~M.~P. Lira, X.~Qin, L.~Wang, and R.~Liao.
\newblock {BoxWorld: Let World Engines Control and Video Generators Render}.
\newblock \emph{arXiv preprint
  \href{https://arxiv.org/abs/2610.36371}{arXiv:2610.36371}},
  2026{\natexlab{c}}.

\bibitem[Xiao et~al.(2026)Xiao, Dai, Qin, Song, et~al.]{xiao2026joyai}
Y.~Xiao, W.~Dai, X.~Qin, L.~Song, et~al.
\newblock {JoyAI-Video-Edit: Real-Time Open-Ended Video Editing with
  Autoregressive Diffusion}.
\newblock \emph{arXiv preprint
  \href{https://arxiv.org/abs/2608.03974}{arXiv:2608.03974}}, 2026.

\bibitem[Xiao et~al.(2025)Xiao, Lan, Zhou, Ouyang, Yang, Zeng, and
  Pan]{xiao2025worldmem}
Z.~Xiao, Y.~Lan, Y.~Zhou, W.~Ouyang, S.~Yang, Y.~Zeng, and X.~Pan.
\newblock {WorldMem: Long-term Consistent World Simulation with Memory}.
\newblock In \emph{Advances in Neural Information Processing Systems},
  volume~38, 2025.
\newblock \doi{10.52202/085713-1659}.

\bibitem[{Xmax AI}(2026)]{xmax2026x2}
{Xmax AI}.
\newblock {XMax X2.0: Real-Time Interactive Video Model}.
\newblock \href{https://platform.xmax.ai/docs/resources/changelog}{Official
  platform and SDK changelog}, July 2026.

\bibitem[Yang et~al.(2025)Yang, Huang, Chu, Xiao, Zhao, Wang, Li, Xie, Chen,
  Lu, Han, and Chen]{longlive}
S.~Yang, W.~Huang, R.~Chu, Y.~Xiao, Y.~Zhao, X.~Wang, M.~Li, E.~Xie, Y.~Chen,
  Y.~Lu, S.~Han, and Y.~Chen.
\newblock {LongLive: Real-time Interactive Long Video Generation}.
\newblock \emph{arXiv preprint
  \href{https://arxiv.org/abs/2509.22622v2}{arXiv:2509.22622}}, 2025.

\bibitem[Yin et~al.(2026)Yin, Cheng, Qin, Huang, Li, and Ding]{yin2026autoue}
L.~Yin, W.~Cheng, Z.~Qin, T.~Huang, Y.~Li, and G.~Ding.
\newblock {AutoUE: Automated Generation of 3D Games in Unreal Engine via
  Multi-Agent Systems}.
\newblock In
  \emph{\href{https://aclanthology.org/2026.findings-acl.111/}{Findings of the
  Association for Computational Linguistics: ACL 2026}}, pages 2341--2364,
  2026.

\bibitem[Yin et~al.(2023)Yin, Gharbi, Zhang, Shechtman, Durand, Freeman, and
  Park]{dmd}
T.~Yin, M.~Gharbi, R.~Zhang, E.~Shechtman, F.~Durand, W.~T. Freeman, and
  T.~Park.
\newblock {One-step Diffusion with Distribution Matching Distillation}.
\newblock \href{https://arxiv.org/abs/2311.18828}{arXiv:2311.18828}, 2023.

\bibitem[Yin et~al.(2024)Yin, Gharbi, Park, Zhang, Shechtman, Durand, and
  Freeman]{dmd2}
T.~Yin, M.~Gharbi, T.~Park, R.~Zhang, E.~Shechtman, F.~Durand, and W.~T.
  Freeman.
\newblock {Improved Distribution Matching Distillation for Fast Image
  Synthesis}.
\newblock \href{https://arxiv.org/abs/2405.14867}{arXiv:2405.14867}, 2024.

\bibitem[Ying et~al.(2026)Ying, Hu, Ren, Li, Chen, Wang, Cao, Cai, and
  Ding]{ying2026wbench}
K.~Ying, H.~Hu, S.~Ren, J.~Li, F.~Chen, Z.~Wang, X.~Cao, X.~Cai, and H.~Ding.
\newblock {WBench: A Comprehensive Multi-turn Benchmark for Interactive Video
  World Model Evaluation}.
\newblock \emph{arXiv preprint
  \href{https://arxiv.org/abs/2605.25874}{arXiv:2605.25874}}, 2026.

\bibitem[Yu et~al.(2025)Yu, Qin, Wang, Wan, Zhang, and Liu]{yu2025gamefactory}
J.~Yu, Y.~Qin, X.~Wang, P.~Wan, D.~Zhang, and X.~Liu.
\newblock {GameFactory: Creating New Games with Generative Interactive Videos}.
\newblock In
  \emph{\href{https://openaccess.thecvf.com/content/ICCV2025/html/Yu_GameFactory_Creating_New_Games_with_Generative_Interactive_Videos_ICCV_2025_paper.html}{IEEE/CVF
  International Conference on Computer Vision}}, pages 11590--11599, 2025.

\bibitem[Zhan et~al.(2026)Zhan, Wang, Zhang, Zhu, Sun, Fang, Yu, Guo, and
  Fu]{zhan2026magpie}
X.~Zhan, X.~Wang, X.~Zhang, H.~Zhu, T.~Sun, P.~Fang, J.~Yu, Y.~Guo, and D.~Fu.
\newblock {Magpie: Real-Time World Renderer for Interactive Games}.
\newblock \emph{arXiv preprint
  \href{https://arxiv.org/abs/2608.27168}{arXiv:2608.27168}}, 2026.

\bibitem[Zhang et~al.(2026{\natexlab{a}})Zhang, Jiang, Chen, Wang,
  et~al.]{zhang2026vidus2}
J.~Zhang, K.~Jiang, J.~Chen, X.~Wang, et~al.
\newblock {Vidu S2: Real-Time Interactive, Editable, and Spatial Video
  Generation}.
\newblock \emph{arXiv preprint
  \href{https://arxiv.org/abs/2609.11638}{arXiv:2609.11638}},
  2026{\natexlab{a}}.

\bibitem[Zhang et~al.(2026{\natexlab{b}})Zhang, Qian, Kong, Jiang, Wang, Chen,
  Li, Chen, Wu, Chen, Li, Huang, Zhang, Li, Yu, Chen, Cheng, and
  Zhou]{zhang2026matrixgame4}
J.~Zhang, R.~Qian, K.~Kong, B.~Jiang, P.~Wang, S.~Chen, T.~Li, G.~Chen, J.~Wu,
  C.~Chen, Y.~Li, K.~Huang, H.~Zhang, Y.~Li, W.~Yu, G.~Chen, Y.~Cheng, and
  Y.~Zhou.
\newblock {Matrix-Game 4.0: Harnessing Video Models for Real-Time Playable Game
  Worlds}.
\newblock \emph{arXiv preprint
  \href{https://arxiv.org/abs/2610.27999}{arXiv:2610.27999}},
  2026{\natexlab{b}}.

\bibitem[Zhang et~al.(2018)Zhang, Isola, Efros, Shechtman, and Wang]{lpips}
R.~Zhang, P.~Isola, A.~A. Efros, E.~Shechtman, and O.~Wang.
\newblock {The Unreasonable Effectiveness of Deep Features as a Perceptual
  Metric}.
\newblock \href{https://arxiv.org/abs/1801.03924}{arXiv:1801.03924}, 2018.

\bibitem[Zhao et~al.(2026)Zhao, Pan, He, Yu, Chen, Ye, Liu, Xie, and
  Han]{zhao2026sanastreaming}
Y.~Zhao, Y.~Pan, Q.~He, J.~Yu, J.~Chen, T.~Ye, H.~Liu, E.~Xie, and S.~Han.
\newblock {SANA-Streaming: Real-time Streaming Video Editing with Hybrid
  Diffusion Transformer}.
\newblock \emph{arXiv preprint
  \href{https://arxiv.org/abs/2605.30409}{arXiv:2605.30409}}, 2026.

\bibitem[Zheng et~al.(2024)Zheng, Yin, Xie, Sun, Huang, Yu, Cao, Kozyrakis,
  Stoica, Gonzalez, Barrett, and Sheng]{zheng2024sglang}
L.~Zheng, L.~Yin, Z.~Xie, C.~Sun, J.~Huang, C.~H. Yu, S.~Cao, C.~Kozyrakis,
  I.~Stoica, J.~E. Gonzalez, C.~Barrett, and Y.~Sheng.
\newblock {SGLang: Efficient Execution of Structured Language Model Programs}.
\newblock In \emph{Advances in Neural Information Processing Systems},
  volume~37, 2024.

\bibitem[Zhuang et~al.(2026)Zhuang, Zhang, Bian, Li, Luo, Liu, Jin, Zhang, He,
  Zhang, Li, Huang, Xue, and Duan]{sgf}
J.~Zhuang, S.~Zhang, Y.~Bian, Y.~Li, Y.~Luo, Y.~Liu, W.~Jin, S.~Zhang, X.~He,
  X.~Zhang, H.~Li, H.~Huang, Z.~Xue, and N.~Duan.
\newblock {Self Gradient Forcing: Native Long Video Extrapolation}.
\newblock \href{https://arxiv.org/abs/2607.20368}{arXiv:2607.20368}, 2026.

\end{thebibliography}
\endgroup

\appendix

\end{document}